\documentclass{article}

\usepackage{PRIMEarxiv}

\usepackage[utf8]{inputenc} % allow utf-8 input
\usepackage[T1]{fontenc}    % use 8-bit T1 fonts
\usepackage{hyperref}       % hyperlinks
\usepackage{url}            % simple URL typesetting
\usepackage{booktabs}       % professional-quality tables
\usepackage{amsfonts}       % blackboard math symbols
\usepackage{nicefrac}       % compact symbols for 1/2, etc.
\usepackage{microtype}      % microtypography
\usepackage{lipsum}
\usepackage{fancyhdr}       % header
\usepackage{graphicx}       % graphics
\graphicspath{{media/}}     % organize your images and other figures under media/ folder
\usepackage{comment}

\usepackage{algorithm2e}
\usepackage{matlab-prettifier}
\usepackage{amsmath,amsfonts,amssymb,amsthm}
\usepackage{mathtools}
\usepackage{commath}
\usepackage{listings}
\usepackage{parskip}
\usepackage{amsmath}%Para ecuaciones
\usepackage{float}

\usepackage{caption}
\usepackage{subcaption}

\usepackage[most]{tcolorbox}
\newtcolorbox{mybox}[2][]{text width=0.9\textwidth,fontupper=\normalsize,
fonttitle=\bfseries\sffamily\normalsize, colbacktitle=black,enhanced,
attach boxed title to top left={yshift=-2mm,xshift=3mm},
boxed title style={sharp corners},top=5pt,bottom=3pt,
title=#2,colback=white}

\usepackage{tikz}
\usepackage{listofitems} % for \readlist to create arrays
\usetikzlibrary{arrows.meta} % for arrow size
\usepackage[outline]{contour} % glow around text
\contourlength{1.4pt}

\tikzset{>=latex} % for LaTeX arrow head
\usepackage{xcolor}
\colorlet{myred}{red!80!black}
\colorlet{myblue}{blue!80!black}
\colorlet{mygreen}{green!60!black}
\colorlet{myorange}{orange!70!red!60!black}
\colorlet{mydarkred}{red!30!black}
\colorlet{mydarkblue}{blue!40!black}
\colorlet{mydarkgreen}{green!30!black}
\tikzstyle{node}=[thick,circle,draw=myblue,minimum size=22,inner sep=0.5,outer sep=0.6]
\tikzstyle{node in}=[node,green!20!black,draw=mygreen!30!black,fill=mygreen!25]
\tikzstyle{node hidden}=[node,blue!20!black,draw=myblue!30!black,fill=myblue!20]
\tikzstyle{node convol}=[node,orange!20!black,draw=myorange!30!black,fill=myorange!20]
\tikzstyle{node out}=[node,red!20!black,draw=myred!30!black,fill=myred!20]
\tikzstyle{connect}=[thick,mydarkblue] %,line cap=round
\tikzstyle{connect arrow}=[-{Latex[length=4,width=3.5]},thick,mydarkblue,shorten <=0.5,shorten >=1]
\tikzset{ % node styles, numbered for easy mapping with \nstyle
  node 1/.style={node in},
  node 2/.style={node hidden},
  node 3/.style={node out},
}
\usepackage[toc,page]{appendix}

\title{Predicting and explaining nonlinear material response using deep Physically Guided Neural Networks with Internal Variables
}

\author{
  Javier Orera-Echeverria \\
  University of Zaragoza \\
  Mariano Esquillor, s.n. 50018, Zaragoza\\
  \texttt{719101@unizar.es} \\
   \And
  Jacobo Ayensa-Jiménez \\
  Aragon Institute of Engineering Research \\
  University of Zaragoza \\
  Mariano Esquillor, s.n. 50018, Zaragoza \\
  \texttt{jacoboaj@unizar.es} \\
  \And
  Manuel Doblaré$^*$ \\
  Aragon Institute of Engineering Research \\
  University of Zaragoza \\
  Mariano Esquillor, s.n. 50018, Zaragoza \\
  \texttt{mdoblare@unizar.es} \\
}

\usepackage{makeidx}
\makeindex
\index{key}

\begin{document}
\maketitle

\begin{abstract}
Nonlinear materials are often difficult to model with classical state model theory because they have a complex and sometimes inaccurate physical and mathematical description or we simply do not know how to describe such materials in terms of relations between external and internal variables. In many disciplines, Neural Network methods have arisen as powerful tools to identify very complex and non-linear correlations. In this work, we use the very recently developed concept of Physically Guided Neural Networks with Internal Variables (PGNNIV) to discover constitutive laws using a model-free approach and training solely with measured force-displacement data. PGNNIVs make a particular use of the physics of the problem to enforce constraints on specific hidden layers and are able to make predictions without internal variable data. We demonstrate that PGNNIVs are capable of predicting both internal and external variables under unseen load scenarios, regardless of the nature of the material considered (linear, with hardening or softening behavior and hyperelastic), unravelling the constitutive law of the material hence explaining its nature altogether, placing the method in what is known as eXplainable Artificial Intelligence (XAI).

\end{abstract}

% keywords can be removed

\section{Introduction}
\label{section:Introduction}

It is of common knowledge that our everyday life is being dramatically challenged by Big Data and Artificial Intelligence (AI). According to the International Data Corporation, the Global Datasphere (the summation of all data, whether created, captured, or replicated) will grow from 33 ZB in 2018 to 175 ZB by 2025 \cite{rydning2018digitization}. This is mainly due to the explosion of social networks, e-commerce and marketing, and the extension of the Internet of Things (IoT). Among the top eight companies in terms of market capitalization, five are based on data value leverage \cite{MVC2022}. This huge amount of available information justifies the prosperity of data science and data-based decision in fields as diverse as sociology, economics, engineering and medicine.

As a response, Machine Learning (ML) methods have become today one of the main tools in business, but also in science and technology. These methodologies enable the extraction of information from data that would be intractable by means of traditional methods \cite{bishop2006pattern}. They try to mimic the process of human knowledge acquisition and structuring taking advantage of the aforementioned advances in data generation, management, and storage, as well as huge improvements in the performance of computers and algorithms \cite{atzori2010internet}. In the special case of Scientific ML \cite{thiyagalingam2022scientific}, the natural adaptation of many supervised as well as unsupervised ML algorithms to the vectorized representation that most physical problems exhibit, makes the study of the convergence between both of special importance.

Data-driven methods are used in many different physical disciplines such as chemical and electrical processes~\cite{nuhic2013health}, biology~\cite{xue2015computational}, spoken language recognition ~\cite{lemon2012data} and a long etcetera. However, the link between classical physical modelling and data-driven methods has not been quite clear so far, since the physical description of most systems was built on the basis of empirical knowledge rather than large data-bases. Nonetheless, the new era of computation and Big Data has opened new perspectives where data can be incorporated in this physical description in a consistent and comprehensive way. Promising improvements can be therefore spotted, such as new forms of empiricism that declare ''the end of theory'' and the impending advancement of data-driven methods over knowledge-driven science \cite{kitchin2014big}.

One of the most prominent strategies that has proven to be especially prolific in recent years is the use of Artificial Neural Networks (ANNs). Since 1958, when Rosenblatt developed the \emph{perceptron} \cite{rosenblatt1958perceptron}, many works have been devoted to ANNs, some of them demostrating their character as universal approximators \cite{cybenko1989approximations,hornik1991approximation,pinkus1999approximation,lu2017expressive,hanin2019universal}. However, it has been only in the last decade of the XX\textsuperscript{th} century, thanks to the important progress of high performance computing capabilities and the combination of back-propagation \cite{chauvin1995backpropagation} with stochastic gradient descent \cite{amari1993backpropagation} algorithms, that ANNs have become a booming technology. The progress has accelerated in the last decade with the advent of Convolutional Neural Networks (CNNs) \cite{krizhevsky2012imagenet} and Recurrent Neural Networks (RNNs) \cite{graves2013speech}, in what is known nowadays as Deep Learning (DL) \cite{lecun2015deep}, and culminating with the attention mechanism \cite{vaswani2017attention}, transformer models and generative AI, whose impact has gained nowadays great popularity thanks to Large Language Models (LLMs) \cite{teubner2023welcome}.

Leaving aside the progress of the DL as a research field, AI approaches have changed the way we conceive science. On the one hand, AI has been used to discover the hidden physical structure of the data and unravel the equations of a system \cite{schmidt2009distilling, brunton2016discovering,udrescu2020ai}. On the other hand, the tremendous predictive power of AI has been blended with the scientific consistency of the explicit mathematical representation of physical systems through the concept of \emph{data-driven} models for simulation-based engineering and sciences (SBES) \cite{chinesta2022empowering}. The latter in turn may be done by the combination of raw data and physical equations \cite{kirchdoerfer2016data,ayensa2019unsupervised}, by enforcing a metriplectic structure to the model, related with the fulfillment of thermodynamic laws \cite{cueto2023thermodynamics} or by defining the specific structure of the model \cite{greydanus2019hamiltonian,jin2020sympnets}. These novel \emph{data-driven science} approaches, coined as  Scientific Machine Learning or Physics-Informed Machine Learning (PIML) arise therefore with the main purpose of turning apparently not-physically meaningful data-driven models, where approaches such as ANNs have excelled, into physics-aware models. 

However, the interplay between data and physical sciences has not been exempt from setbacks of different nature. In fact, the use of complex DL models does not fit well with the study of physical problems. Furthermore, in many physical problems, many variables are involved, interacting in complex and non-stationary ways. This requires huge amounts of data to get accurate predictions using ANN techniques, sometimes in regions of the solution space that are difficult to access or uncommon and therefore difficult to be sampled. As a consequence, due to the bias–variance trade-off, poorly extrapolation capacity is obtained out of the the usual data range for models looking for recreating complex physics \cite{ying2019overview}. 

In addition, a physically based model is not only useful for making predictions, but also to help in acquiring new knowledge by the interpretation of its structure, parameters, and mathematical properties. Physical interpretability is, in most  cases, at least as important as predictive performance. However, it is known that interpretability is one of the main weaknesses of ANNs, as the acquired  knowledge is encoded in the strength of multiple connections, rather than stored at specific locations. That explains the huge efforts that are currently being made towards “whitening” the “black-box” character of ANN \cite{mahendran2015understanding,shwartz2017opening} in what has been coined as eXplainable Artificial Intelligence (XAI) \cite{adadi2018peeking,xu2019explainable}. In the context of data-driven simulation-based engineering and sciences (DDSBES) \cite{ibanez2017data}. Two ways of proceeding can be distinguished: building specific ANNs structures endowed with the problem equations \cite{ayensa2021prediction,masi2021thermodynamics}, also known as the \emph{inductive bias} approach, and/or by regularizing the loss function using this same physical information  \cite{karpatne2017theory,muralidhar2018incorporating, raissi2019physics}.

The approach to be followed depends decisively upon the data availability and the way in which this data is used. If we follow a supervised approach, there are two possibilities. 

\begin{itemize}
    \item The first one assumes that we know the whole physics of the problem. In that situation, supervised ML is used for the sake of computational requirements. In that sense, ANNs act as Reduced Order Models (ROMs) that can be used as a surrogate in problems involving optimization or control. Hybridizing DL with physical information is a way of improving standard DL methods in terms of data requirements, less expensive training or noise filtering at the evaluation step, thanks to regularization \cite{ayensa2021prediction,karpatne2017theory}. Therefore, we are interested in the predictive capacity of the approach.
    \item The second one assumes that we know some of the physics of the problem. In that situation, we are interested in knowing the hidden physics that remains unknown, expressed in terms of some model parameters \cite{raissi2019physics} or functional relations \cite{ayensa2022understanding}. For that reason, we are rather interested in the explanatory character of the method.
\end{itemize}

In other situations, we follow unsupervised approaches. This may be indeed due to two different possibilities:

\begin{itemize}
    \item If we know the whole physics of the problem, the use of ANNs is merely instrumental and is used as an alternative way of solving numerically some system of Partial Differential Equations (PDEs) that require an important computational effort \cite{lagaris2000neural}.
    \item When some of the physics of the problem are unknown, the intrinsic variability of the data (for instance when measuring spatial or temporal fields) may be exploited for an unsupervised discovery of some hidden constitutive models \cite{flaschel2022discovering,tartakovsky2018learning}. For that reason, we are rather interested in the explanatory character of the method.
\end{itemize}

However, in the context of the IoT, where data quantity generally dominates over data quality, the data availability and variability is key, and it is difficult to guarantee whether, considering for instance the specific case of computational solid mechanics,  ``the combination of geometry and loading generates sufficiently diverse and heterogeneous strain states to train a generalizable constitutive model with just a single experiment'' \cite{thakolkaran2022nn}. Therefore, there are no other alternatives rather than introducing all the control or measurable variables in the workflow, while maintaining the desirable properties of the PIML approach, namely, its ability to get fast predictions in real time (for optimization and control issues) together with its explanatory capacity.

In this work, we demonstrate, how, in the context of computational solid mechanics, Physically Guided Neural Networks with Internal Variables (PGNNIVs) enable the compliance with these requirements particularly well. PGNNIVs comprise ANNs that are able to incorporate some of the known physics of the problem, expressed in terms of some measurable variables (for instance forces and displacements) and some hidden ones (for instance stresses). Their predictive character, improving many of the features of conventional ANN \cite{ayensa2021prediction}, allow for fast and accurate predictions. In addition, PGNNIVs are able to unravel the constitutive equations of different materials from unstructured data, that is, uncontrolled test data obtained from system monitorization.

The content of this paper is structured as follows. First, the methodology is described, including a brief overview of the state of the art, the use of PGNNIVs in computational mechanics, the computational treatment of the physical tensorial fields and operators, as well as the data-set generation and training process. Then, the main results are presented, of both the predictive and explanatory capacity of the method. Finally, conclusions are summarized, together with the main limitations and a brief overview of future work.

\section{Methods}
\label{section:Methods}

\subsection{Brief overview of Physics Informed Machine Learning in computational solid mechanics}
\label{subsection:RW}

PDEs are the standard way to describe physical systems under the continuum setting thanks to their overarching capacity to model extremely different and complex phenomena. However,  analytic solutions are most times difficult or even impossible to find. That is the reason why numerical methods have become the universal tool to obtain approximate but accurate solutions to PDEs. These methods consider a given discretization in space and time, which results in an algebraic (in general non-linear) system that is then solved by means of standard matrix manipulation. 

In the last three decades, however, attempts to solve PDEs from a data-driven point of view have been numerous. First tentatives \cite{van1995neural,lagaris1998artificial} generalized earlier ideas for Ordinary Differential Equations (ODEs). Since then, many different approaches have been proposed, from collocation methods \cite{rudd2015constrained,sirignano2018dgm,abueidda2020deep}, variational/energy approaches \cite{yu2018deep,samaniego2020energy,nguyen2020deep}, loss regularization using physical or domain knowledge \cite{karpatne2017physics,stewart2017label,muralidhar2018incorporating,magiera2020constraint}, to the most recent approaches using automatic differentiation \cite{raissi2019physics}, that is nowadays known as Physics-Informed Neural Networks (PINNs). Other works have extensively tried to address this challenge using the stochastic representations of high-dimensional parabolic PDEs \cite{weinan2017deep,han2018solving,hure2020deep}.

In order to provide the data-driven models with a meaningful physical character, remarkable efforts have been done in the recent years to embed physical information into data-driven  descriptions. The potential of solving inverse problems with linear and non-linear behavior in solid mechanics, for example, has been explored using DL methods ~\cite{tamaddon2020data}, where the forward problem is solved first to create a database, which is then used to train the ML algorithms and determine the boundary conditions from assumed measurements. Other approaches initially build a constitutive model into the framework by enforcing constitutive constraints, and  aim at calibrating the constitutive parameters ~\cite{hamel2022calibrating}.

In this context, clearly differentiating between external and internal variables becomes an important factor when approaching complex physical problems, but this is always disregarded from a data-driven viewpoint. External variables are those observable, measurable variables of the system, that can be obtained directly from physical sensors such as position, temperature or forces; internal variables are non-observable (not directly measurable) variables, that integrate locally other observable magnitudes and depend on the particular internal structure of the system~\cite{ayensa2020application}. This is very important to consider in the ML framework since predictions associated to an internal state model require explicitly the definition of the cloud of experimental values that identifies the internal state model~\cite{karapiperis2021data}. This implies ``measuring'', or better, assuming values for non-observable variables \cite{rao2021physics,bharadwaja2022physics}. This is for example, the case of stresses in continuum mechanics that can be determined \emph{a priori} only after making strong assumptions such as their uniform distribution in the center section of a sample under uniform tension.

An alternative is the use of PGNNIVs. This new methodological appraisal permits to predict the values of the internal variables by mathematically constraining some hidden layers of a deep ANN (whose structural topology is appropriately predefined) by means of the fundamental laws of the continuum mechanics such as conservation of energy or mechanical momenta that relate internal non-measurable with external observable variables. With this, it is possible to transform a pure ML based model into a physically-based one without giving up the powerful tools of DL, including the implicit correlation between observable data and the derived predictive capacity. This way, not only the real internal variables of the problem are predicted, but also the data needed to train the network decreases, convergence is reached faster, data noise is better filtered and the extrapolation capacity out of the range of the training data-set is improved, as recently demonstrated in ~\cite{ayensa2021prediction}. In line with the terminology and general framework used in PIML \cite{karniadakis2021physics}, PGNNIVs showcase an intuitive interplay between an \emph{inductive bias} approach and a \emph{learning bias} appraisal, where physical constrains are incorporated by means of a physics-informed likelihood, i.e. additional terms in the loss function (also known as \emph{collocation} or \emph{regularization losses}).
 
\subsection{Physically Guided Neural Networks with Internal Variables in computational mechanics} \label{subsection:PGNNIV}

\subsubsection{Revisiting PGNNIVs} \label{subsubsection:Revisiting_PGNNIV}

PGNNIVs are in essence a generalization of PINNS. In the latter, physical equations constrain the values of output variables to belong to a certain physical manifold that is built from the information provided by the data and the specific form of the PDE considered. One of the shortcomings of PINNs is that, in general, only simple ordinary PDEs that contain a few free parameters and are closed-form can be considered. In contrast, PGNNIVs do not only apply to scenarios where PDEs involve many parameters and have complex forms, but also to those where a mathematical description is not available. In fact, this new paradigm embodies a unique architecture where the values of the neuron variables in some intermediate layers acquire physical meaning in an unsupervised way, providing the network with an inherent explanatory capacity. 

The main differentiating features that, up to the authors' knowledge, constitute a relevant contribution to the advances of data-driven physical modelling and its particular application to nonlinear mechanics are two-folded: first, physical constraints are applied in predefined internal layers (PILs), in contrast to previous works on PINNs. Secondly, and even most important, PGNNIVs are able to predict and explain the nature of the system all at once, i.e. predictability of the variables as well as explainability of the constitutive law are ensured and learned altogether. In all related works, modeling assumptions on the constitutive law of the correspondent materials are directly imposed, so that the material response complies with certain constrains. On the contrary, PGNNIVs only enforce universal laws of the system (e.g. balance equations) and no prior knowledge on the constitutive model is incorporated.

Classical Deep Neural Networks (DNNs) are often represented as \emph{black boxes} that theoretically can compute and learn any kind of function correlating the input and output data ~\cite{nielsen2015neural}. In particular, they perform very well in  areas of science and technology where complex functions convey good approximations of the governing phenomena. Although there exist some heuristic rules ~\cite{walczak1999heuristic}, these \emph{black boxes} are usually trained via trial and error. Adding a physical meaning to the hidden layers and constraining them by adding an extra term to the cost function, has already proven to have significant advantages such as less data requirements, higher accuracy and faster convergence in real physical problems, as well as model unravelling capacities~\cite{ayensa2021prediction}. The basic principles of PGNNIVs are briefly exposed in the next lines.
 
Let us consider a set of continuous Partial Differential Equations (PDEs) of the form
\begin{subequations} \label{eq:PDEs}
\begin{align}
\mathcal{F}(u,v)&=f, \text{ in } \Omega, \label{eq:PDEs_1}\\ 
\mathcal{G}(u,v)&=g, \text{ in } \partial\Omega, \label{eq:PDEs_2}\\ 
\mathcal{H}(u)&=v, \text{ in } \Omega, \label{eq:PDEs_3}
\end{align}
\end{subequations}
where $u$ and $v$ are the unknown fields of the problem, $\mathcal{F}$ and $\mathcal{H}$ are functionals representing the known and unknown
physical equations of the specific problem, $\mathcal{G}$ is a functional that specifies the boundary conditions, and $f$ and $g$ are known fields.

The continuous problem has its analogous discretized representation in finite-dimensional spaces in terms of vectorial functions $\boldsymbol{F}$, $\boldsymbol{G}$ and $\boldsymbol{H}$ and nodal values $\boldsymbol{u}$, $\boldsymbol{v}$, $\boldsymbol{f}$ and $\boldsymbol{g}$. Particularly, $\boldsymbol{u}$ are the solution field nodal values and $\boldsymbol{v}$ are the unknown internal field variables at the different nodes. The discretization may be done using any discretization technique, such as the Finite Element Method (FEM). Hence, Eqs. (\ref{eq:PDEs}) become
\begin{subequations} \label{eq:DPDEs}
\begin{align} 
\boldsymbol{F}(\boldsymbol{u},\boldsymbol{v})=\boldsymbol{f}, \text{ in } \Omega, \label{eq:DPEs_1}\\  
\boldsymbol{G}(\boldsymbol{u},\boldsymbol{v})=\boldsymbol{g}, \text{ in } \partial\Omega, \label{eq:DPEs_2}\\ 
\boldsymbol{H}(\boldsymbol{u})=\boldsymbol{v}, \text{ in } \Omega. \label{eq:DPEs_3}
\end{align}
\end{subequations}

A PGNNIV may be defined for a problem of type (\ref{eq:DPDEs}) in the following terms:
\begin{subequations} \label{eq:pgnniv}
\begin{align}
\boldsymbol{y}&=\mathsf{Y}(\boldsymbol{x}), \nonumber\\ 
\boldsymbol{v}&=\mathsf{H}(\boldsymbol{u}),\nonumber\\
\boldsymbol{x}&=\boldsymbol{I}(\boldsymbol{u},\boldsymbol{f},\boldsymbol{g}),\nonumber\\
\boldsymbol{y}&=\boldsymbol{O}(\boldsymbol{u},\boldsymbol{f},\boldsymbol{g}),\nonumber\\
\boldsymbol{R}(\boldsymbol{u},\boldsymbol{v},\boldsymbol{f},\boldsymbol{g})&=0,\nonumber
\end{align}
\end{subequations}
where:
\begin{enumerate}
    \item $\boldsymbol{u}$, $\boldsymbol{f}$ and $\boldsymbol{g}$ are the measurable variables of the problem.
    \item $\boldsymbol{x}$ and $\boldsymbol{y}$ are the input and output variables respectively and will be defined depending on which relation $\boldsymbol{x} \mapsto \boldsymbol{y}$ wants to be predicted.
    \item $\boldsymbol{I}$ and $\boldsymbol{O}$ are functions that compute the input $\boldsymbol{x}$ and the output $\boldsymbol{y}$ of the problem from the measurable variables. In other words, functions $\boldsymbol{I}$ and $\boldsymbol{O}$ define the data used as starting point to make predictions, $\boldsymbol{x}$, and the data that we want to predict, that is, $\boldsymbol{y}$.
    \item $\boldsymbol{R}$ are the physical constraints, related to the relations given by $\boldsymbol{F}$ and $\boldsymbol{G}$.
    \item $\mathsf{Y}$ and $\mathsf{H}$ are DNN models:
    \begin{itemize}
        \item  $\mathsf{Y}$ is the \textbf{predictive model}, whose aim is to infer accurate values for the output variables for a certain input set, that is, to surrogate the relation $\boldsymbol{x} \mapsto \boldsymbol{y}$.
        \item $\mathsf{H}$ is the \textbf{explanatory model}, whose objective is to unravel the hidden physics of the relation $\boldsymbol{u} \mapsto \boldsymbol{v}$. 
        \end{itemize}
\end{enumerate}

\subsubsection{Adaptation to computational solid mechanics.} \label{subsection:PGNNIV_in_CM}
Our aim now is to reframe Eqs. (\ref{eq:PDEs}) in the context of solid mechanics. To fix ideas, although it is not difficult to adapt the methodology to other constitutive models, we restrict this analysis to hyperelastic solids with constant and known density $\rho$. First, we have to consider equilibrium equations (momentum conservation) in the domain $\Omega$. In spatial coordinates, equilibrium reads
\begin{equation} \label{eq:equilibrium_sc}
\mathrm{div} (\boldsymbol{\sigma}) + \rho \boldsymbol{b} = \boldsymbol{0},
\end{equation}
where $\boldsymbol{\sigma}$ is Cauchy stress tensor, $\rho$ is the density, $\boldsymbol{b}$ the spatial volumetric body force field and $\mathrm{div}$ is the divergence operator in spatial coordinates. In material coordinates, equilibrium reads 
\begin{equation} \label{eq:equilibrium_mc}
\mathrm{DIV} (\boldsymbol{P}) + \rho \boldsymbol{B} = \boldsymbol{0},
\end{equation}
where now $\boldsymbol{P}=\mathrm{det}(\boldsymbol{F})\boldsymbol{\sigma}\boldsymbol{F}^{-\intercal}$ is the first Piola-Kirchhoff stress tensor, $\boldsymbol{B}= \mathrm{det}(\boldsymbol{F})\boldsymbol{b}$ is the reference volumetric body force field, and $\mathrm{DIV}$ is the divergence operator in material coordinates.

If $\boldsymbol{\xi} = \chi(\boldsymbol{\Xi})$ is the motion function that relates spatial ($\boldsymbol{\xi}$) and material ($\boldsymbol{\Xi}$) coordinates, we define the deformation gradient tensor $\boldsymbol{F}$ as
\begin{equation}\label{eq:Fdef}
    \boldsymbol{F} = \mathrm{GRAD}(\chi) = \mathrm{GRAD} \otimes \boldsymbol{\xi},
\end{equation}
where $\mathrm{GRAD}$ is the gradient operator in material coordinates.
Eq. (\ref{eq:Fdef}) shows that $\boldsymbol{F}$ is a potential tensor field so $\boldsymbol{F}$ must satisfy the following compatibility equation in each connected component of $\Omega$:
\begin{equation}\label{eq:compatibility}
    \mathrm{ROT}(\boldsymbol{F}) = \boldsymbol{0},
\end{equation}
where $\mathrm{ROT}$ is the rotational in material coordinates.

For hyperelastic materials, the Cauchy stress tensor $\boldsymbol{\sigma}$ is related to the deformation gradient tensor by means of the equation
\begin{equation} \label{eq:constitutive}
\boldsymbol{\sigma} = \frac{1}{\mathrm{det}(\boldsymbol{F})}\frac{\partial \Psi}{\partial \boldsymbol{F}} \cdot \boldsymbol{F}^\intercal,
\end{equation}
where $\Psi$ is the strain energy function expressed as a function of the deformation state given by $\boldsymbol{F}$, $\Psi=\mathfrak{F}(\boldsymbol{F})$. Obtaining this particular function is the subject of research of material sciences and many approaches are possible, ranging from phenomenological descriptions to mechanistic and statistical models. 

Finally, Eqs. (\ref{eq:equilibrium_sc}) or (\ref{eq:equilibrium_mc}), (\ref{eq:compatibility}) or (\ref{eq:Fdef}), and (\ref{eq:constitutive}) must be supplemented with appropriate boundary conditions. We distinguish here between \emph{essential} boundary conditions and \emph{natural} boundary conditions. The former define the motion of the solid at some boundary points $\Gamma_E \subset \partial \Omega$:
\begin{equation} \label{eq:boundary_essential}
\boldsymbol{\xi} = \bar{\boldsymbol{\xi}},
\end{equation}
whereas the latter define the traction vector at some other boundary points $\Gamma_N \subset \partial \Omega$:
\begin{equation} \label{eq:boundary_natural_sc}
\boldsymbol{\sigma} \cdot \boldsymbol{n} = \bar{\boldsymbol{t}},
\end{equation}
where $\boldsymbol{n}$ is the outwards normal vector (in the spatial configuration) and $\bar{\boldsymbol{\xi}}$ and $\bar{\boldsymbol{t}}$ are known values of the solid motion and spatial traction forces respectively. The material analogous of Eq. (\ref{eq:boundary_natural_sc}) is
\begin{equation} \label{eq:boundary_natural_mc}
\boldsymbol{P} \cdot \boldsymbol{N} = \bar{\boldsymbol{T}},
\end{equation}
where now $\boldsymbol{N}$ and $\bar{\boldsymbol{T}}$ are the material analogous of $\boldsymbol{n}$ and $\bar{\boldsymbol{t}}$.

Let us assume that it is possible to measure (for instance using Digital Image Correlation techniques \cite{chu1985applications}) the system response in terms of its motion given by the map $\boldsymbol{\xi} = \chi(\boldsymbol{\Xi})$, that is, $\boldsymbol{\xi}$ is a measurable variable. Let us also assume that we can measure the volumetric loads, $\boldsymbol{b} = \boldsymbol{b}(\boldsymbol{\xi})$, as well as the prescribed traction forces $\bar{\boldsymbol{t}}$ (or, in an equivalent manner, $\boldsymbol{B}=\boldsymbol{B}(\boldsymbol{\Xi})$ and $\bar{\boldsymbol{T}}$). Therefore, using Eqs. (\ref{eq:Fdef}) and  (\ref{eq:constitutive}) it is possible to express:
\begin{subequations}
\begin{align}
    \boldsymbol{F} &= \mathcal{A}(\boldsymbol{\xi}), \\
    \boldsymbol{\sigma} &= \mathcal{B}(\Psi,\boldsymbol{\xi}),
\end{align}
\end{subequations}
for some appropriate differential operators $\mathcal{A}$ and $\mathcal{B}$. Therefore, it is possible to recast hyperelastic solid mechanics as
\begin{subequations} \label{eq:problem_mechanics}
\begin{align}
    \mathrm{div} (\boldsymbol{\sigma}(\Psi,\boldsymbol{\xi})) &= -\rho \boldsymbol{b} \quad \mathrm{in} \quad \Omega, \label{eq:problem_mechanics_1}\\
    \boldsymbol{\xi}  &= \bar{\boldsymbol{\xi}} \quad \mathrm{in} \quad \Gamma_E, \label{eq:problem_mechanics_2} \\
    \boldsymbol{\sigma} \cdot \boldsymbol{n} &= \bar{\boldsymbol{t}} \quad \mathrm{in} \quad \Gamma_N, \label{eq:problem_mechanics_3}\\
    \Psi &= \mathfrak{F}(\boldsymbol{F}(\boldsymbol{\xi})) \quad \mathrm{in} \quad \Omega. \label{eq:problem_mechanics_4}
\end{align}
\end{subequations}
where Eqs. (\ref{eq:problem_mechanics_1}) and (\ref{eq:problem_mechanics_3}) may be eventually substituted by their material analogous. Groupping Eqs. (\ref{eq:problem_mechanics_2}) and (\ref{eq:problem_mechanics_3}), it is clear that we can express computational solid mechanics for hyperelastic materials as
\begin{subequations} \label{eq:problem_mechanics_final}
\begin{align}
\mathcal{F}(\boldsymbol{\xi},\boldsymbol{\sigma})=\boldsymbol{b}, \text{ in } \Omega, \label{eq:problem_mechanics_final_1}\\  
\mathcal{G}(\boldsymbol{\xi},\boldsymbol{\sigma})=(\bar{\boldsymbol{\xi}},\bar{\boldsymbol{t}}) \text{ in } \partial\Omega, \label{eq:problem_mechanics_final_2}\\ 
\mathcal{H}(\boldsymbol{\xi})=\boldsymbol{\sigma}, \text{ in } \Omega, \label{eq:problem_mechanics_final_3}
\end{align}
\end{subequations}
that is in the form of Eqs. (\ref{eq:PDEs}) with $u=\boldsymbol{\xi}$, $v = \boldsymbol{\sigma}$, $f = \boldsymbol{b}$ and $g=(\bar{\boldsymbol{\xi}},\bar{\boldsymbol{t}})$.

\paragraph{Particularization to small strains solid mechanics.} In small strains solid mechanics,  it is common to work with the displacement field $\boldsymbol{U} = \boldsymbol{\xi} - \boldsymbol{\Xi}$ 
and to define the displacement gradient tensor $\boldsymbol{J} = \mathrm{GRAD}(\boldsymbol{U})$. In that case, the constitutive equation is formulated as
\begin{equation}
    \boldsymbol{\sigma} = \mathfrak{G}(\boldsymbol{\varepsilon}),
\end{equation}
where $\boldsymbol \varepsilon$ is the Cauchy small strain tensor
\begin{equation}
    \boldsymbol{\varepsilon} = \mathrm{symgrad}(\boldsymbol{U}) = \frac{1}{2}(\boldsymbol{J} + \boldsymbol{J}^\intercal),
\end{equation}
and $\mathfrak{G}$ is a tensor map.
With these considerations, the equations of the problem are now written as
\begin{subequations} \label{eq:problem_mechanics_ifinal}
\begin{align}
\mathcal{F}(\boldsymbol{U},\boldsymbol{\sigma})&=\boldsymbol{b}, \text{ in } \Omega, \label{eq:problem_mechanics_ifinal_1}\\  
\mathcal{G}(\boldsymbol{U},\boldsymbol{\sigma})&=(\bar{\boldsymbol{U}};\bar{\boldsymbol{t}}), \text{ in } \partial\Omega, \label{eq:problem_mechanics_ifinal_2}\\ 
\mathcal{H}(\boldsymbol{U})&=\boldsymbol{\sigma}, \text{ in } \Omega, \label{eq:problem_mechanics_ifinal_3}
\end{align}
\end{subequations}
again in the form of Eqs. (\ref{eq:PDEs}) with $u=\boldsymbol{U}$, $v = \boldsymbol{\sigma}$, $f = \boldsymbol{b}$ and $g = (\bar{\boldsymbol{U}},\bar{\boldsymbol{t}})$.

Once we have discretized the problem, our aim is to predict a motion field $\boldsymbol{\xi}$ (or a displacement field $\boldsymbol{U}$) from a particular load case, expressed in terms of the volumetric loads and the natural boundary conditions\footnote{It is important to recall that, as essential boundary conditions can be measured as an output variable, it is not necessary to include them as inputs of our problem.}, $\bar{\boldsymbol{t}}$, therefore $\boldsymbol{I}(\boldsymbol{u},\boldsymbol{f},\boldsymbol{g}) = (\boldsymbol{f},\boldsymbol{g}) = (\boldsymbol{b},\bar{\boldsymbol{t}})$. With these last remarks, the PGNNIV problem is stated for finite solid mechanics as
\begin{subequations} \label{eq:problem_solid_mechanics}
\begin{align}
\boldsymbol{\xi}&=\mathsf{Y}(\bar{\boldsymbol{t}})\\ 
\boldsymbol{\sigma}&=\mathsf{H}(\mathrm{KIN}(\boldsymbol{\xi}))\\
\boldsymbol{x}&=(\bar{\boldsymbol{t}},\boldsymbol{b})\\
\boldsymbol{y}&=\boldsymbol{\xi}\\
\boldsymbol{R}(\boldsymbol{\xi},\boldsymbol{\sigma},\bar{\boldsymbol{t}})&=(\mathrm{div}(\boldsymbol{\sigma}) - \rho \boldsymbol{b}; \boldsymbol{\sigma} \cdot \boldsymbol{n}- \bar{\boldsymbol{t}};\boldsymbol{\xi}-\bar{\boldsymbol{\xi}}). 
\end{align}
\end{subequations}
where $\mathrm{KIN}(\boldsymbol{\xi})$ is a selected kinematic descriptor of the strain state, such as the deformation gradient $\boldsymbol{F} = \mathrm{GRAD}(\boldsymbol{\xi})$, the right Cauchy - Green deformation tensor $\boldsymbol{C} = \boldsymbol{F}^\intercal \boldsymbol{F}$, or the Green - Lagrange strain tensor $\boldsymbol{E} = \frac{1}{2}(\boldsymbol{C} - \boldsymbol{I})$, among others. For small strains solid mechanics, the methodology simplifies to
\begin{subequations} \label{eq:problem_solid_mechanics_inf}
\begin{align}
\boldsymbol{U}&=\mathsf{Y}(\bar{\boldsymbol{t}})\\ 
\boldsymbol{\sigma}&=\mathsf{H}(\mathrm{symgrad}(\boldsymbol{U}))\\
\boldsymbol{x}&=(\bar{\boldsymbol{t}},\boldsymbol{b})\\
\boldsymbol{y}&=\boldsymbol{U}\\
\boldsymbol{R}(\boldsymbol{U},\boldsymbol{\sigma},\bar{\boldsymbol{t}})&=(\mathrm{div}(\boldsymbol{\sigma}) - \rho \boldsymbol{b}; \boldsymbol{\sigma} \cdot \boldsymbol{n}- \bar{\boldsymbol{t}};\boldsymbol{U}-\bar{\boldsymbol{U}}). 
\end{align}
\end{subequations}

The appropriate structure and architecture of $\mathsf{H}$ and $\mathsf{Y}$ depend on the complexity of the material in hands, which is discussed later.

\subsubsection{Case study: geometry and external forces.}
\label{subsubsection:case}
\begin{figure}
    \centering %Para centrar la imagen
    \includegraphics[scale=0.6, trim=80 80 80 80]{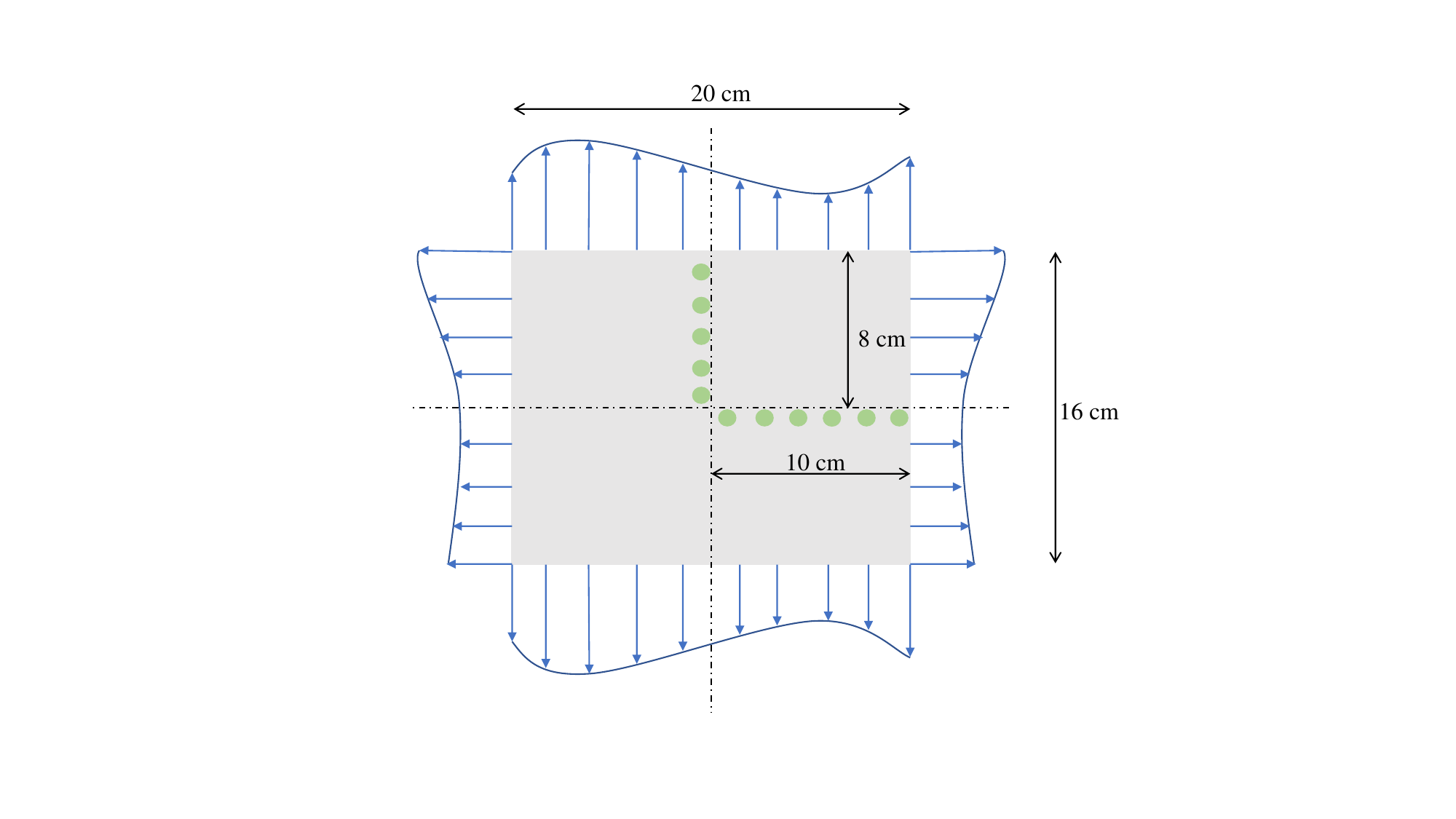}   %[Para su tamaño:anchura con respecto al ancho de la hoja]{Imágenes\Nombre_de_la_imagen}
    \caption{\textbf{Dimensions and representation of the non-uniform biaxial test on a 2D plate.} Load $p = p(s)$ acting perpendicularly to the contour is  arbitrary, provided that it is compatible with the symmetry of the problem. We locate the origin $(x,y)=(0,0)$ at the geometrical center of the plate.}
    \label{fig:plate}   %Nombre para referirse a la imagen
\end{figure}

For illustration purposes, the case study considered in this work consists in a non-uniform biaxial test on a rectangular plate of height $L_1 = 16$ cm and width $L_2 = 20$ cm, under plane stress. No volumetric loads are incorporated, that is, $\boldsymbol{b} = \boldsymbol{0}$. We impose a certain arbitrary compression load profile $p = p(s)$ (where $s$ is the coordinate along the right and top contour). To accelerate computations we consider a load profile that is symmetric with respect to the vertical and horizontal axis and acts perpendicularly to the plate contour, as shown in Figure \ref{fig:plate}. The symmetry of the problem allows therefore for the analysis of an equivalent problem by extracting the upper-right portion of the plate and applying the corresponding symmetry boundary conditions.

\subsection{Data and operators representation}
\label{subsection:set_up}

In this section we discuss how the different mathematical objects in our use case problem (scalar, vectorial and tensorial fields and operators) are represented. First, we describe how the different fields are encoded and related to the measured data. Then, we explain how the different operators involved are built. This includes both the known operators (equilibrium and compatibility) as well as the unknown relationships comprising the predictive and explanatory networks, $\mathsf{Y}$ and $\mathsf{H}$ respectively. Then, we discuss how physical constrains are hardwired into the ANN, so that the built PGNNIV is tailored towards the discovery of constitutive models that comply with the physics of the solid mechanics problem, constraining therefore the learning space and bypassing the parametrization of the constitutive law.

%In the following, we particularize the approach for infinitesimal solid mechanics for the sake of simplicity, even if the generalization to finite strains is straightforward as discussed in the previous section. Indeed, by comparing Eqs. (\ref{eq:problem_mechanics}) and (\ref{eq:problem_solid_mechanics_inf}) we note that the methodology only changes formally in the specification of the input variable of the explanatory network $\mathsf{H}$. In infinitesimal strains, the small deformation tensor $\boldsymbol{\varepsilon}$ is used, whereas in finite strains we use the Green - Lagrange deformation tensor $\boldsymbol{E} = \frac{1}{2}(\boldsymbol{F}^\intercal \boldsymbol{F} - \boldsymbol{I})$.

\subsubsection{Data structures}\label{subsubsection:data}

The data that contains the nodal and element-wise variables (that is, displacements and stresses/strains respectively) is stored in array structures. Now we introduce the notation that will be used for referring to a given tensor field, that is represented by an array $\mathtt{I}$ containing the information of the tensorial field itself. The different dimensions of the data are represented using the indexation $\mathtt{I}[I|J|K]$ where $I$ is a multi-index associated with the discretization of the problem, $J$ with the tensorial character and $K$ with the data instance. Thus, considering that we have a data-set of size $N$ and a discretization of size $n_x \times n_y$, the displacement field is represented by $\mathtt{U}[i,j|k|l]$ where $i=1,\ldots,n_x$, $j=1,\ldots,n_y$, and where $k=1,2$ (2D problem) and $l = 1\ldots,N$, so that $$\mathtt{U}[i,j|k|l] = u_k(x_i,y_j)$$ is the $k$ component of the displacement field evaluated at $(x_i,y_j)$ (that is, the node $(i,j)$) corresponding to the data $l$. Analogously, the strain and stress fields are represented respectively by $\mathtt{E}[i,j|k,l|m]$ and $\mathtt{S}[i,j|k,l|m]$ where $i=1,\ldots,n_x-1$, $j=1,\ldots,n_y-1$, $k,l=1,2$ and $m = 1\ldots,N$, such that, for instance, $$\mathtt{E}[i,j|k,l|m] = E_{kl}\left(\frac{1}{2}(x_i + x_{i+1}),\frac{1}{2}(y_j + y_{j+1})\right)$$ is the $k,l$ component of the strain tensor evaluated at the element $(i,j)$  corresponding to the data $m$.

Finally, the traction forces, $\bar{\boldsymbol{t}}^{\mathrm{top}}$ and $\bar{\boldsymbol{t}}^{\mathrm{right}}$, which are treated as the inputs of our problem (provided the volume forces are not considered), are represented by $\mathtt{t}^{\mathrm{top}}[i|j|k]$ where $i=1,\ldots,n_x$, $j=1,2$, and $k = 1\ldots,N$, so that $$\mathtt{t}^{\mathrm{top}}[i|j|k] = t^{\mathrm{top}}_j(x_i,L_1/2),$$ and $\mathtt{t}^{\mathrm{right}}[i|j|k]$ where $i=1,\ldots,n_y$, $j=1,2$, and $k = 1\ldots,N$, so that $$\mathtt{t}^{\mathrm{right}}[i|j|k] = t^{\mathrm{right}}_j(L_2/2,y_i),$$ both associated with the data instance $k$.

\subsubsection{Operator construction}
\label{subsubsection:operators}

In this section we specify the details to build the predictive and explanatory networks ($\mathsf{Y}$ and $\mathsf{H}$), as well as the constraint operator $\boldsymbol{R}$. The definition of the complexity of a  PGNNIV adapted to solid mechanics stems from the architecture of the predictive and explanatory networks, $\mathsf{Y}$ and $\mathsf{H}$, as they must be able to learn the non-linearities between variables $\bar{\boldsymbol{t}} \mapsto \boldsymbol{U}$ or $\bar{\boldsymbol{t}} \mapsto \boldsymbol{\xi} $ and  $\boldsymbol{E} \mapsto \boldsymbol{P}$ (or $\boldsymbol{\varepsilon} \mapsto \boldsymbol{\sigma}$), as explained in Section \ref{subsection:PGNNIV_in_CM}.

\paragraph{Predictive network.} The predictive network must be able to represent the data variability, so typically it has an autoencoder-like structure. Its complexity is therefore associated with the latent dimensionality and structure of the volumetric loads and boundary conditions. Although more sophisticated approaches coming from Manifold Learning theory are possible for analyzing data dimensionality and structure \cite{ma2012manifold}, this is not the main interest of this work and therefore is out of the scope of this particular study. Here we follow a much simpler approach, where we build an ANN that is sufficiently accurate when predicting the output $\boldsymbol{y}$ from an input $\boldsymbol{x}$.

Since we consider a biaxial quadratic load applied to the plate, the complexity of the network depends on the data variability. For non-uniform loads, the values of the elemental loads are the input of a DNN whose output are the nodal displacements. For the uniform load, we use a much simpler autoencoder-like DNN. It is important to note that a single, complex enough, autoencoder-like DNN would be able to represent the data variability even in the more complex scenario (the case with the largest latent space in the process of data generation). However, we have decided to use two different network architectures for illustrating this particular feature of PGNNIV: the predictive network is associated to data variability, rather than data nature. Therefore, we can adapt the network architecture to our problem characteristics, aiming either at a better network performance (avoiding overfitting) or at a lower computational cost.

In Appendix \ref{app:network_details}, the particular architecture of the two predictive networks is detailed, both for the non-uniform biaxial test and for the uniform biaxial test. Anyway, although the $\mathsf{Y}$ network architecture was handcrafted, it is expected that the more and variate data is available, the less relevant the hand-engineering of the network becomes.

%The implementation details of both $\mathsf{Y}$ and $\mathsf{H}$ networks architectures can be found in Appendix \ref{sec:hyper}.
\paragraph{Explanatory network.} As in this work we restrict ourselves to the elastic regime, the input variable is the given strain state at an arbitrary node $\mathtt{E}[i,j|k,l|m]$ ($\mathsf{E}$ represents the Cauchy deformation tensor for infinitesimal theory and the Green -  Lagrange deformation tensor for finite strains theory)  and the output variable is the associated stress state at that same element, $\mathtt{S}[i,j|k,l|m]$ (again, $\mathsf{S}$ represents the Cauchy stress tensor or the first Piola-Kirchhoff tensor depending on whether we are in the infinitesimal or finite strains theory). Note that under the homogeneity assumption (and postulating that the stress state depends only on the value of the deformation at the same point), the explanatory network is a nonlinear map $\mathbb{R}^3 \rightarrow \mathbb{R}^3$, due the symmetry of both tensors, that may be expressed symbolically as $$\mathtt{S}[i,j|\cdot ,\cdot |m] = \mathsf{H}\left(\mathtt{E}[i,j|\cdot ,\cdot |m]\right).$$ For non-local materials, given the described discretization, the explanatory network could be in principle a map $\mathbb{R}^{3n_xn_y} \rightarrow \mathbb{R}^{3n_xn_y}$ represented as $$\mathtt{S}[\cdot,\cdot|\cdot ,\cdot |m] = \mathsf{H}\left(\mathtt{E}[\cdot,\cdot|\cdot ,\cdot |m]\right).$$ 

\begin{figure}
    \centering %Para centrar la imagen
    \includegraphics[scale=0.5, trim=80 00 80 80]{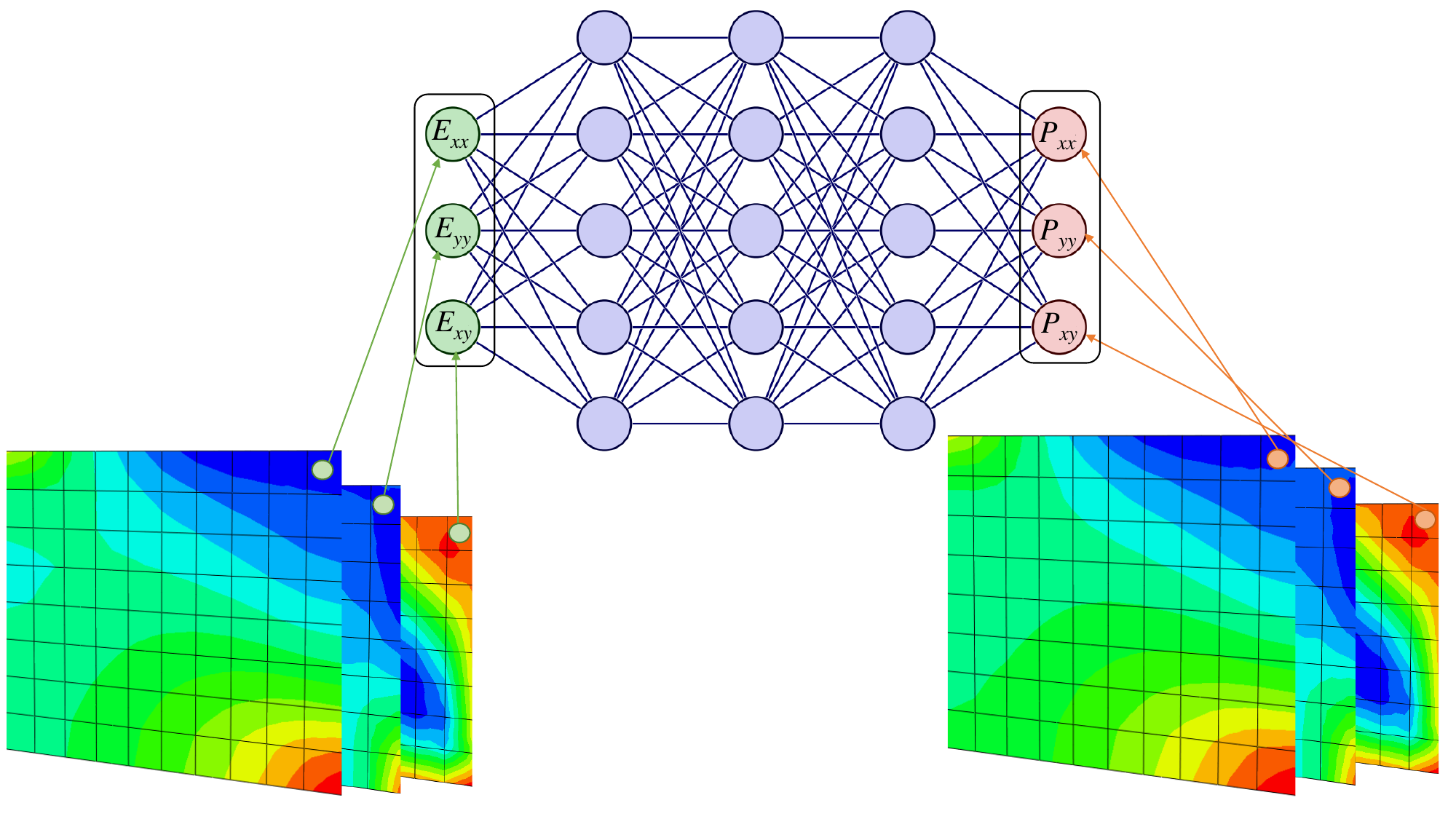}   %[Para su tamaño:anchura con respecto al ancho de la hoja]{Imágenes\Nombre_de_la_imagen}
    \caption{\textbf{Representation of the explanatory network for the 2D plane stress problem.} On the left, the strain field computed from the displacement field predicted by $\mathsf{Y}$ is represented on the plate. In an homogeneous material, the DNN is fed with the strain state on each element, and the correspondent stresses are obtained. The set of weights of this DNN moves across the elements of the plate and updates after each iteration of the optimization (see Figure \ref{fig:scheme} for the whole picture), acting as a 2D-convolutional filter. We call this type of architecture a moving Multilayer Perceptron (mMLP).}
    \label{fig:scheme2}   %Nombre para referirse a la imagen
\end{figure}

In particular, $\mathsf{H}$ has to be able to capture the highly nonlinear dependencies that may exist between variables. This is in theory possible thanks to the universal approximation theorem: by adding more internal layers (also known as hidden layers) to the DNN model, we can provide the network with the learning capability and complexity that a particular nonlinear constitutive law might require. The implementation of the explanatory network for homogeneous materials is efficiently implemented using a \emph{convolutional} filter to move across the domain element by element, but expanding the features in a higher dimensional spaces, as illustrated in Fig. \ref{fig:scheme2}. We call this type of architecture a moving Multilayer Perceptron (mMLP).

For the very particular case of linear elasticity, we can explicitly parameterize the explanatory network, $\mathsf{H}(\boldsymbol{\sigma}) = \boldsymbol{H}(\boldsymbol{\sigma};\boldsymbol{D})$ where $\boldsymbol{D}$ is the elastic tensor and $\sigma_{ij} = D_{ijkl}\varepsilon_{kl}$. Using Voigt convention, the tensor $\boldsymbol{D}$ is expressed, under plane stress conditions and in the most general case, as a $3\times 3$ matrix.
\begin{equation} \label{eq:D_anisotropic}
\boldsymbol{D} =\begin{pmatrix}
\ d_{11}  & d_{12} & d_{13}\\
\ d_{12}  & d_{22} & d_{23}\\
\ d_{13}  & d_{23} & d_{33}\\
\end{pmatrix},
\end{equation}
where $d_{ij}$ are free model parameters, whereas for an isotropic elastic material we have
\begin{equation} \label{eq:D_isotropic}
\boldsymbol{D} =\frac{E}{1-\nu^2}\begin{pmatrix}
\ 1  & \nu & 0\\
\ \nu   & 1 & 0\\
\ 0  & 0 & 1 - \nu\\
\end{pmatrix},
\end{equation}
where the elastic modulus $E$ and the Poisson ratio $\nu$ are the only free fitting parameters learned during the training process.

An analogous reasoning holds for more complex parametric dependencies. It is possible to express the explanatory network $\mathsf{H}$ as a parametric model relating the strain and the stress states, that is $$\mathtt{S}[i,j|\cdot ,\cdot |m] = \mathbf{H}\left(\mathtt{E}[i,j|\cdot ,\cdot |m];\boldsymbol{\Lambda}\right).$$ where $\boldsymbol{\Lambda}$ are some pre-defined fitting parameters that are learned during the training step. In particular, an homogeneous material is described as $$\mathtt{S}[i,j|\cdot ,\cdot |m] = \mathbf{H}\left(\mathtt{E}[i,j|\cdot ,\cdot |m];\boldsymbol{\Lambda}_{ij}\right).$$

In this work, this approach is illustrated with different types of materials ranging from the more simple case of a linear elastic material under infinitesimal strain theory to an hyperelastic Ogden material under finite strains theory.

\paragraph{Coupling the two networks using physical constraints.} The definition and subsequent formulation of the PGNNIV framework implies that the loss function includes a term proportional to the quadratic error between the predictions and true values of the output variable (minimization of the maximum likelihood of the data given the parameters) and other penalty terms related to some (physical) equations i.e. equilibrium constraints. Therefore the different terms involved are:
\begin{enumerate}
    \item Loss term associated with the measurement of the displacement field:
    \begin{equation}
 \mathrm{MSE} = \frac{1}{N}\sum_{i=1}^N||\bar{\boldsymbol{U}}^{(i)}-\mathsf{Y}(\boldsymbol{t}^{(i)})||^2, 
    \end{equation}
where $\bar{\boldsymbol{U}}^i$ is the observed displacement corresponding to sample $i$.
    \item Constraint associated with the equilibrium equation.
    \begin{equation} \label{eq:loss_eqd}
    \boldsymbol{\nabla}\cdot\boldsymbol{P} = \boldsymbol{0}, \quad \mathrm{or}  \quad \boldsymbol{\nabla}\cdot\boldsymbol{\sigma} = \boldsymbol{0}.
    \end{equation}
    \item Constraint associated with
 the compatibility in the domain.
    \begin{equation} \label{eq:loss_cd}
        \boldsymbol{E} - \frac{1}{2}\left(\boldsymbol{F}^\intercal \boldsymbol{F} - \boldsymbol{I}\right) = \boldsymbol{0}, \quad \mathrm{or}\quad \boldsymbol{\varepsilon}-\frac{1}{2}(\nabla\otimes \boldsymbol{U}+\boldsymbol{U}\otimes\nabla)=\boldsymbol{0}.
    \end{equation}
    \item Constraint associated with
 the equilibrium of the stresses in the boundary.
    \begin{equation} \label{eq:loss_eqb}
    \boldsymbol{P}\cdot\boldsymbol{N} -\boldsymbol{T} =\boldsymbol{0},\quad \mathrm{or} \quad \boldsymbol{\sigma}\cdot\boldsymbol{n} -\boldsymbol{t} =\boldsymbol{0}, \text{ in } \Gamma_{N}.
    \end{equation}
    \item Constraints associated with
 the compatibility of the displacements in the boundary.
   \begin{equation} \label{eq:loss_cb}
    U_x(x=0,y)=0, \quad U_y(x,y=0)=0.
    \end{equation}
\end{enumerate}

The global cost function (which turns out to be a \emph{virtual} physics-informed likelihood in a Bayesian formulation or, equivalently, a regularized cost function in the most common terminology) can be computed as a weighted sum of $\mathrm{MSE}$ and $\mathrm{PEN}$, with $\mathrm{PEN}$ referring to the physical terms, that is, Eqs. (\ref{eq:loss_eqd}), (\ref{eq:loss_cd}), (\ref{eq:loss_eqb}) and (\ref{eq:loss_cb}). As Eq. (\ref{eq:loss_cd}) may be expressed as an explicit relation between $\boldsymbol{E}$ (or $\boldsymbol{\varepsilon}$) and $\boldsymbol{U}$, it is directly embedded in the network architecture. Therefore, the loss is expressed as:
\begin{equation}
\mathrm{CF} = \mathrm{MSE} + \mathrm{PEN},
\label{fig:cf}
\end{equation}
where
\begin{equation}
\mathrm{MSE} = \frac{1}{N}\sum_{i=1}^N\left[p_1\|\bar{\boldsymbol{U}}^{(i)}-\mathsf{Y}\left(\bar{\boldsymbol{t}}^{(i)}\right)\|^2\right],
\label{fig:cf1}
\end{equation}
and
\begin{equation}
\mathrm{PEN} = \frac{1}{N}\sum_{i=1}^N\left[p_2\|\boldsymbol{\nabla}\cdot\boldsymbol{\sigma}^{(i)}\|^2 +
p_3\|\boldsymbol{\sigma}^{(i)}\cdot\boldsymbol{n}-\bar{\boldsymbol{t}}^{(i)}\|^2+
p_4\\\|U_x^{(i)}(x=0,y)+U_y^{(i)}(x,y=0)\|^2\right],
\label{fig:cf2}
\end{equation}
where with the superscript $(i)$ we refer to the $i$-th piece of data and $p_{j}, j=1,2,3,4$ are penalty coefficients that account for the relative importance of each term in the global CF (and may be seen as Lagrange multipliers that softly enforce the constrains). Recall that no penalty for the compatibility in the domain is included since $\boldsymbol{E} - \frac{1}{2}\left(\boldsymbol{F}^\intercal \boldsymbol{F} - \boldsymbol{I}\right)$ (or $\mathbf{\boldsymbol{\varepsilon}}-\frac{1}{2}\mathbf{(\nabla\otimes \boldsymbol{U}}+\boldsymbol{U}\otimes\nabla)$) is identically $\textbf{0}$.

The ANN minimization problem reads therefore:
\begin{equation}
\min_{\boldsymbol{W}}\mathrm{CF}(\mathcal{E};\boldsymbol{W}),
\label{min}
\end{equation}
where $\boldsymbol{W}$ are the network parameters and $\mathcal{E} = \{\bar{\boldsymbol{t}}^i,\bar{\boldsymbol{U}}^i| i=1,\cdots,N\}$ is a given training data-set. By minimizing this function (and assuring that not overfitting is observed by examining the predictions for test data) we will obtain predictions of displacement, stresses and strains. For simplicity, Algorithm \ref{algorithm1} details a stochastic gradient descent version of the optimization, even if in this work we always used the Adam optimizer. 

From the theoretical point of view, Eq. (\ref{min}) presents a complex constrained optimization problem that has been widely studied in the context of applied mathematics, i.e. Langrange multipliers. However, when NNs come into play along with PDEs, the optimization becomes more involved as the complex nature of the Pareto front, extensively studied in \cite{rohrhofer2021pareto} for Physics-Informed Neural Networks (PINNs), determines that the optimum is a state where an individual loss cannot be further decreased without increasing at least one of the others, and therefore the optimal set of weighting hyperparameters $p_i$ cannot be inferred in advance. This weighting hyperparameters $p_i$, commonly referred as penalties, arise in a natural way if they are regarded as real numbers scaling the covariance matrix of the variables' \emph{virtual} maximum likelihood probability distribution. This concept was introduced in \cite{kaltenbach2020incorporating} and \cite{kaltenbach2021physics} in the context of state-space particle dynamics.

\RestyleAlgo{ruled}

\begin{figure}[H]
    \centering %Para centrar la imagen
    \includegraphics[scale=0.5]{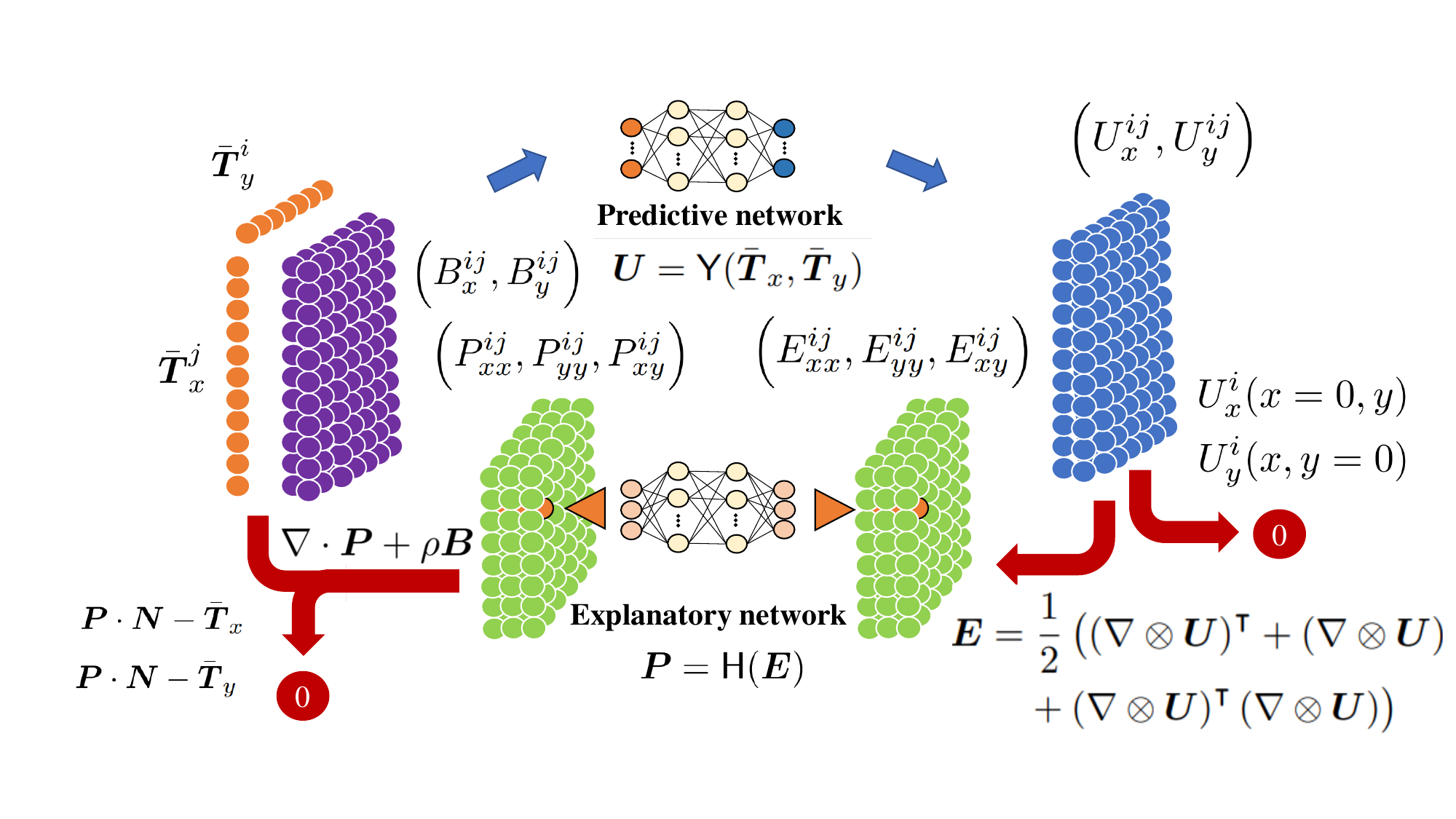}   %[Para su tamaño:anchura con respecto al ancho de la hoja]{Imágenes\Nombre_de_la_imagen}
    \caption{\textbf{Graphical representation of the designed 2D-planar stress PGNNIV.} All significant tensorial fiels of the problem are represented: the input variables (top and right tractions and volume forces, where the latter are assumed to be null so removed formally from the input), the output variables (displacement field at each nodal value), as well as the internal variables of the problem (stress and strain fields, represented in Voigt notation).}
    \label{fig:scheme}   %Nombre para referirse a la imagen
\end{figure}

%% This is needed if you want to add comments in
%% your algorithm with \Comment
\SetKwComment{Comment}{/* }{*/}
\begin{algorithm}[H]
\caption{PGNNIV learning algorithm}\label{alg:three}
\KwIn{PGNNIV architecture, batch size $n_b$, penalties $p_k$, $k=1,2,3,4$,  and number of iterations $M$;}
\KwData{external forces $\bar{\boldsymbol{t}}^{(i)}$, measured displacements $\bar{\boldsymbol{U}}^{(i)}$, $i=1,\ldots,N$\;}
 Initialization of PGNNIV parameters, $\boldsymbol{w} = \boldsymbol{w}^{0}$, $j = 0$\;
 \Repeat{$j = M$}{
    \For{$i=1,\ldots,n_b$}{
  $\mathtt{U}^{(i)} \gets \mathsf{Y}\left(\bar{\mathtt{T}}^{(i)};\boldsymbol{w}\right)$;\Comment{Predictive network}
  $\mathtt{E}^{(i)} = \mathsf{KIN}\left(\mathtt{U}^{(i)}\right)$;\Comment{Green-Lagrange or Cauchy strain tensor}
  $\mathtt{S}^{(i)} \gets \mathsf{H}\left(\mathtt{E}^{(i)};\boldsymbol{w}\right)$;\Comment{Explanatory network}
  }
  $\mathrm{MSE} = \frac{1}{n_b}\sum_{i=1}^{n_b}\left[p_1||\bar{\mathtt{U}}^{(i)}-\mathtt{U}^{(i)}||^2\right]$\;
  $\mathrm{PEN} = \frac{1}{n_b}\sum_{i=1}^{n_b}\left[p_2||\mathsf{DIV}(\mathtt{S}^{(i)})||^2 +p_3||\mathtt{S}^{(i)}\cdot\boldsymbol{N}-\bar{\mathtt{T}}^{(i)}||^2+
  p_4\left(\|\mathtt{U}_x^{(i)}(x=0,y)\|^2+\|\mathtt{U}_y^{(i)}(x,y=0)\|^2\right)\right]$\;
  $\mathrm{CF} = \mathrm{MSE} + \mathrm{PEN}$\;
  $\boldsymbol{w} \gets \boldsymbol{w}-\nabla_w \mathrm{CF}$; \Comment{Stochastic gradient descent step}
  $j \gets j+1$\;
 }
 \KwOut{Optimal parameters $\boldsymbol{w}^*=\boldsymbol{w}$ for $\mathsf{Y}$ and $\mathsf{H}$\;}
 \label{algorithm1}
\end{algorithm}
Figure \ref{fig:scheme} shows a graphical representation of the different structures involved (tensorial fields) and the links between them (known and unknown operators) for finite strains solid mechanics.

\subsubsection{Details about the discretization.}

The discretization of both space and time domains lies on the basis of numerical methods. In the particular case of solid mechanics under the hypotheses considered here, time discretization turns out not to be relevant for the overall computations, since loads are applied in a quasi-static way and the sole discretization of the geometry provides a very good approximation of how the continuum solid behaves.

Traditional FEMs follow a matrix-based approach to have algebraic systems whose solution is an approximate solution. By subdividing the whole domain into small parts (\emph{finite elements}), PDEs governing the physical phenomena occurring in the particular geometry can be approximated by means of computable functions to generate algebraic systems even for complex geometries. However, FEMs require exact knowledge of the properties of the material and are usually time-consuming. On the contrary, PGNNIVs require no information about the material properties since these are learned during the training process of the network, and the calculation time for the forward problem is reduced to seconds at prediction time in the online loop. 

The discrete nature of methods such as FEM, which subdivide space in small elements, very closely resembles that of PGNNIVs, which comprise a number of discrete units (neurons) to represent field variables. Moreover, commonly used differential operators in these methods are also subject of a suitable description in the PGNNIV framework using convolutional filters. 

For instance, it is possible to define the discrete gradient filter $\mathtt{GRAD}$ acting on the nodes for obtaining values on the elements or acting on the elements for obtaining values on the nodes. For example, if $\mathtt{W} = \mathsf{GRAD} \otimes \mathtt{U}$, then
\begin{align}
    \mathtt{W}[i,j|1,1|m] = \frac{1}{2h_x}\left(\Delta_x\mathtt{U}[i,j+1|1|m] + \Delta_x\mathtt{U}[i,j-1|1|m] \right), \nonumber \\
    \mathtt{W}[i,j|1,2|m] = \frac{1}{2h_y}\left(\Delta_y\mathtt{U}[i+1,j|1|m] + \Delta_y\mathtt{U}[i-1,j|1|m] \right), \nonumber \\
    \mathtt{W}[i,j|2,1|m] = \frac{1}{2h_x}\left(\Delta_x\mathtt{U}[i,j+1|2|m] + \Delta_x\mathtt{U}[i,j-1|2|m] \right), \nonumber \\
    \mathtt{W}[i,j|2,2|m] = \frac{1}{2h_y}\left(\Delta_y\mathtt{U}[i+1,j|2|m] + \Delta_y\mathtt{U}[i-1,j|2|m] \right), \nonumber
\end{align}
where $\Delta_x \mathtt{U}[i,\cdot|1|m] = \mathtt{U}[i+1,\cdot|1|m] - \mathtt{U}[i-1,\cdot|1|m]$ and $\Delta_y \mathtt{U}[\cdot ,j|1|m] = \mathtt{U}[\cdot,j+1|1|m] - \mathtt{U}[\cdot,j-1|1|m]$. Analogously, if $\mathtt{R} = \mathsf{GRAD} \cdot \mathtt{T}$

\begin{align}
    \mathtt{R}[i,j|1|m] = \frac{1}{2h_x}\left(\Delta_x\mathtt{T}[i,j+1|1,1|m] + \Delta_x\mathtt{T}[i,j-1|1,1|m] \right) + \frac{1}{2h_y}\left(\Delta_y\mathtt{T}[i+1,j|1,2|m] + \Delta_y\mathtt{T}[i-1,j|1,2|m] \right), \nonumber \\
    \mathtt{R}[i,j|2|m] = \frac{1}{2h_x}\left(\Delta_x\mathtt{T}[i,j+1|2,1|m] + \Delta_x\mathtt{T}[i,j-1|2,1|m] \right) + \frac{1}{2h_y}\left(\Delta_y\mathtt{T}[i+1,j|2,2|m] + \Delta_y\mathtt{T}[i-1,j|2,2|m] \right), \nonumber  
\end{align}

Now we can define the different discretized differentials. For instance the 2D-discretized symmetric gradient of a vector field $\boldsymbol{V}$ is $
\mathsf{SGRAD}(\boldsymbol{V})=\frac{1}{2}\left(\mathsf{GRAD} \otimes \boldsymbol{V} + \boldsymbol{V}\otimes\mathsf{GRAD}\right),$ where $\mathsf{GRAD}$ is the discrete gradient operator. Therefore, for large strain we have
\begin{equation}
\mathsf{E}=\frac{1}{2}\left(\mathsf{GRAD} \otimes \boldsymbol{U} + \boldsymbol{U}\otimes\mathsf{GRAD} + \left(\mathsf{GRAD} \otimes \boldsymbol{U} \right) \left(\boldsymbol{U}\otimes\mathsf{GRAD} \right) \right) \nonumber
\end{equation}
and for small strains
\begin{equation}
\mathsf{E}=\mathsf{SGRAD}(\boldsymbol{U}) = \frac{1}{2}\left(\mathsf{GRAD} \otimes \boldsymbol{U} + \boldsymbol{U}\otimes\mathsf{GRAD}\right). \nonumber
\end{equation}

Similarly, the 2D-discretized divergence of a tensor field $\boldsymbol{T}$ is defined as:
\begin{equation} 
\mathsf{DIV}(\boldsymbol{T})=\mathsf{GRAD} \cdot \boldsymbol{T}, \nonumber
\end{equation}
so we can express the equilibrium equation both in infinitesimal as well as finite strains theory as $\mathsf{DIV}(\boldsymbol{\sigma})=\boldsymbol{0}$ or $
\mathsf{DIV}(\boldsymbol{P})=\boldsymbol{0}$ respectively.

\subsection{Data generation and training process}
\label{subsec:dat}

In this work we generate synthetic data although the methodology is the same for real experimental data. Synthetic data generation is  more practical, as it allows for error validation by comparing to the real solutions, and inexpensive, given that an accurate numerical solution is available via FEM Analysis.

\paragraph{Small strains: linear, softening and hardening elastic materials.} For the creation of the linear, softening and hardening materials, we used Matlab, in co-simulation with Abaqus CAE/6.14-2. Matlab was used to automatically and iteratively generate an Abaqus input file that contained the geometry and load profiles. Once each numerical simulation is completed, the results are stored back into Matlab. For all the test cases, the geometry is the same, whereas the variability in the data-set is achieved by randomly changing the load profiles so that all the examples correspond with different experiments. The load profiles are parabolic and are generated for both the right and top contours.

We consider three elastic materials with different constitutive laws. On the one hand, we choose an isotropic linear elastic material with elastic modulus $E=1000$ Pa and Poisson's coefficient $\nu=0.3$. 
On the other hand, we considered two nonlinear materials, one with softening properties and another one with hardening properties. In Abaqus, they were modeled as plastic materials with no discharging effects caused by the removal of the load, and strain ranges confined within very small values, that allowing for the compliance of the nonlinear constitutive law with the infinitesimal strains hypothesis. A relation of the type $\sigma = K\varepsilon^n$ was used, with values of $K$ and $n$ specified in Table \ref{tab:soft_parameters}.
%As mentioned before, we use Abaqus CAE/6.14-2 to create the FEM-model and solve it iteratively for the described parabolic profiles acting on the plate. 

The data-set comprises $N=10^3$ FEM-simulations for the linear  material and $N=10^4$ FEM-simulations for the hardening and softening materials.

\begin{table}[H]
\centering
\begin{tabular}{|l|l|l|}
\hline
\textbf{Material} &  $K$ [Pa] &  $n$ [-]\\ \hline
Softening & $18.69$  & $0.45$  \\ \hline
Hardening & $1.869\times10^{12}$  & $3.5$  \\ \hline
\end{tabular}
\caption{Parameter values for the softening and hardening materials.}
\label{tab:soft_parameters}
\end{table}

\paragraph{Finite strains: Ogden-like hyperelastic material.} For the finite strains case, an incompressible Ogden-like hyperelastic material of order 3 is used for the data-set generation. The incompressible Ogden hyperelastic material of order $m$ is defined in terms of its strain-energy density function \cite{ogden1972large}:
\begin{equation}
    \Psi(\boldsymbol{C}) = \Psi(\lambda_1,\lambda_2,\lambda_3) = \sum_{i=p}^m\frac{\mu_p}{\alpha_p}\left(\lambda_1^{\alpha_p} + \lambda_2^{\alpha_p} +\lambda_3^{\alpha_p} - 3\right).
\end{equation}
In Table \ref{tab:ogden_parameters} we report the material parameters used for the data generation. We produce $N=10^4$ examples corresponding to uniform biaxial tests, where $\lambda_1,\lambda_2 \in [1;1.10]$. For an incompressible membrane under biaxial deformation, assuming a plane stress state, the solution of the problem using an Ogden's strain-energy function has analytical solution \cite{holzapfel2002nonlinear}. The displacement fields corresponding to uniform biaxial deformations are
\begin{equation}
U_x(x,y) = \lambda_1 x, \quad  U_y(x,y) = \lambda_2 y, \quad U_z(x,y) = \frac{1}{\lambda_1 \lambda_2}x, \nonumber
\end{equation}
so, the non-vanishing components of the Green-Lagrange deformation tensor, $\boldsymbol{E}$, are
\begin{equation}
    E_{xx} = \frac{1}{2}(\lambda_1^2 - 1), \quad E_{yy} = \frac{1}{2}(\lambda_2^2 - 1), \quad E_{zz} = \frac{1}{2}((\lambda_1 \lambda_2)^{-2} - 1).  \nonumber
\end{equation}
As we are assuming plane stress, that is $\sigma_{zz} = \sigma_{xz} =\sigma_{yz}= 0$, the non-vanishing components of the first order Piola-Kirchhoff stress tensor are
\begin{equation}
    P_{xx} = \frac{1}{\lambda_1}\sum_{k=1}^3\mu_k\left(\lambda_1^{\alpha_k} - (\lambda_1\lambda_2)^{-\alpha_k}\right), \quad P_{yy} = \frac{1}{\lambda_2}\sum_{k=1}^3\mu_k\left(\lambda_2^{\alpha_k} - (\lambda_1\lambda_2)^{-\alpha_k}\right).  \nonumber
\end{equation}

\begin{table}[H]
\centering
\begin{tabular}{|c|c|}
\hline
\textbf{Parameter} &  \textbf{Value} \\ \hline
$\mu_1$   & $281$ Pa  \\ \hline
$\mu_2$ & $-280$ Pa  \\ \hline
$\mu_3$ &  $0.31$ Pa  \\ \hline
$\alpha_1$   & $1.66$           \\ \hline
$\alpha_2$ & $ 1.61$ \\ \hline
$\alpha_3$ &  $38.28$  \\ \hline
\end{tabular}
\caption{Parameter values for the Ogden hyperelastic material.}
\label{tab:ogden_parameters}
\end{table}

\paragraph{Training process.} For the evaluation of the methodology, we have trained four PGNNIVs corresponding to four cases:
\begin{itemize}
    \item Linear material with parametric explanatory network.
    \item Linear, softening and hardening materials with non-parametric explanatory network.
    \item Ogden-like material with parametric explanatory network.
    \item Ogden-like material with non-parametric explanatory network.
\end{itemize}
For all the data-sets considered, there is a number of hyperparameters that have been tuned for obtaining the proceeding results with the different networks, namely the learning rate $\beta$ and the four penalty coefficients $p_i$, $i=1,\ldots,4.$. The specific values of these hyperparameters are reported together with the different network topologies in Appendix \ref{sec:hyper}.

%%%%%%%%%%%%%%%%%%%%%%%%%%%%%%%%%%%%%%%%%%%%%%%%%%%%%%%%%%%%%%%%%%%%%%%%%

\section{Results}\label{results_nonlinear}

When used as forward solvers, PGNNIVs can either predict measurable variables if force-displacement data is available, for example, through Digital Image Correlation (DIC) techniques, or explain the internal state of the solid if this is needed for a certain application. In this section, we validate the performance of PGNNIVs acting as a forward-solver against standard FEM solutions for the plate using the different materials described in Section \ref{subsec:dat}, and also as a method for constitutive equation discovery.

\subsection{Predictive capacity} \label{subsec:prediction}

\subsubsection{Infinitesimal strains case}
We first evaluate the prediction capacity of PGNNIVs for the different tested materials under random parabolic loads. For a quantitative evaluation of the predictive capacity of the PGNNIV, we define the Relative Error ($\mathrm{RE}$) of an array  field $\mathtt{I}$ as:
\begin{equation}
    \mathrm{RE}(\mathtt{I})=\frac{\sum_{I,J,K}\left(\hat{\mathtt{I}}[I|J|K] - \mathtt{I}[I|J|K]\right)^2}{\sum_{I,J,K}\mathtt{I}[I|J|K]^2},
    \label{relative_error_x_1}
\end{equation}
where $\hat{\mathtt{I}}$ is the predicted value and $\mathtt{I}$ the value obtained using FEM.
For instance, for the displacement field $\boldsymbol{U}$ represented by the array $\mathtt{U}$:
\begin{equation}
    \mathrm{RE}(\mathtt{U})=\sqrt{\frac{\sum_{i=1}^{n_x}\sum_{j=1}^{n_y}\sum_{k=1}^2\sum_{l=1}^N\left(\hat{\mathtt{U}}[i,j|k|l] - \mathtt{U}[i,j|k|l]\right)^2}{\sum_{i=1}^{n_x}\sum_{j=1}^{n_y}\sum_{k=1}^2\sum_{l=1}^N\mathtt{U}[i,j|k|l]^2}},
    \label{relative_error_x_2}
\end{equation}
or using standard field notation:
\begin{equation}
\mathrm{RE}(\boldsymbol{U}) = \sqrt{\frac{\sum_{i=1}^{n_x}\sum_{j=1}^{n_y}\sum_{k=1}^2\sum_{l=1}^N\left(\hat{U}^{(l)}_k(x_i,y_j) - U^{(l)}_k(x_i,y_j)\right)^2}{\sum_{i=1}^{n_x}\sum_{j=1}^{n_y}\sum_{k=1}^2\sum_{l=1}^N(U^{(l)}_k(x_i,y_j))^2}},
\end{equation}
where $U_k^{(l)}(x,y)$ represents the value of $U_k(x,y)$ corresponding to the data sample $l$.

Table \ref{tab:results_pre_1} shows the Relative Error (RE) obtained by the PGNNIV for the softening, hardening and linear materials. We observe predictive errors below $10\%$ for all the cases.

\begin{table}[H]
\centering
\begin{tabular}{|l|l|l|l|l|l|l|}
\hline
\textbf{Type of material} & $\mathrm{RE}(\boldsymbol{U})$  (\%) & $\mathrm{RE}(\boldsymbol{\sigma})$  (\%)   & $\mathrm{RE}(\boldsymbol{\varepsilon})$  (\%)                             \\ \hline
Linear   & 0.53 & 3.21  & 3.61         \\ \hline
Softening & 0.92 & 2.07 & 4.55 \\ \hline
Hardening  &  1.17 & 4.48 & 5.51 \\ \hline
%Ogden & 26 & 7.0 & 26 \\ \hline
\end{tabular}
\caption{PGNNIV predictive errors for the linear, softening and hardening materials.}
\label{tab:results_pre_1}
\end{table}

\subsubsection{Finite strains case}

We evaluate now the Ogden material under finite strains. Table \ref{tab:results_pre_2} shows the Relative Error (RE) obtained by the PGNNIV for the Ogden material, resulting in that case in errors below $2\%$ (due to the lower data-set variability).

\begin{table}[H]
\centering
\begin{tabular}{|l|l|l|l|l|l|l|}
\hline
& $\mathrm{RE}(\boldsymbol{U})$  (\%) & $\mathrm{RE}(\boldsymbol{P})$  (\%)   & $\mathrm{RE}(\boldsymbol{E})$  (\%)                             \\ \hline

Ogden material &  0.34 & 1.56 & 0.37 \\ \hline
%Ogden & 26 & 7.0 & 26 \\ \hline
\end{tabular}
\caption{PGNNIV predictive errors for the hyperelastic material.}
\label{tab:results_pre_2}
\end{table}

\subsection{Explanatory capacity} \label{subsec:explanation}

The explanatory capacity of the method is evaluated here in two levels: parametric identification and state model unveiling. 

\subsubsection{Infinitesimal strains}

\paragraph{Parameter identification.} First, we train and evaluate a linearized version of the PGNNIV (where $\mathsf{Y}$ and $\mathsf{H}$ are replaced with weight matrices with no activation functions) with the data-set corresponding to the isotropic linear elastic material parametrized with elastic modulus $E$ and Poisson ratio $\nu$. It is possible to evaluate the relative error when learning an elastic tensor $\boldsymbol{D}$ by means of
\begin{equation} \label{eq:error_r_matrix}
\epsilon_r(\boldsymbol{D}) = \frac{\|\hat{\boldsymbol{D}} - \boldsymbol{D}\|_\mathrm{F}}{\|\boldsymbol{D}\|_\mathrm{F}},
\end{equation}
where $\hat{\boldsymbol{D}}$ is the predicted value, $\boldsymbol{D}$ the real one, and  $\|\cdot\|_\mathrm{F}$ depicts the Fröbenius norm, $\|\boldsymbol{A}\|_F = \sqrt{\boldsymbol{A} : \boldsymbol{A}}$. For assessing the identifiability of the structural parameters , it is possible to evaluate the prediction of the model parameters associated with Eqs. (\ref{eq:D_anisotropic}) and (\ref{eq:D_isotropic}) by computing the relative errors for a given parameter $\lambda$, that are defined as:
\begin{equation} \label{eq:error_r_lambda}
\epsilon_r(\lambda) = \frac{|\hat{\lambda}-\lambda|}{\lambda},
\end{equation}
where $\hat{\lambda}$ and $\lambda$ are the predicted and real values respectively. The learned anisotropic and isotropic tensors, up to a precision of $1$ Pa are:

$$
\boldsymbol{D}_\mathrm{aniso} = 
\begin{pmatrix}
1099  & 330 & 0\\
330  & 1099 & 0\\
0  & 0 & 511\\
\end{pmatrix}
, \quad 
\boldsymbol{D}_\mathrm{iso} = 
\begin{pmatrix}
1098  & 331 & 0\\
331  & 1098 & 0\\
0  & 0 & 767\\
\end{pmatrix}.
$$

The errors of the elastic tensor when using the anisotropic or the isotropic elastic material model are respectively
$\varepsilon_r(\boldsymbol{D}_\mathrm{aniso}) = 14.4\%$ and $\varepsilon_r(\boldsymbol{D}_\mathrm{iso}) = 0.2\%$, i.e when the assumed hypotheses are true, the explanatory capacity of the network increases.  Notwithstanding, the general anisotropic model has a certain explanatory capacity, as we may detect the supplementary structural symmetries in the resulting elastic tensor, that is, $d_{13},d_{23} \ll d_{11}, d_{12},d_{22},d_{33}$, $d_{11} \simeq d_{22}$. The last symmetry condition $d_{33} = d_{11} - d_{12}$ is not fulfilled, as the data-set is not very rich in large pure shear-stress states, so the model is not able to detect this symmetry.

Moving to specific structural parameter identification,  Table \ref{tab:parameter_identification} describes how the model is able to predict accurately the different elastic parameters. As commented before, the worst prediction is observed for the anisotropic model and the parameter that correlates shear stresses and strains. If, using the anisotropic model, we were interested in finding the value of the isotropic model parameters $E$ and $\nu$, it is possible to compute $\nu$ using the relations
$$\nu = \frac{d_{12}}{d_{11}} = \frac{d_{12}}{d_{22}},$$ and then to compute $E$ using
$$ \quad E = d_{11}(1-\nu^2) = d_{22}(1-\nu^2) = d_{12}\frac{1-\nu^2}{\nu} = d_{33}(1+\nu) .$$ Of course, we obtain a different accuracy for each of these expressions. Using the anisotropic model, we obtain values of $0.1\%$ and $0.1\%$ for $\epsilon_r(\nu)$ and values of $0.04\%$, $0.04\%$, $0.4\%$ and $34\%$ for $\epsilon_r(E)$, in agreement with the previous observations.

\begin{table}[H]
\centering
\begin{tabular}{|c|c|} \hline
\textbf{Parameter}, $\lambda$ & \textbf{Relative error}, $\epsilon_r(\lambda)$ (\%) \\ \hline
Anisotropic model     &    \\ \hline
$d_{11}$    & $0.02$\\
$d_{12}$    &  $0.08$\\
$d_{13}$    &  $\infty$\\
$d_{22}$    &  $0.02$\\
$d_{23}$    & $\infty$\\
$d_{33}$    & $33.59$\\ \hline \hline
Isotropic model    & \\ \hline
$E$         & $0.20$\\ \hline
$\nu$       & $0.52$\\ \hline
\end{tabular}
\caption{Predicted and real values of the model parameters.}
\label{tab:parameter_identification}
\end{table}

\begin{figure}[htbp]
     \centering %Para centrar la imagen
     \begin{subfigure}[t]{0.495\linewidth}
         \centering %Para centrar la imagen
        \includegraphics[scale=0.28]{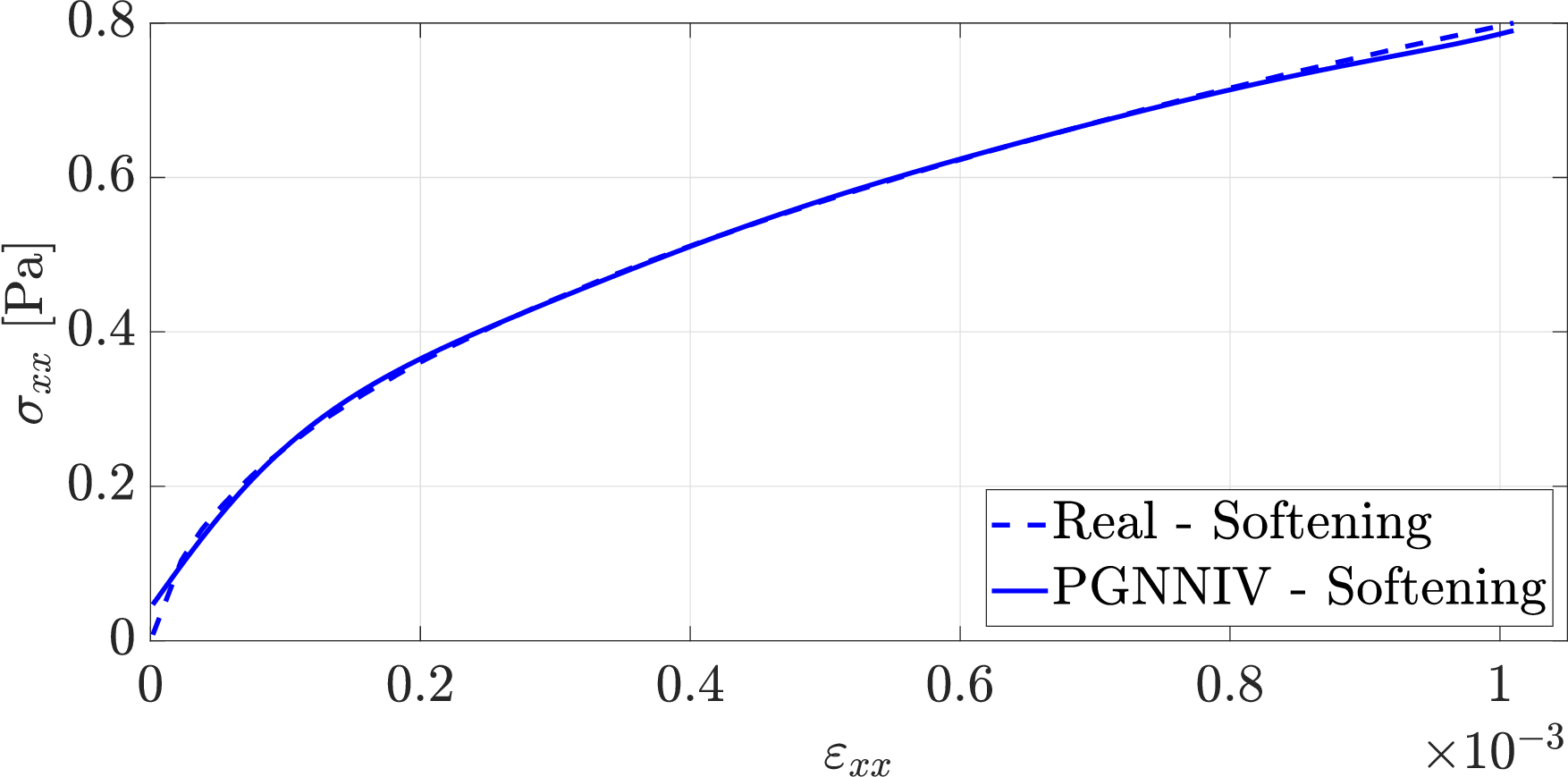}   %[Para su tamaño:anchura con respecto al ancho de la hoja]{Imágenes\Nombre_de_la_imagen}
        \caption{Softening material.}
        \label{soft1}
     \end{subfigure}
     \hfill
     \begin{subfigure}[t]{0.495\linewidth}
        \centering %Para centrar la imagen
        \includegraphics[scale=0.28]{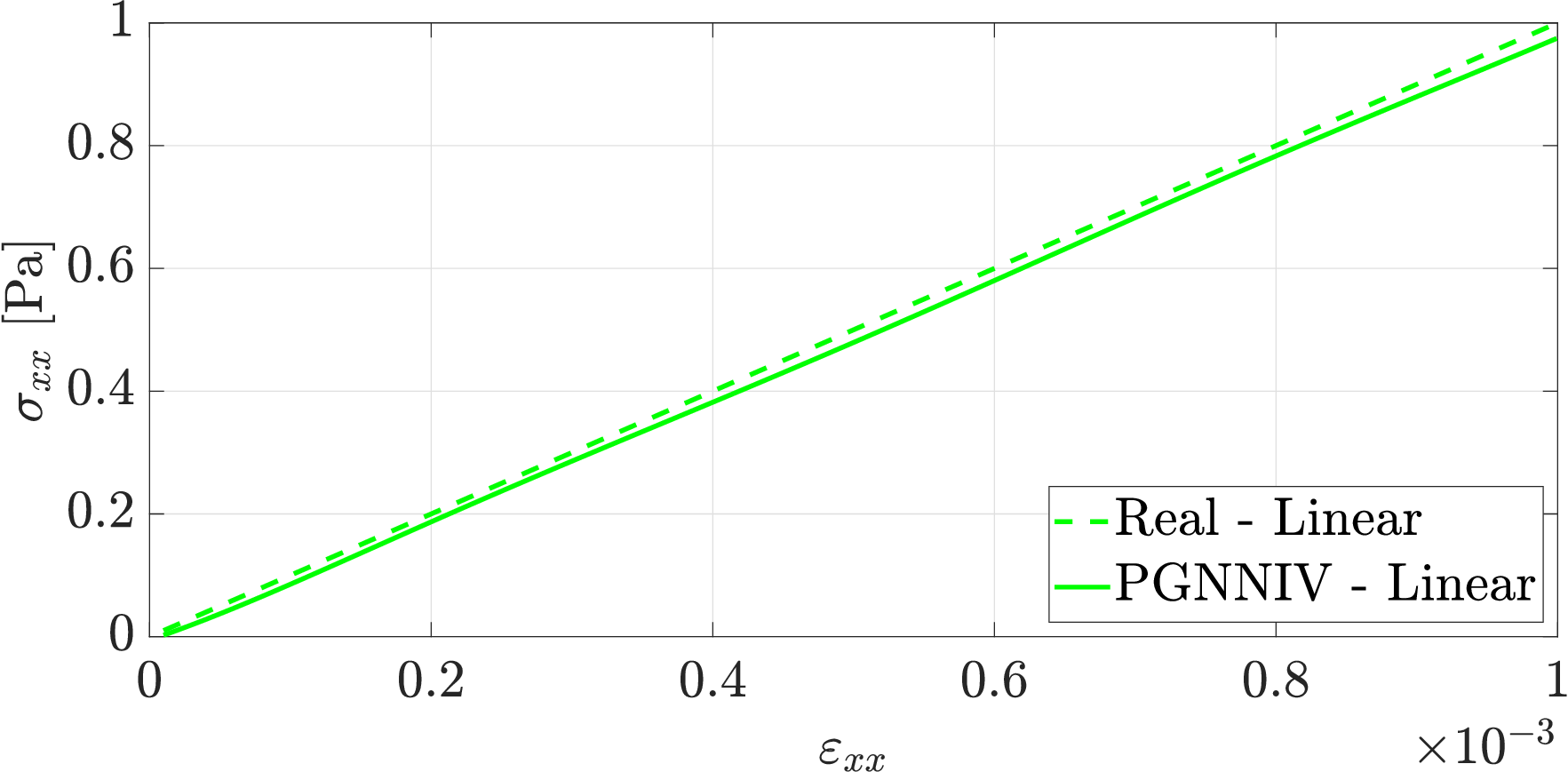}   %[Para su tamaño:anchura con respecto al ancho de la hoja]{Imágenes\Nombre_de_la_imagen}
        \caption{Linear material.}
        \label{lin1}
     \end{subfigure}
     \begin{subfigure}[t]{0.495\linewidth}
         \centering %Para centrar la imagen
        \includegraphics[scale=0.28]{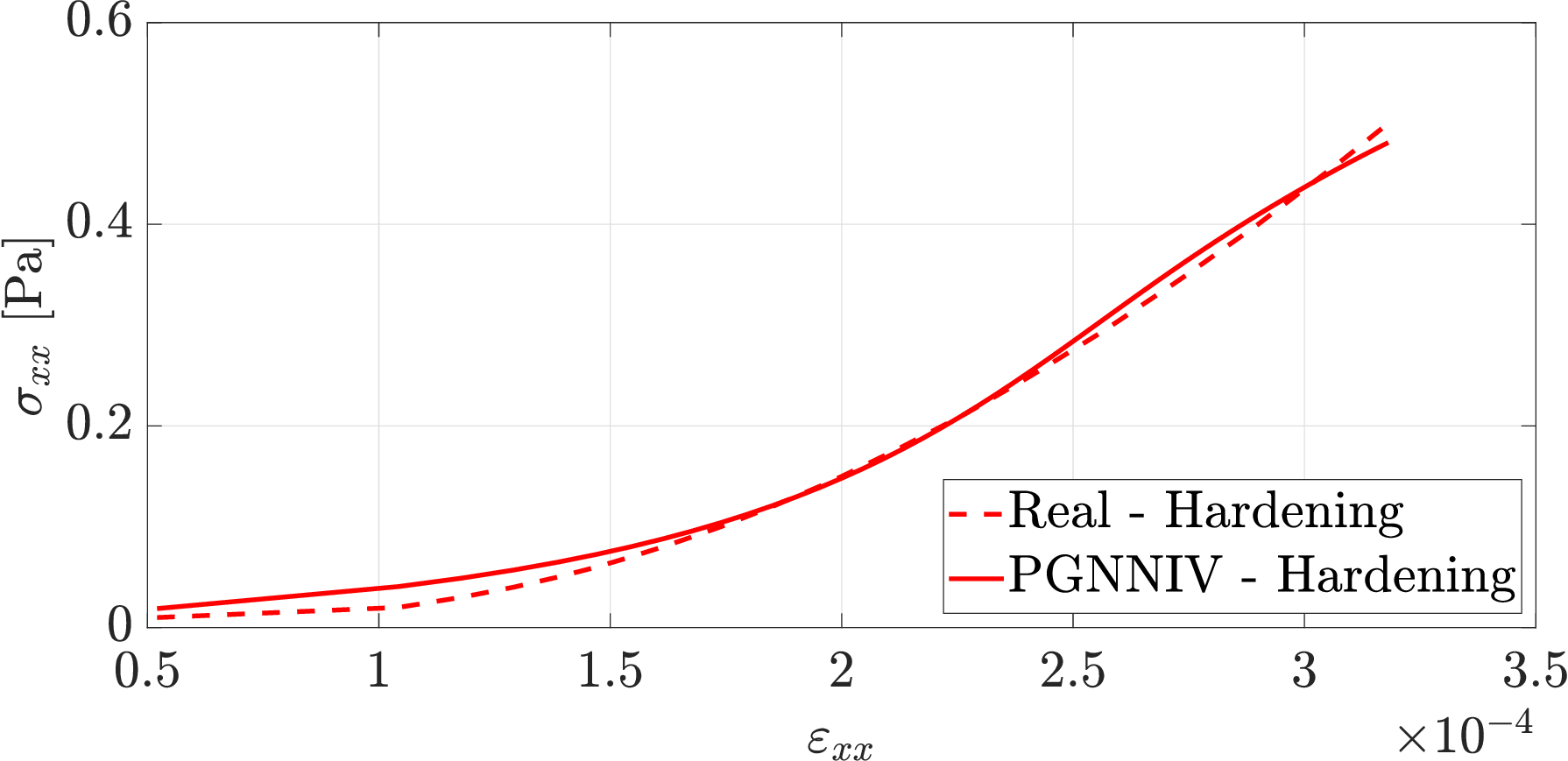}   %[Para su tamaño:anchura con respecto al ancho de la hoja]{Imágenes\Nombre_de_la_imagen}
        \caption{Hardening material.}
        \label{hard1}
     \end{subfigure}
     \caption{\textbf{PGNNIV prediction versus FEM solution of the uniaxial test curve for the different data-sets.}  We observe good agreement between FEM solution (continuous line) and PGNNIV prediction (dashed line), for the softening (a), linear (b) and hardening (c) materials.}
    \label{curve_soft_global}   %Nombre para referirse a la imagen
\end{figure}

\paragraph{State model discovery.} While the predictive capacity of PGNNIVs does not necessarily surpass that of a classical (unconstrained) NN, the significant improvement is visible when assessing the explanatory capacity of the PGNNIV, which can be evaluated by its ability to learn the material constitutive law. We perform a virtual uniaxial test using the explanatory network, which corresponds to the functional representation of $\sigma_{xx}= \mathsf{H}_1(\varepsilon,0,0)$ for $\varepsilon\in [\varepsilon_\mathrm{min};\varepsilon_\mathrm{max}]$ and we compare the PGNNIV predictions with a virtual uniaxial test produced with FEM, as described in Section \ref{subsec:dat}. Results are shown in Figure \ref{curve_soft_global}.

The explanatory error is quantified as the normalized area confined between the real uniaxial test curve and the PGNNIV-predicted one in Figure \ref{curve_soft_global}. It is expressed as:
\begin{equation}
   \mathrm{RE}(\mathsf{H})=\sqrt{\frac{\int_{\varepsilon_{\mathrm{min}}}^{\varepsilon_{\mathrm{max}}} (\hat{\sigma}_{xx}(\varepsilon)-\sigma_{xx}(\varepsilon))^2\,\mathrm{d}\varepsilon}{\int_{\varepsilon_{\mathrm{min}}}^{\varepsilon_{\mathrm{max}}} \sigma_{xx}^2(\varepsilon)\,\mathrm{d}\varepsilon}},
\label{eq:sigma_under}
\end{equation}
where $\hat{\sigma}_{xx}$ and $\sigma_{xx}$ are the predicted and FEM stresses respectively, which result from strains $\varepsilon \in [\varepsilon_{min}; \varepsilon_{max}]$.

\begin{table}[H]
\centering
\begin{tabular}{|l|l|}
\hline
\textbf{Type of material} & $\mathrm{RE}(\mathsf{H})$ (\%) \\ \hline
Linear    & 3.02  \\ \hline  
Softening & 0.96 \\ \hline 
Hardening & 5.92 \\ \hline 
\end{tabular}
\caption{Explanatory errors for the $\mathsf{H}$ model subjected to uniform uniaxial test.}
\label{tab:exp_errors_small}
\end{table}

\subsubsection{Finite strains}

We explore now the explanatory capacity for the finite strains case. If $\lambda \in [\lambda_\mathrm{min};\lambda_\mathrm{max}]$ is the longitudinal stretch, $\lambda_1$, the relative explanatory errors for a given transversal stretch $\lambda_2$ are defined as:
\begin{align}
   \mathrm{RE}_{xx}(\mathsf{H};\lambda_2)&=\sqrt{\frac{\int_{\lambda_{\mathrm{min}}}^{\lambda_{\mathrm{max}}} (\hat{P}_{xx}(\lambda,\lambda_2)-P_{xx}(\lambda,\lambda_2))^2\,\mathrm{d}\lambda}{\int_{\lambda_{\mathrm{min}}}^{\lambda_{\mathrm{max}}} P_{xx}^2(\lambda,\lambda_2)\,\mathrm{d}\lambda}},  \\
   \mathrm{RE}_{yy}(\mathsf{H};\lambda_2)&=\sqrt{\frac{\int_{\lambda_{\mathrm{min}}}^{\lambda_{\mathrm{max}}} (\hat{P}_{yy}(\lambda,\lambda_2)-P_{yy}(\lambda,\lambda_2))^2\,\mathrm{d}\lambda}{\int_{\lambda_{\mathrm{min}}}^{\lambda_{\mathrm{max}}} P_{yy}^2(\lambda,\lambda_2)\,\mathrm{d}\lambda}}. 
\label{sigma_under}
\end{align}

Note that, as the roles of $x$ and $y$ are symmetrical in the considered biaxial test (which is uniform, meaning that $P_{xx}$ on the top contour has the same vale as $P_{yy}$ on the right contour), the indicated errors are sufficient for illustrating the explanatory capacity of the method. For structural parameter identification, the formula used for error quantification is Eq. (\ref{eq:error_r_lambda}).

\paragraph{Parameter identification.} As explained previously, we first explicitly state the parametric shape of the constitutive equation, that is, we prescribe the material to be Ogden-like. Under these assumptions and for the uniform biaxial test considered, the constitutive relation writes
\begin{align}
    P_{xx} &= \frac{1}{\sqrt{2E_{xx} + 1}}\sum_{p=1}^3\mu_k\left[\left(2E_{xx} + 1\right)^{\alpha_k/2} - \left((2E_{xx}+1)(2E_{yy}+1)\right)^{-\alpha_k/2}\right], \nonumber \\ 
    P_{yy} &= \frac{1}{\sqrt{2E_{yy} + 1}}\sum_{p=1}^3\mu_k\left[\left(2E_{yy} + 1\right)^{\alpha_k/2} - \left((2E_{xx}+1)(2E_{yy}+1)\right)^{-\alpha_k/2}\right]. \nonumber \\ 
\end{align}

\begin{figure}[H]
 \centering
 \begin{subfigure}[t]{0.495\linewidth}
    \centering %Para centrar la imagen
    \includegraphics[scale=0.43, trim=80 00 00 00]{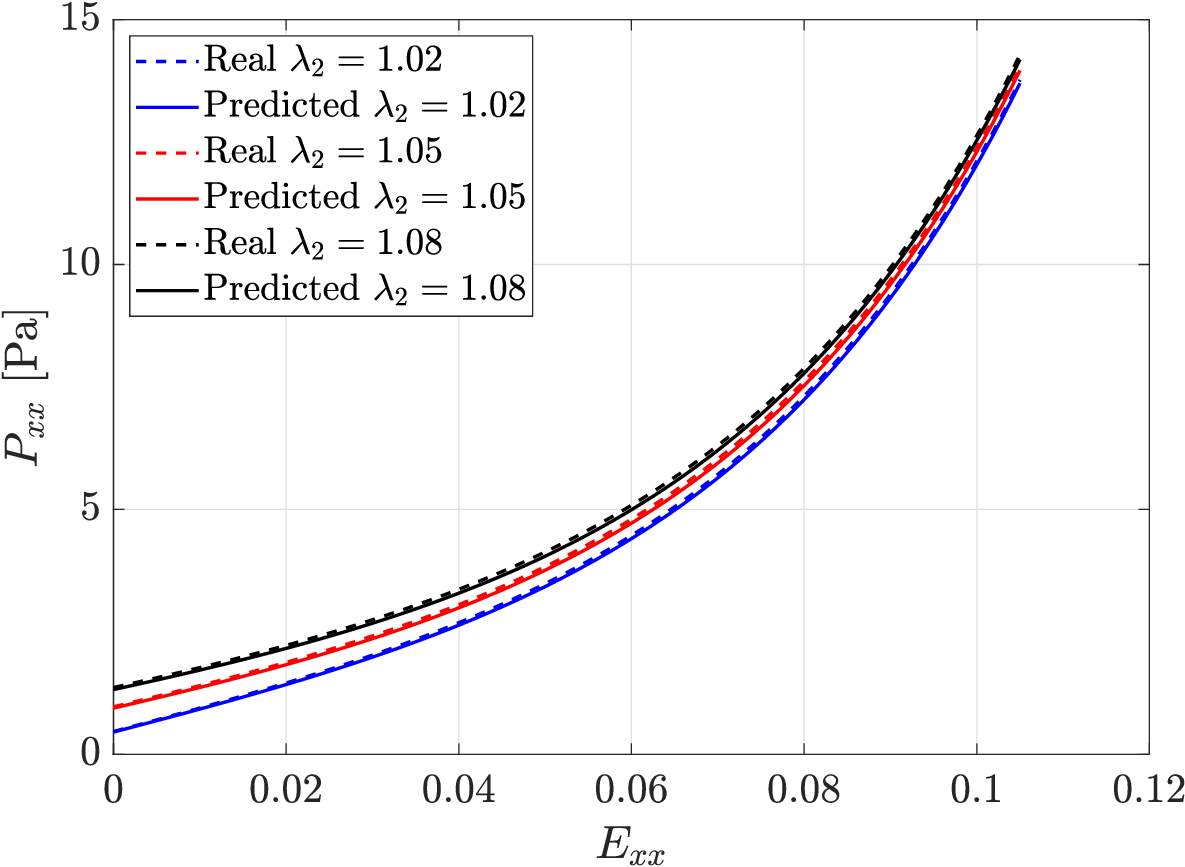}   %[Para su tamaño:anchura con respecto al ancho de la hoja]{Imágenes\Nombre_de_la_imagen}
    \caption{Component $xx$.}
    \label{og1p}   %Nombre para referirse a la imagen
 \end{subfigure}
 \hfill
 \begin{subfigure}[t]{0.495\linewidth}
    \centering %Para centrar la imagen
    \includegraphics[scale=0.43]{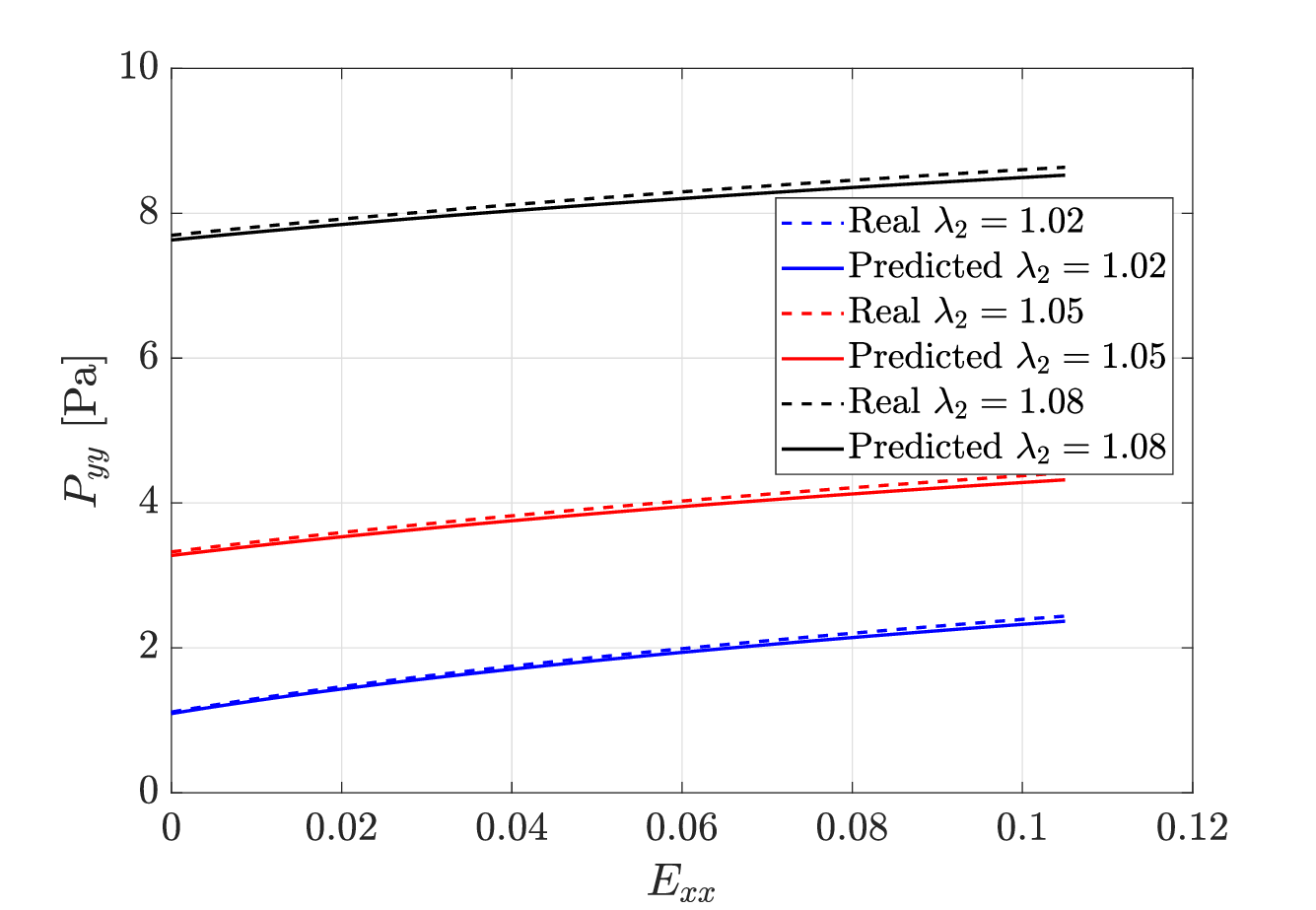}   %[Para su tamaño:anchura con respecto al ancho de la hoja]{Imágenes\Nombre_de_la_imagen}
    \caption{Component $yy$.}
    \label{og2p}   %Nombre para referirse a la imagen
 \end{subfigure}
\caption{\textbf{Parametric PGNNIV prediction versus analytic solution of the uniform biaxial test curve for the Ogden material.} We observe good agreement between analytical solution (continuous line) and PGNNIV prediction (dashed line), for the different values of $\lambda_2$. This indicates that the network has a good explicability capacity even though some superflous model parameters are not accurately fitted.}
\label{fig:parameter_identification_ogden}
\end{figure}

Therefore, the parameters $\alpha_k$, $\mu_k$, $k=1,2,3$ are, in principle, the ones that ought to be learned by the explanatory network. We obtain values for the parameters of $\mu_1 = 276\, \mathrm{Pa}$, $\mu_2 = -277\, \mathrm{Pa}$, $\mu_3 = 0.31 \, \mathrm{Pa}$, $\alpha_1 = 1.53$, $\alpha_2 = 1.47$ and $\alpha_3 = 38.32$. The relative errors for the different parameters are shown in Table \ref{tab:parameter_identification_ogden}. It is important to note that there are some parameters that are more accurately predicted than others, i.e. they are superfluous. This fact relies on the capacity of each of the parameters for explaining the material response, as observed in Fig. \ref{fig:parameter_identification_ogden}, where we compare the theoretical constitutive relation with the one obtained using the learned parameters. The explanatory errors are reported in Table \ref{tab:exp_errors_large}, adding evidence of the explanatory power of the method despite the discrepancies in some parameters.

\begin{table}[H]
\centering
\begin{tabular}{|c|c|} \hline
\textbf{Parameter}, $\lambda$ & \textbf{Relative error}, $\epsilon_r(\lambda)$ (\%) \\ \hline
$\mu_1$    & $1.7$\\
$\mu_2$    &  $-1.1$\\
$\mu_3$    &  $0.5$\\
$\alpha_1$    &  $7.7$\\
$\alpha_2$    & $8.6$\\
$\alpha_3$    & $0.1$\\ \hline
\end{tabular}
\caption{Predicted and real values of the model parameters for the Ogden material.}
\label{tab:parameter_identification_ogden}
\end{table}

\paragraph{State model discovery.} We now evaluate the model discovered by the PGNNIV with a virtual biaxial test using the explanatory network. The functional representation is now $(P_{xx},P_{yy},P_{xy})= \mathsf{H}(E_{xx},E_{yy},0)$ for $E_{xx} \in [E_\mathrm{min};E_\mathrm{max}]$ and we compare the PGNNIV predictions with the model used for the data generation. 

\begin{figure}[H]
 \centering
 \begin{subfigure}[t]{0.495\linewidth}
     \centering %Para centrar la imagen
    \includegraphics[scale=0.43, trim=80 00 00 00]{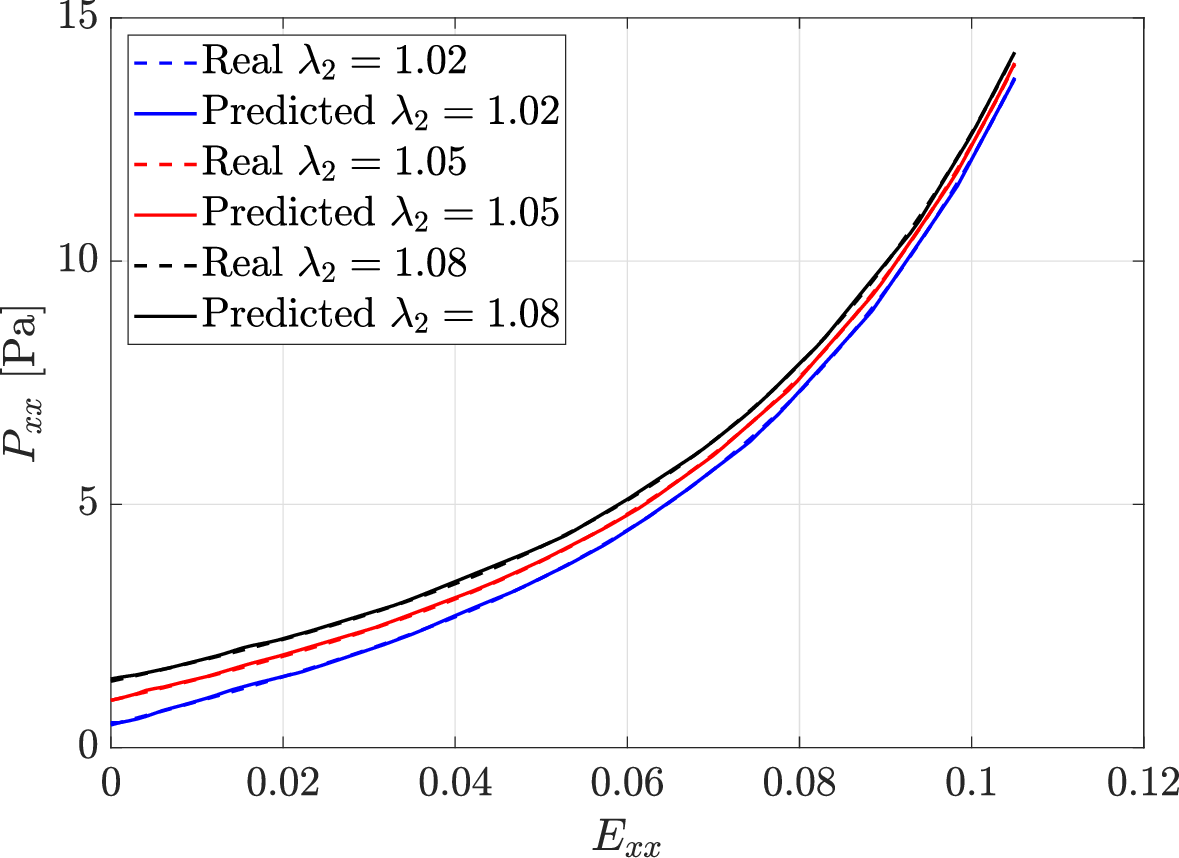}   %[Para su tamaño:anchura con respecto al ancho de la hoja]{Imágenes\Nombre_de_la_imagen}
    \caption{Component $xx$.}
    \label{og1}   %Nombre para referirse a la imagen
 \end{subfigure}
 \hfill
 \begin{subfigure}[t]{0.495\linewidth}
    \centering %Para centrar la imagen
    \includegraphics[scale=0.43]{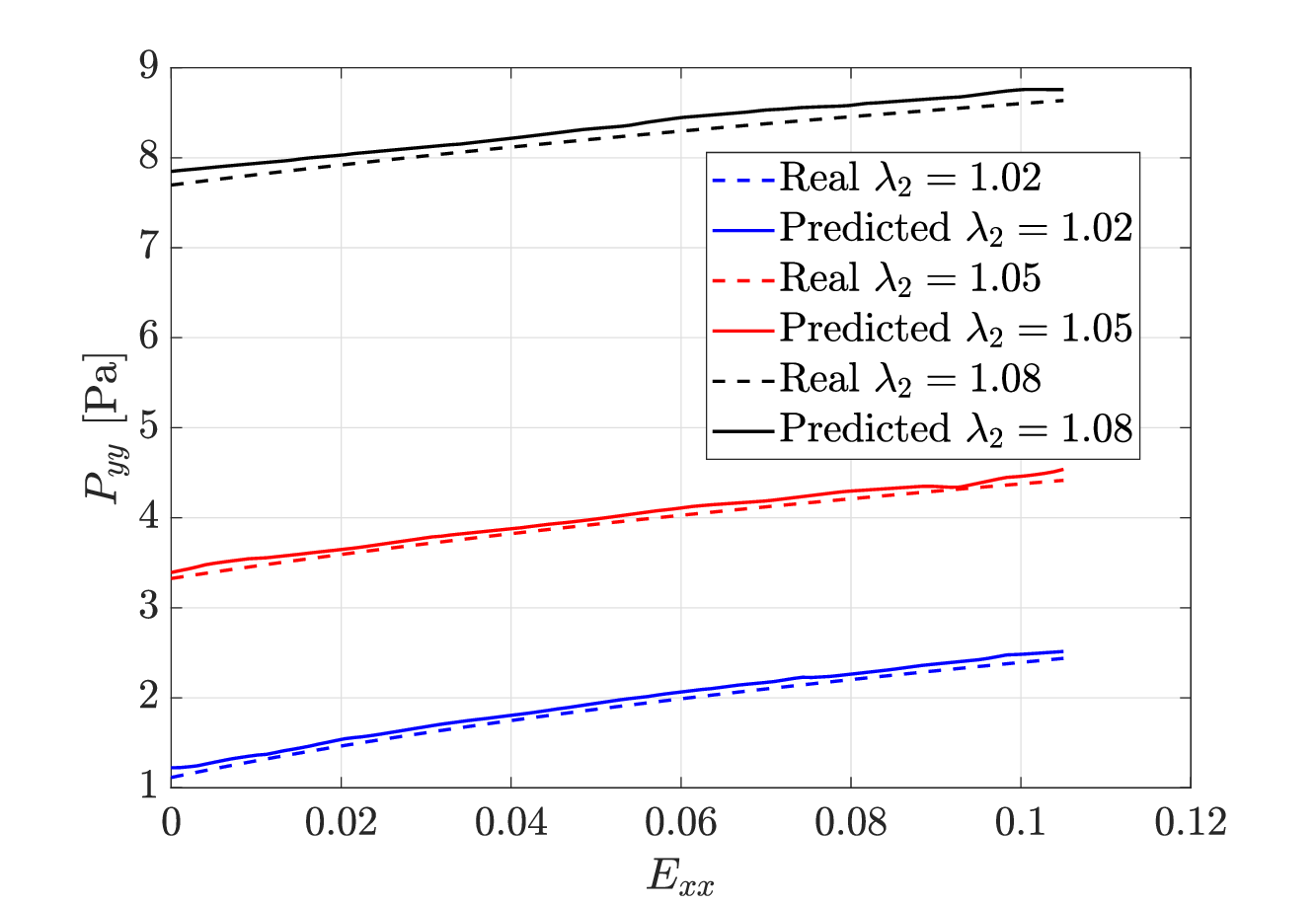}   %[Para su tamaño:anchura con respecto al ancho de la hoja]{Imágenes\Nombre_de_la_imagen}
    \caption{Component $yy$.}
    \label{og2}   %Nombre para referirse a la imagen
 \end{subfigure}
\caption{\textbf{Non-parametric PGNNIV prediction versus analytic solution of the biaxial test curve for the Ogden material.} We observe good agreement between FEM solution (continuous line) and PGNNIV prediction (dashed line), for the different values of $\lambda_2$.}
\label{fig:curve_global_large}
\end{figure}

The results are shown in Fig. \ref{fig:curve_global_large} for three different values of $\lambda_2$, and the errors, computed according to Eqs.(\ref{sigma_under}) are displayed in Table \ref{tab:exp_errors_large}.

\begin{table}[H]
\centering
\begin{tabular}{|c|c|c|c|c|}
\hline
$Y$ stretch ratio & \multicolumn{2}{|c|}{\textbf{Parametric model}} & \multicolumn{2}{|c|}{\textbf{Non-parametric model}}\\\hline
 & $\mathrm{RE}_{xx}(\mathsf{H};\lambda_2)$ (\%) & $\mathrm{RE}_{yy}(\mathsf{H};\lambda_2)$ (\%) & $\mathrm{RE}_{xx}(\mathsf{H};\lambda_2)$ (\%) & $\mathrm{RE}_{yy}(\mathsf{H};\lambda_2)$ (\%) \\ \hline
$\lambda_2 = 1.02$ & 0.98  & 2.65 & 0.34  & 3.57\\ \hline
$\lambda_2 = 1.05$ & 1.15  & 1.93 & 0.33   &1.77\\ \hline
$\lambda_2 = 1.08$ & 1.32  & 1.10 & 0.39  & 1.52 \\ \hline
\end{tabular}
\caption{Explanatory errors for the finite strains problem subjected to uniform biaxial test.}
\label{tab:exp_errors_large}
\end{table}

\section{Discussion,  conclusions and future work} 
\label{section:conclusion}

Throughout this work we have presented the mathematical foundations of PGNNIVs in the field of computational solid mechanics, which demonstrates to be a particularly interesting niche for the use of such methodology.  We have demonstrated that PGNNIVs have both predictive and explanatory capacity:
\begin{itemize}
    \item \textbf{Predictive capacity}: PGNNIVs are able to accurately predict the solid response to new external stimuli in real time, something fundamental for optimization, control and probabilistic problems. They are also able to predict not only the solid response in terms of displacement field, but also the deformation and stress fields, without the need of any extra post-processing. This has been demonstrated in Section \ref{subsec:prediction}, where we have obtained relative errors always below $10\%$. Controlling non-primary fields is sometimes important in engineering problems, as high stresses cause damage, plasticity or structural failure. As the explanatory network, once trained, encodes all the information about the material properties, it can be used for the prediction of stresses directly from the displacement fields, if necessary.
    \item \textbf{Explanatory capacity:} PGNNIVs are able to unveil hidden state model equations, that is, the constitutive equations of computational solid mechanics. First, for parameter identification and fitting, PGNNIV are able to identify inherent material symmetries (such as isotropy) and also to predict the value of the structural model parameters with high accuracy. In the latter, PGNNIVs are in a certain sense an alternative to conventional least-square minimization problems \cite{dennis1996numerical} (e.g. using standard methods such as Levenberg-Marquardt algorithm), but making the use of software and hardware tools associated with ANN technology: Graphical Processor Units (GPUs) and Tensor Processor Units (TPUs), distributed and cloud computation, scalability, transfer and federate learning strategies among others. In addition, PGNNIVs address the more challenging problem of  model-free unravelling of nonlinear materials constitutive laws. In Section \ref{subsec:explanation}, we have demonstrated the explanatory capacity of PGNNIVs both for parameter identification and state model discovery with many examples (linear and nonlinear materials both in the infinitesimal and finite strains framework). The relative error when predicting structural parameters is always below $2\%$ except if the data-set does not contain information about the material response to stimuli associated with a certain parameter or the parameter itself has not a direct impact in the explanatory capacity (superfluous parameters). Besides, the explanatory relative error is below $10\%$ for all the cases analyzed.
\end{itemize}

One important characteristic of PGNNIVs related to this double capacity is the fact that it is possible to decouple two sources of variability in the data obtained from a mechanical system that can be measured using any kind of sensor, namely, the stimuli variability and the variability related to system response. 

\begin{itemize}
\item The stimuli variability is stored in the predictive network, which acts as an autoencoder. The encoder is able to map data that lives in a space of dimension $D$, to a latent space of dimension $d\ll D$. The size of the latent space is therefore informative about the data variability, and the values of the latent variables are a compressed representation of data. In addition to theoretical considerations, this fact has important practical consequences: if we know the sources of variability of our system, it is possible to design the predictive network accordingly.
\item The physical interpretable knowledge, that is, the constitutive equation of the material, is delocalized, spread and diluted in the weights of the predictive network, but is also encoded in the relation $\boldsymbol{\varepsilon} \mapsto \boldsymbol{\sigma}$ or $\boldsymbol{E} \mapsto \boldsymbol{P}$ that is learned by the explanatory network, in a much more structured way. Indeed, if it is intended to identify and adjust some structural parameters, this is possible. Otherwise, the constitutive model as a whole may be also unravelled using the expressiveness of ANN methods.
\end{itemize}

In that sense, PGNNIVs are knowledge generators from data as most of ML techniques are, with the difference that for this particular method, the physical knowledge is directly distilled in a separated component. This particularity is what makes the difference between PINNs and PGNNIVs. In PINNs, data and mathematical physics models are seamlessly integrated for solving parametric PDEs of a given problem \cite{raissi2019physics,karniadakis2021physics}, but, by construction, the information cannot be extrapolated to other situations. That means that, in the context of solid mechanics, the network trained for a given problem using PINNs cannot be used for predicting the response of the system under different volume loads or boundary conditions, which greatly weakens its ability as a predictive method. PGNNIVs overcome this difficulty precisely by distilling the physical information from the intrinsic variability of the stimuli.

There is another paramount characteristic of PGNNIVs for computational solid mechanics that has been largely discussed and has explained their emergence. There is no need to have access to the values of internal variables, that are, strictly speaking, non-measurable as they are mathematical constructs coming from a scientific theory. In that sense, even if the bases for thermodynamically-appropriated ANN for constitutive equations in solid mechanics have been investigated \cite{linka2023new}, it is important to recall that stress fields, such as $\boldsymbol{P}$ or $\boldsymbol{\sigma}$ are not accessible without the need of extra hypothesis (geometry or load specific configurations). This fundamental issue should not be overlooked and many recent works have worked for this purpose. Efficient Unsupervised Constitutive Law Identification and Discovery (EUCLID) method is one of the most acclaimed efforts in that direction, either using sparse identification \cite{flaschel2021unsupervised,flaschel2022discovering,flaschel2023automated}, clustering \cite{marino2023automated}, Bayesian methods \cite{joshi2022bayesian} or ANN \cite{thakolkaran2022nn}. However, EUCLID paradigm relies on the fact that the geometry and loads are appropriate enough to ensure strain-stress fields variability in a single specimen. When the geometry or the data acquisition capabilities do not satisfy this requirement, the only possibility is to take action on the data-set or the network, or at least to track the different load conditions and incorporate them to the computational pipeline.

Finally, among the different PIML methods, PGNNIVs are more transparent than other approaches that have demonstrated to be very performant for computing the evolution of dynamical systems by incorporating thermodynamical constraints. Structure Preserving Neural Networks enforce first and second laws of thermodynamics as regularization term \cite{hernandez2021structure}. Even if in the cited work the GENERIC structure allows for a split between reversible and non-reversible (dissipative) components,  the physical information is again diluted in all the network weights, rather than in some specific components. A dimensionality reduction of the dynamic data was also explored \cite{hernandez2021deep}. This may be interpreted also as an information reduction to distil physical knowledge, although the interpretability is still dark. In PGNNIVs, however, interpretable physical information (that is, knowledge) is located in specific ANN components.

Nevertheless, the presented methodology still has some limitations and there exists room for exploration in several directions:
\begin{enumerate}
\item The data requirements for the problem in hands are high. In this work, we have generated data synthetically, but in reality sensors usually collect noisy data from experimental tests and there exist also important limitations concerning the size of the data-sets. A probabilistic (Bayesian) viewpoint will enable a new interpretation of PGNNIVs in the \emph{small data} regime, although this methodology is rather thought for systems where intensive data mining is possible, in which data quantity prevails over data quality.

\item Defining a suitable architecture for the $\mathsf{Y}$ and $\mathsf{H}$ networks is not a simple task and requires an iterative process, including the tuning of many hyperparameters. In addition, PGNNIVs require extra hyper-parameters, i.e. penalty coefficients $p_i$, making the process even more involved and time consuming. We have presented some insights into to the complexity and architecture of both predictive and explanatory networks, related with both their prediction and explanation character, but either a great intuition for network design, or time-intensive trial-and-error iterative testing are needed.

\item In this work, we have made used of relatively coarse discretizations, but  finer meshes will result in more expensive training processes due to the exponentially larger number of parameters required. More powerful computational strategies (distributed computing, parallelization) as well as more advanced hardware (GPUs and TPUs) will enable the acceleration of the PGNNIVs training processes although the problem of dealing with multidimensional and unstructured meshes still remains open.
\end{enumerate}

Many challenges lie ahead of the development of a more general PGNNIV framework. Next lines of research will address the formulation of PGNNIVs under finite strain assumptions using a much more theoretical basis, such as the presented in some recent works \cite{linka2023new,linka2023automated,linden2023neural}, leveraging its predictive and explanatory power. Furthermore, extensions of the 2D planar stress architecture to general 3D problems with more complex geometries and load scenarios as well as to more complex constitutive laws that might depend on time (visco-elasticity), or heterogeneous conditions, still pose major challenges for the future.

Finally, the usage of real data from sensors, for example, through Digital Image or Volume Correlation tests (DIC, DVC) \cite{chu1985applications} or even more advanced methods such as Finite Element Model Updating (FEMU) \cite{mottershead2011sensitivity} or Virtual Fields Methods (VFM) \cite{grediac2006virtual} will put to the test the applicability of PGNNIVs to real scientific problems in the field of engineering.

In conclusion, we have demonstrated that in the context of computational solid mechanics, PGNNIVs are a family of ANN for accurately predicting measurable and non-measurable variables such as displacement and stress fields, in real time, and are also able to describe or unravel the constitutive model with high accuracy for different linear and nonlinear (hyper-)elastic materials. Even if this work is preliminary, the ingredients that it comprises correspond to a general approach and the methodology can be applied to cases of scientific interest with the necessary adaptations of the network architectures.

%Bibliography
\bibliographystyle{unsrt}  
\bibliography{references}  

\newpage

\appendix
\renewcommand\thesection{\Alph{section}}

\renewcommand{\appendixname}{Appendix}

% Appendix content here...
\begin{appendices}

\section{Details about network architectures and hyperparameters.} \label{app:network_details}

\subsection{Network architectures used}

As explained in Section \ref{subsubsection:operators}, we use different combinations of architectures for both predictive and explanatory networks, $\mathsf{Y}$ and $\mathsf{H}$, depending on the problem. Table \ref{hyper1} shows the architecture details for the unique PGNNIV trained on the linear, softening and hardening materials in small strains theory when the parametric structure of the constitutive equation is not prescribed. For the anisotropic and isotropic parametric discovery, the explanatory network is substituted by the direct parametric expression. The $\mathsf{Y}$ network layer includes empty elements (null value) corresponding to the most general natural boundary conditions where shear loads are applied on the contour. Therefore, the input layer comprises $2 \times (n_x - 1) \times (n_y - 1) = 2 \times (10 +8) = 36$ neurons. The number of neurons of the output layer corresponds to $2 \times n_x \times n_y =  2\times 11\times 9  = 198$.

\begin{table}[H]
\centering
\begin{tabular}{|l|l|l|}
\hline
\textbf{Network} & \textbf{Class of architecture} & \textbf{Detailed layers (input and output neurons included)}  \\
\hline
 Predictive network $\mathsf{Y}$ & MLP & 36/50/tanh/50/tanh/50/tanh/50/tanh/198         \\ \hline
 Explanatory network $\mathsf{H}$ & mMLP & 3/100/tanh/100/tanh/100/tanh/100/tanh/100/tanh/100/tanh/100/tanh/3                \\ \hline
\end{tabular}
\caption{Architecture of the PGNNIV for the softening, hardening and linear materials.}
\label{hyper1}
\end{table}

Table \ref{hyper2} shows the implementation details for the unique PGNNIV trained on the Ogden material in finite strain theory when the parametric structure of the constitutive equation is not prescribed. For the Ogden constitutive model parametric discovery, the explanatory network is substituted by the direct parametric expression. 
\begin{table}[H]
\centering
\begin{tabular}{|l|l|l|}
\hline
\textbf{Network} & \textbf{Class of architecture} & \textbf{Detailed layers (input and output neurons included)} \\
\hline
Predictive network $\mathsf{Y}$  & autoencoder &18/9/tanh/3/tanh/3/tanh/9/tanh/198                   \\ \hline
Explanatory network $\mathsf{H}$ & mMLP &3/60/ReLU/60/ReLU/60/ReLU/60/ReLU/60/ReLU/3                   \\ \hline
\end{tabular}
\caption{Architecture of the PGNNIV for the Ogden material.}
\label{hyper2}
\end{table}

\subsection{Hyperparameters}

\label{sec:hyper}

For reproducibility purposes, Table \ref{pre} shows a summary of the hyperparameters that were found to give optimal results for the optimization process.

\begin{table}[H]
\centering
\begin{tabular}{|l|l|l|l|l|l|l|}
\hline
& \multicolumn{4}{|c|}{\textbf{Loss function coefficients}} & \multicolumn{2}{|c|}{\textbf{Learning algorithm}} \\ \hline
& $p_1$  & $p_2$ & $p_3$ & $p_4$  & $\alpha$ (Learn. rate) & $n_b$ (batch size) \\ \hline
Linear  (model-free)  & $1\times 10^3$  & $1\times 10^{-3}$  & $1\times 10^{-3}$ & $1\times 10^3$ & $5 \times 10^{-5}$  &  $100$              \\ \hline
Linear (anisotropic model)   & $1\times 10^3$  & $1\times 10^{-3}$  & $1\times 10^{-3}$ & $1\times 10^3$ & $5 \times 10^{-5}$  &   $100$             \\ \hline
Linear (isotropic model)   & $1\times 10^3$  & $1\times 10^{-3}$  & $1\times 10^{-3}$ & $1\times 10^3$ & $5 \times 10^{-5}$  &   $100$              \\ \hline
Softening (model-free)& $1\times 10^3$ & $1\times 10^{-3}$  & $1\times 10^{-3}$ & $1\times 10^3$ & $5 \times 10^{-5}$   &   $100$            \\ \hline
Hardening (model-free)&  $1\times 10^4$ & $1\times 10^{-8}$  & $1\times 10^{-8}$ & $1\times 10^4$ & $5 \times 10^{-4}$   &   $100$           \\ \hline
Ogden (model-free)& $1.0\times 10^0$ & $1\times 10^{-3}$ & $1\times 10^{-3}$ & $1\times 10^{-3}$ & $1\times 10^{-3}$ & $128$\\ \hline
Ogden (model-based) & $1.0\times 10^0$ & $1\times 10^{-3}$ & $1\times 10^{-3}$ & $1\times 10^{-3}$ & $1\times 10^{-3}$ & $128$\\ \hline
\end{tabular}
\caption{Value of the hyperparameters for all the training processes.}
\label{pre}
\end{table}

\clearpage
\section{Some field solution predictions of the PGNNIV for the linear, softening, hardening and Ogden materials.}
\label{sec:colormaps}

As an illustration of the predictive capacity of the methodology, in this Appendix we illustrate some solution fields  associated to the linear, softening and hardening material problems, as well as for the Ogden-like material.

\subsection{Infinitesimal theory.}
\subsubsection{Linear material.}

\begin{figure}[htbp]
 \centering
 \begin{subfigure}[t]{0.495\linewidth}
     \centering %Para centrar la imagen
    \includegraphics[scale=0.3]{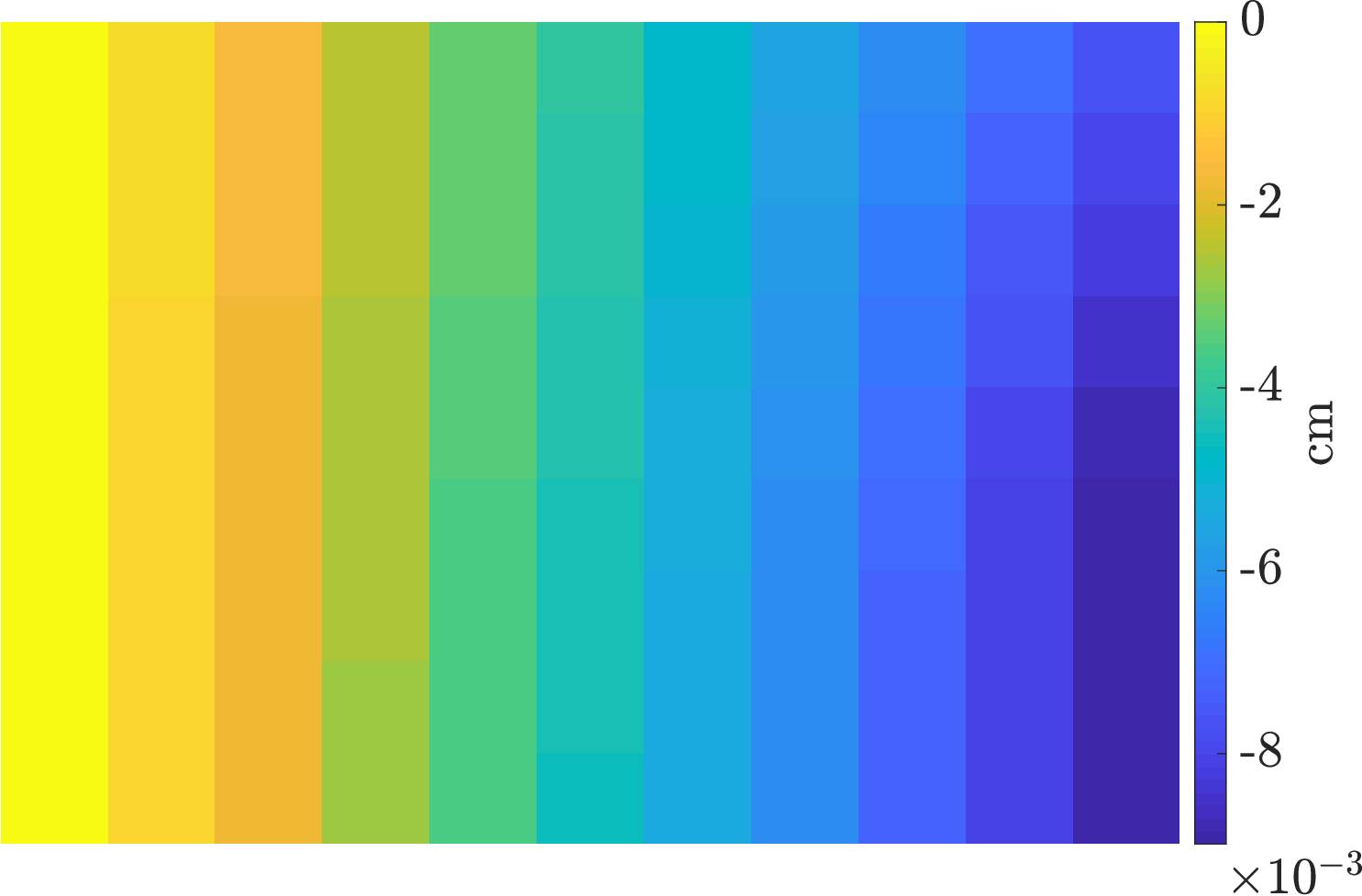}   %[Para su tamaño:anchura con respecto al ancho de la hoja]{Imágenes\Nombre_de_la_imagen}
    \caption{Real horizontal component ($u_x$).}
 \end{subfigure}
 \hfill
 \begin{subfigure}[t]{0.495\linewidth}
    \centering %Para centrar la imagen
    \includegraphics[scale=0.3]{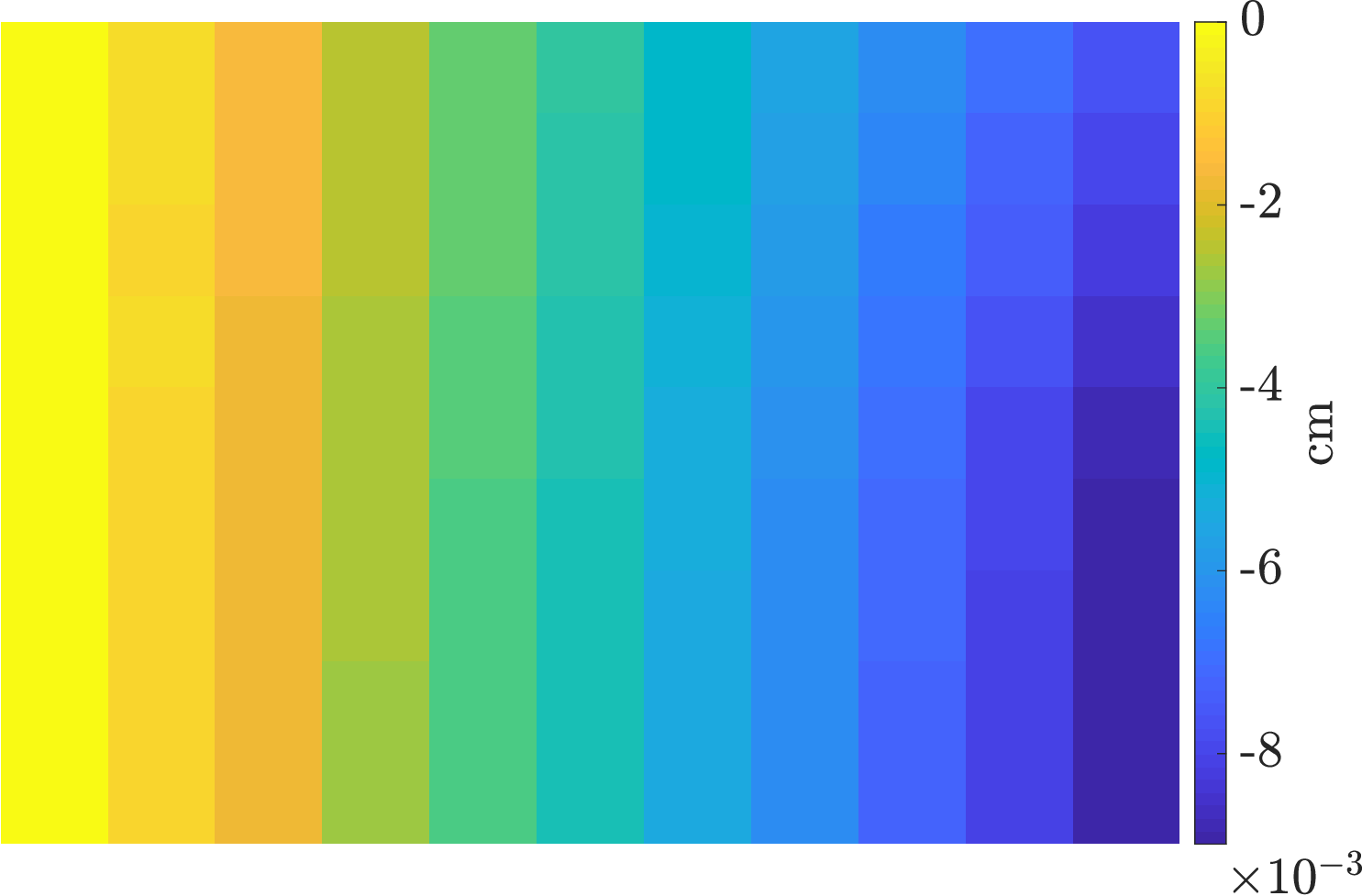}   %[Para su tamaño:anchura con respecto al ancho de la hoja]{Imágenes\Nombre_de_la_imagen}
    \caption{Predicted horizontal component ($u_x$).}
 \end{subfigure}
 \begin{subfigure}[t]{0.495\linewidth}
     \centering %Para centrar la imagen
    \includegraphics[scale=0.3]{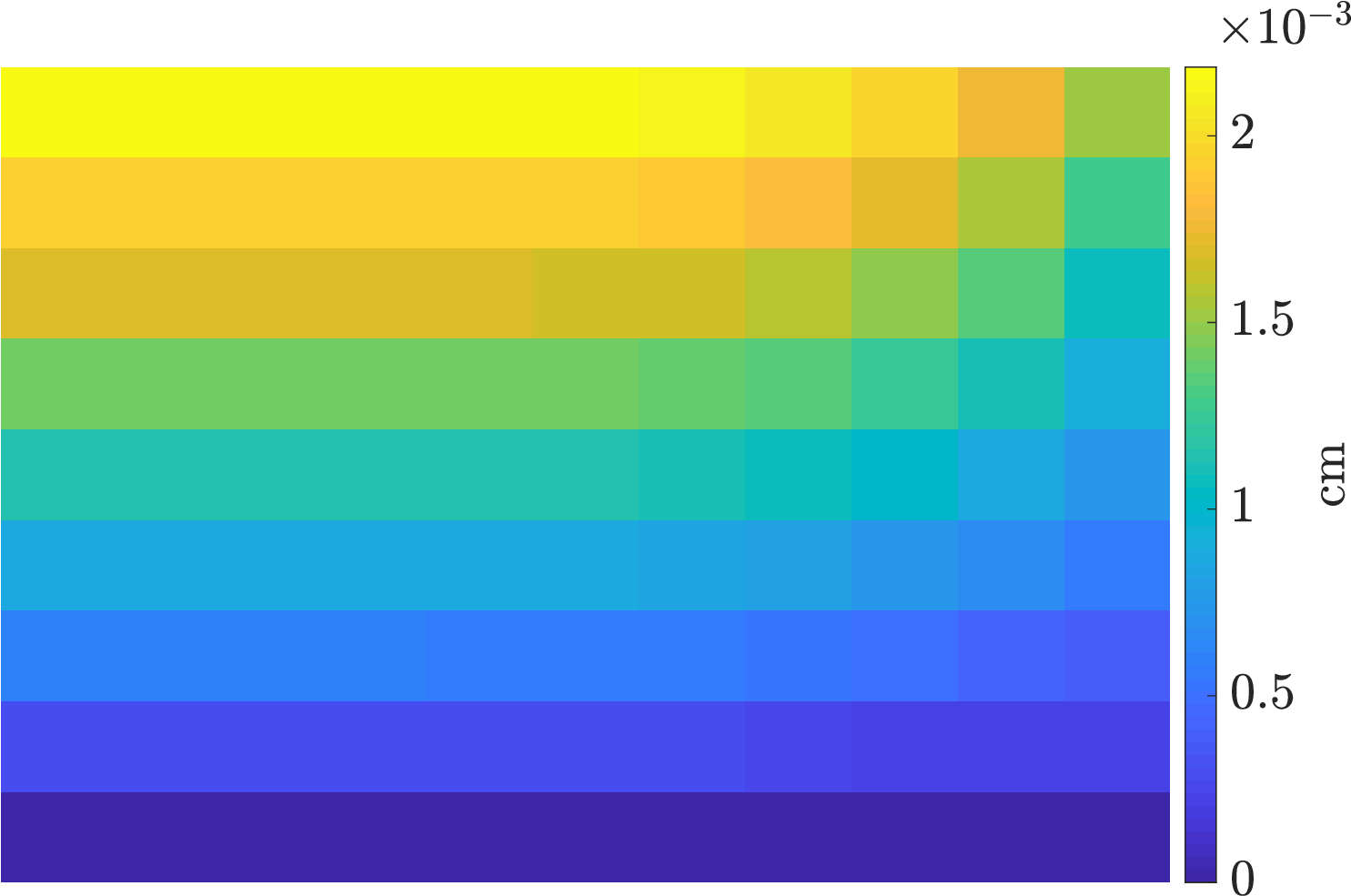}   %[Para su tamaño:anchura con respecto al ancho de la hoja]{Imágenes\Nombre_de_la_imagen}
    \caption{Real vertical component ($u_y$).}
 \end{subfigure}
 \hfill
 \begin{subfigure}[t]{0.495\linewidth}
    \centering %Para centrar la imagen
    \includegraphics[scale=0.3]{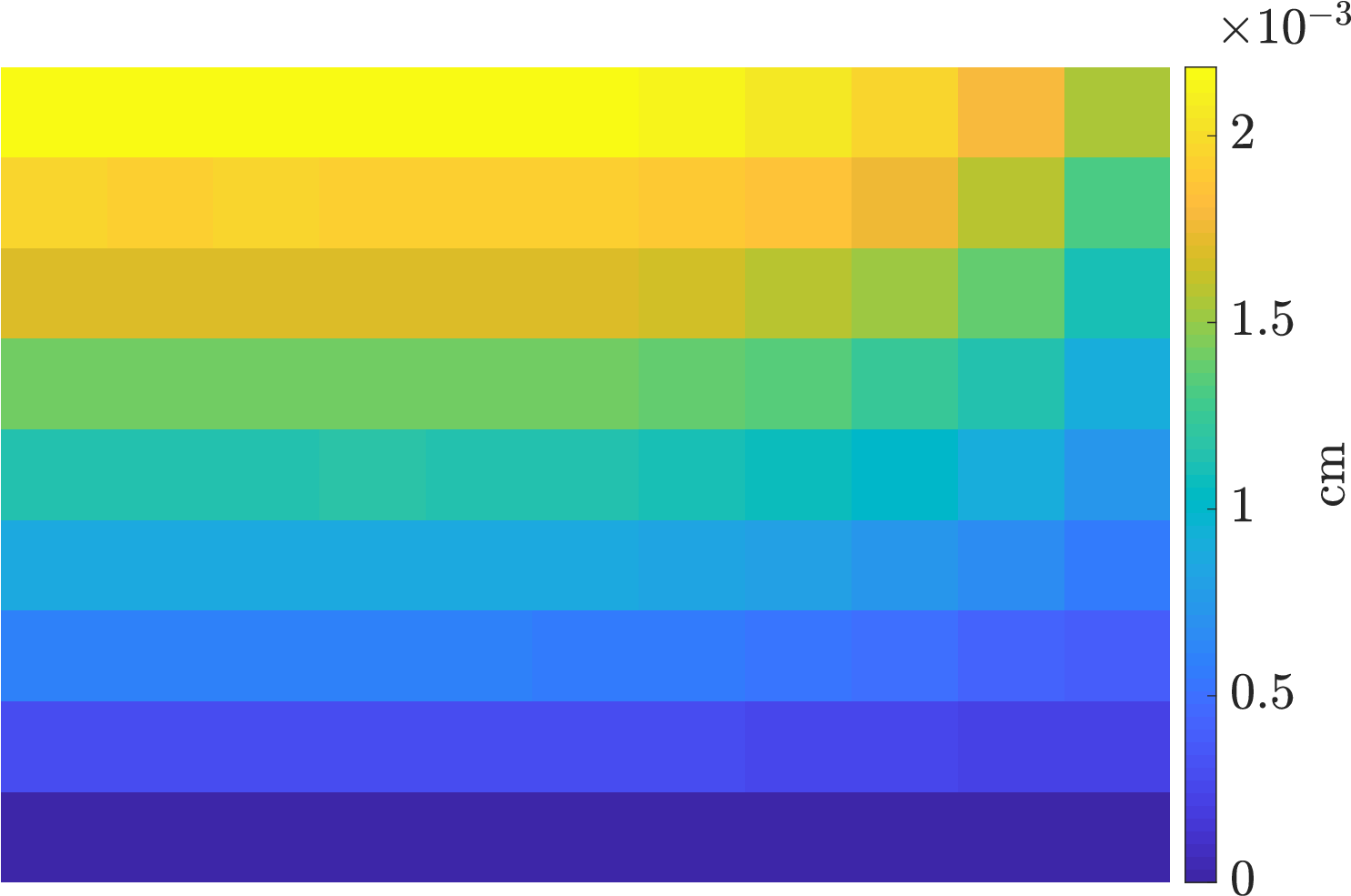}   %[Para su tamaño:anchura con respecto al ancho de la hoja]{Imágenes\Nombre_de_la_imagen}
    \caption{Predicted vertical component ($u_y$).}
 \end{subfigure}
\caption{\textbf{PGNNIV prediction versus FEM solution of the components of the displacement field for a single test-set example of the linear material.}}
\label{linu}
\end{figure}

\begin{figure}[H]
 \centering
 \begin{subfigure}[t]{0.495\linewidth}
     \centering %Para centrar la imagen
    \includegraphics[scale=0.30]{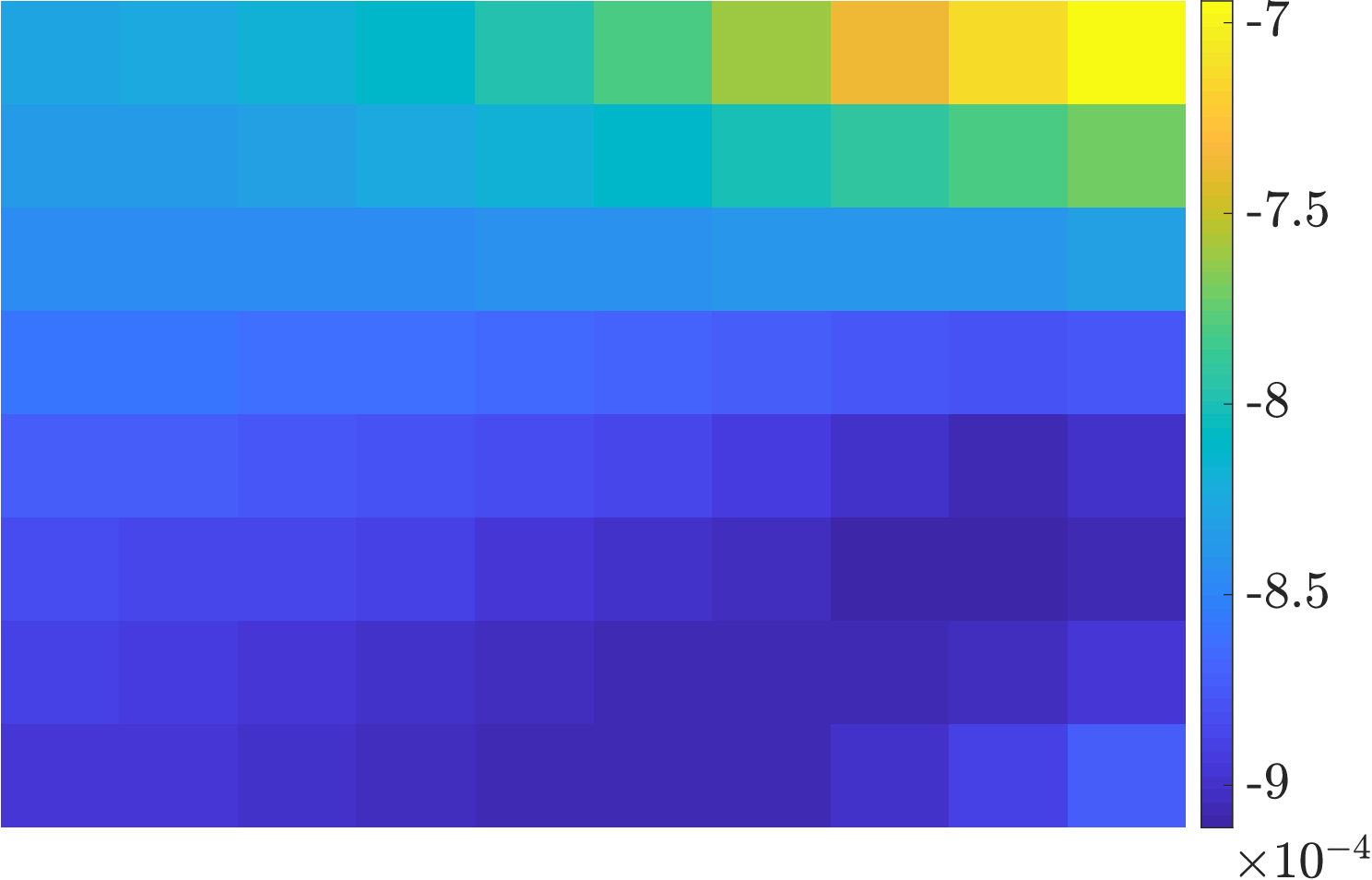}   %[Para su tamaño:anchura con respecto al ancho de la hoja]{Imágenes\Nombre_de_la_imagen}
    \caption{Real normal component ($\varepsilon_{xx}$).}
 \end{subfigure}
 \hfill
 \begin{subfigure}[t]{0.495\linewidth}
     \centering %Para centrar la imagen
    \includegraphics[scale=0.30]{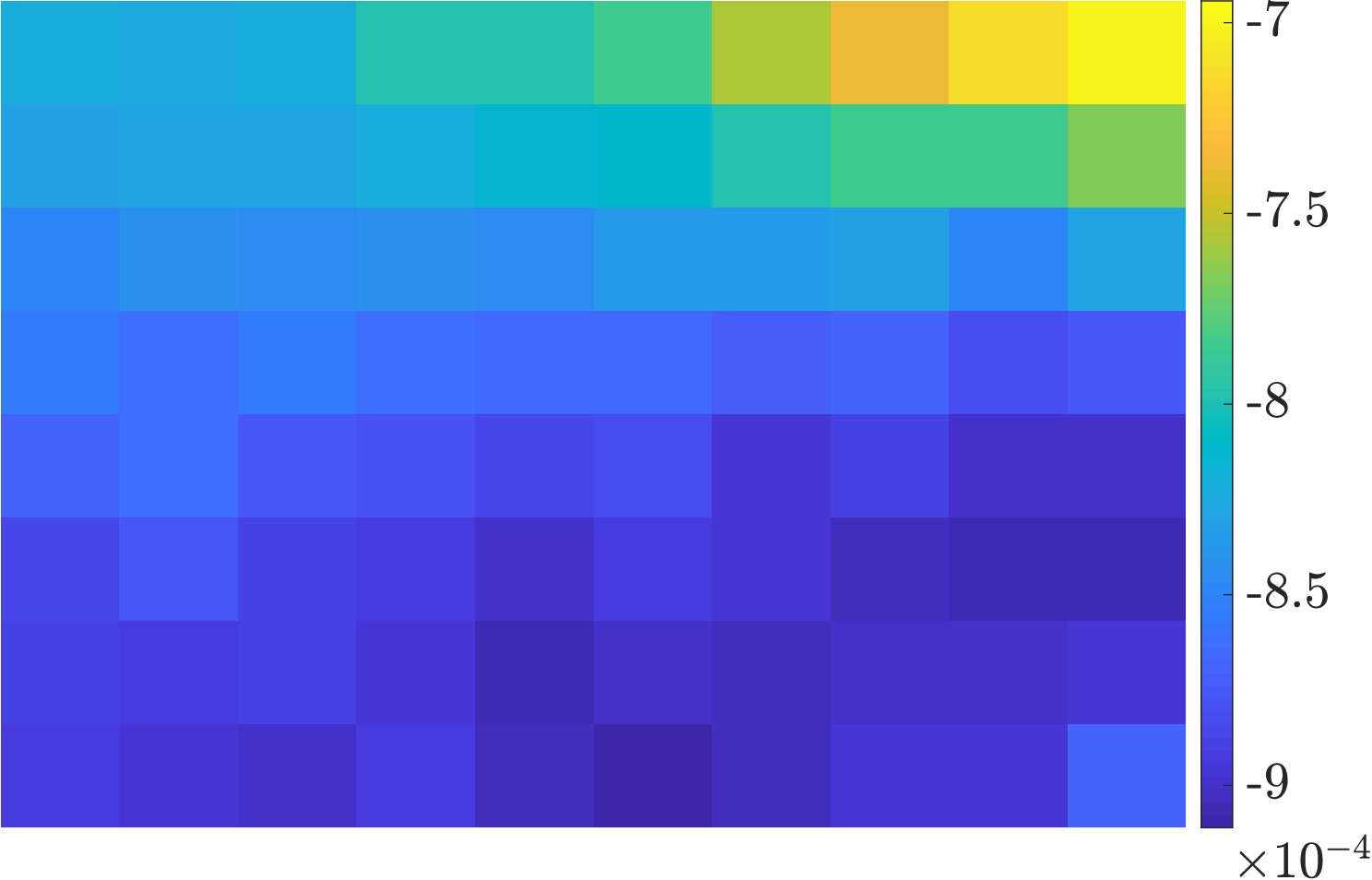}   %[Para su tamaño:anchura con respecto al ancho de la hoja]{Imágenes\Nombre_de_la_imagen}
    \caption{Predicted normal component ($\varepsilon_{xx}$).}
 \end{subfigure}
 \hfill
 \begin{subfigure}[t]{0.495\linewidth}
    \centering %Para centrar la imagen
    \includegraphics[scale=0.30]{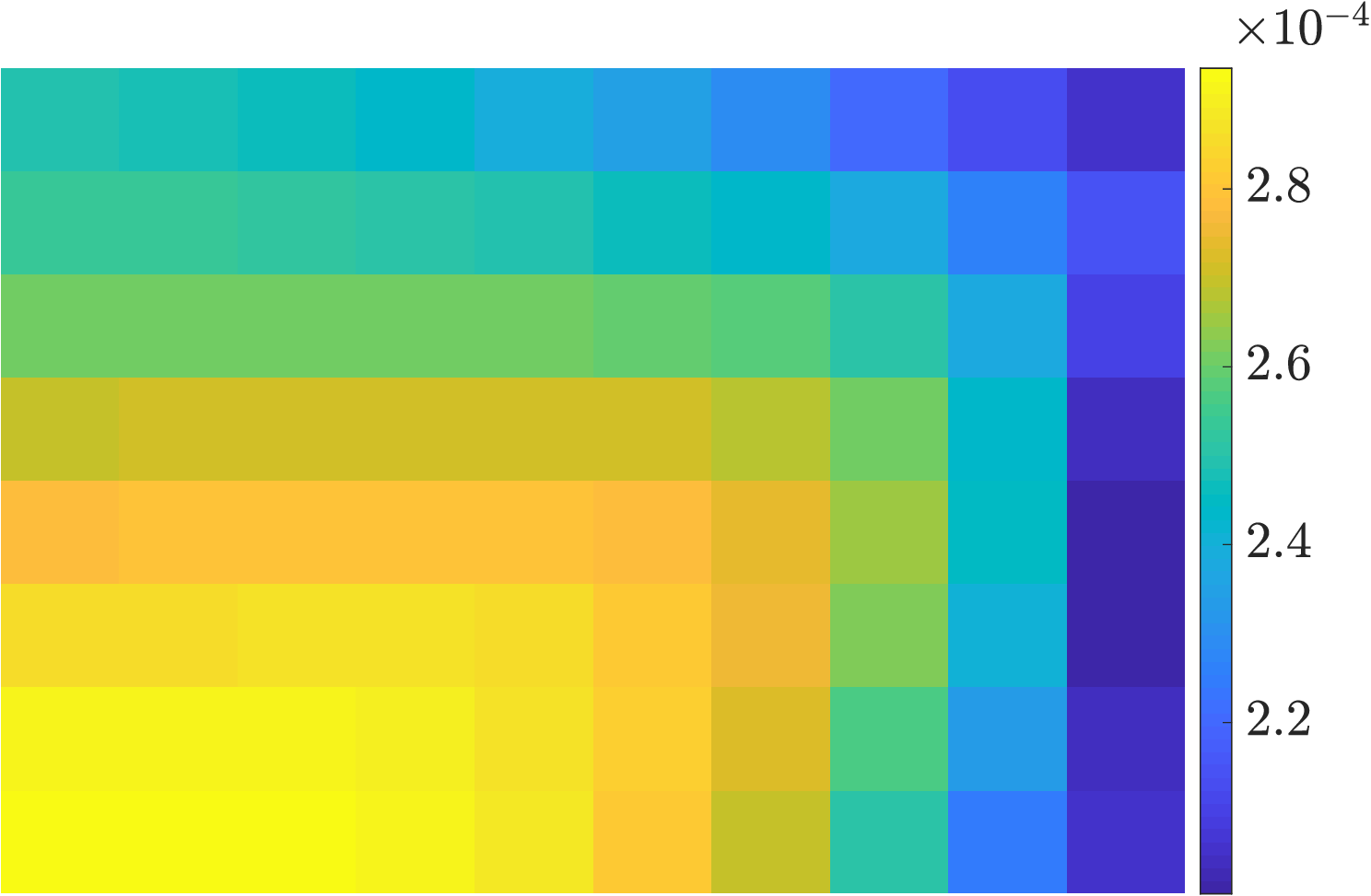}   %[Para su tamaño:anchura con respecto al ancho de la hoja]{Imágenes\Nombre_de_la_imagen}
    \caption{Real normal component ($\varepsilon_{yy}$).}
 \end{subfigure}
  \begin{subfigure}[t]{0.495\linewidth}
     \centering %Para centrar la imagen
    \includegraphics[scale=0.30]{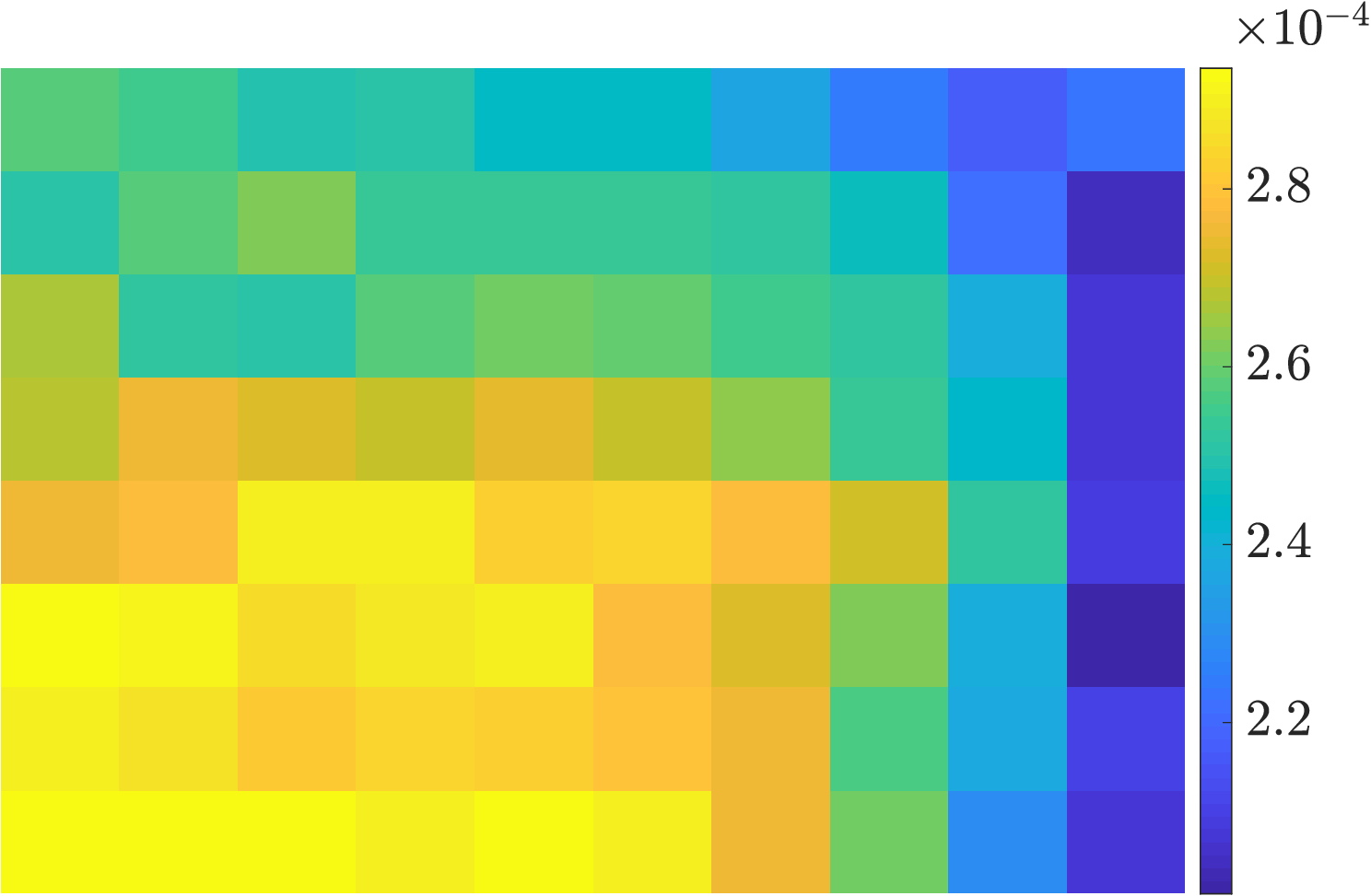}   %[Para su tamaño:anchura con respecto al ancho de la hoja]{Imágenes\Nombre_de_la_imagen}
    \caption{Predicted normal component ($\varepsilon_{yy}$).}
 \end{subfigure}
 \hfill
 \begin{subfigure}[t]{0.495\linewidth}
     \centering %Para centrar la imagen
    \includegraphics[scale=0.30]{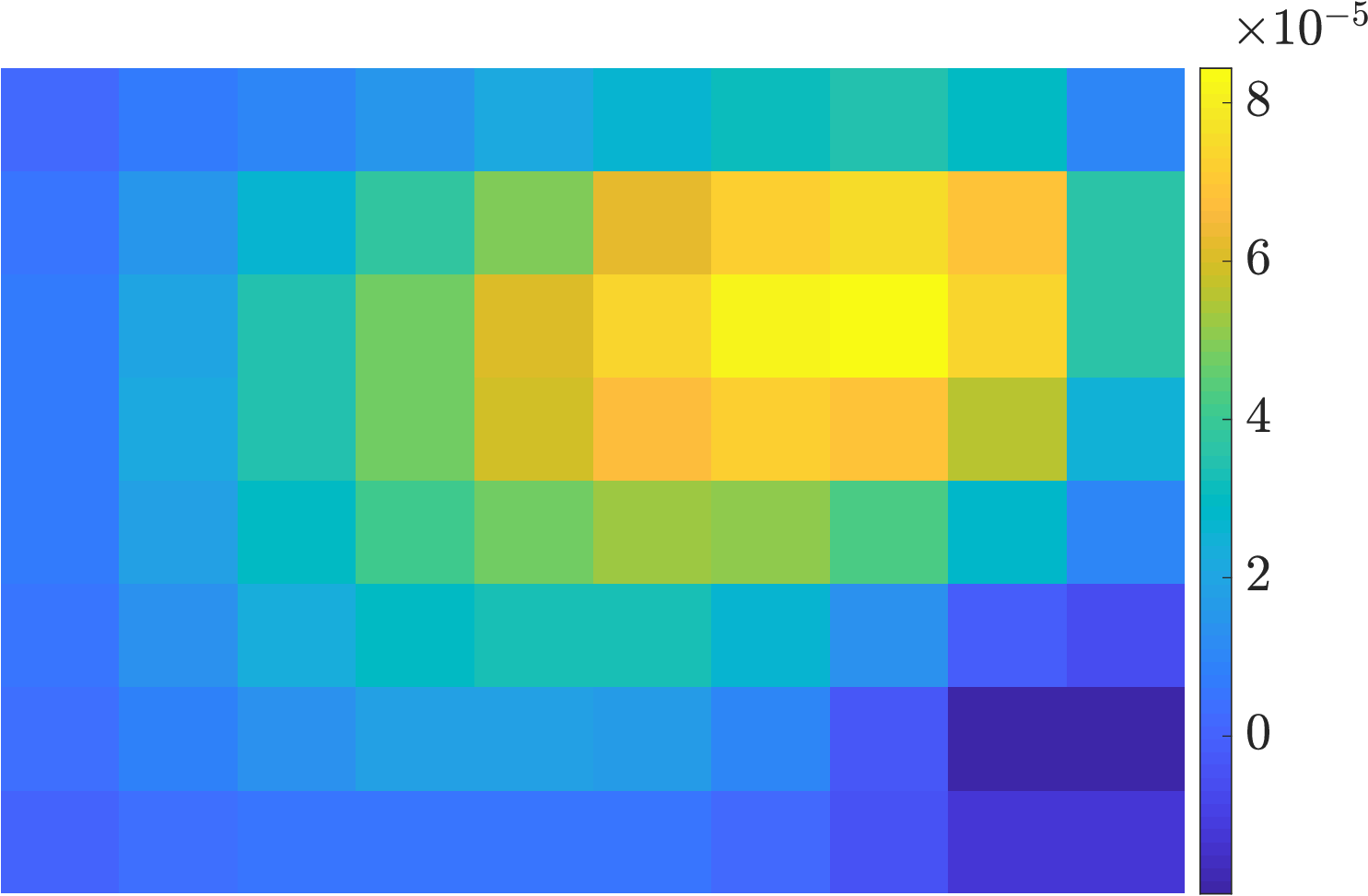}   %[Para su tamaño:anchura con respecto al ancho de la hoja]{Imágenes\Nombre_de_la_imagen}
    \caption{Real shear component ($\varepsilon_{xy}$).}
 \end{subfigure}
 \hfill
 \begin{subfigure}[t]{0.495\linewidth}
    \centering %Para centrar la imagen
    \includegraphics[scale=0.30]{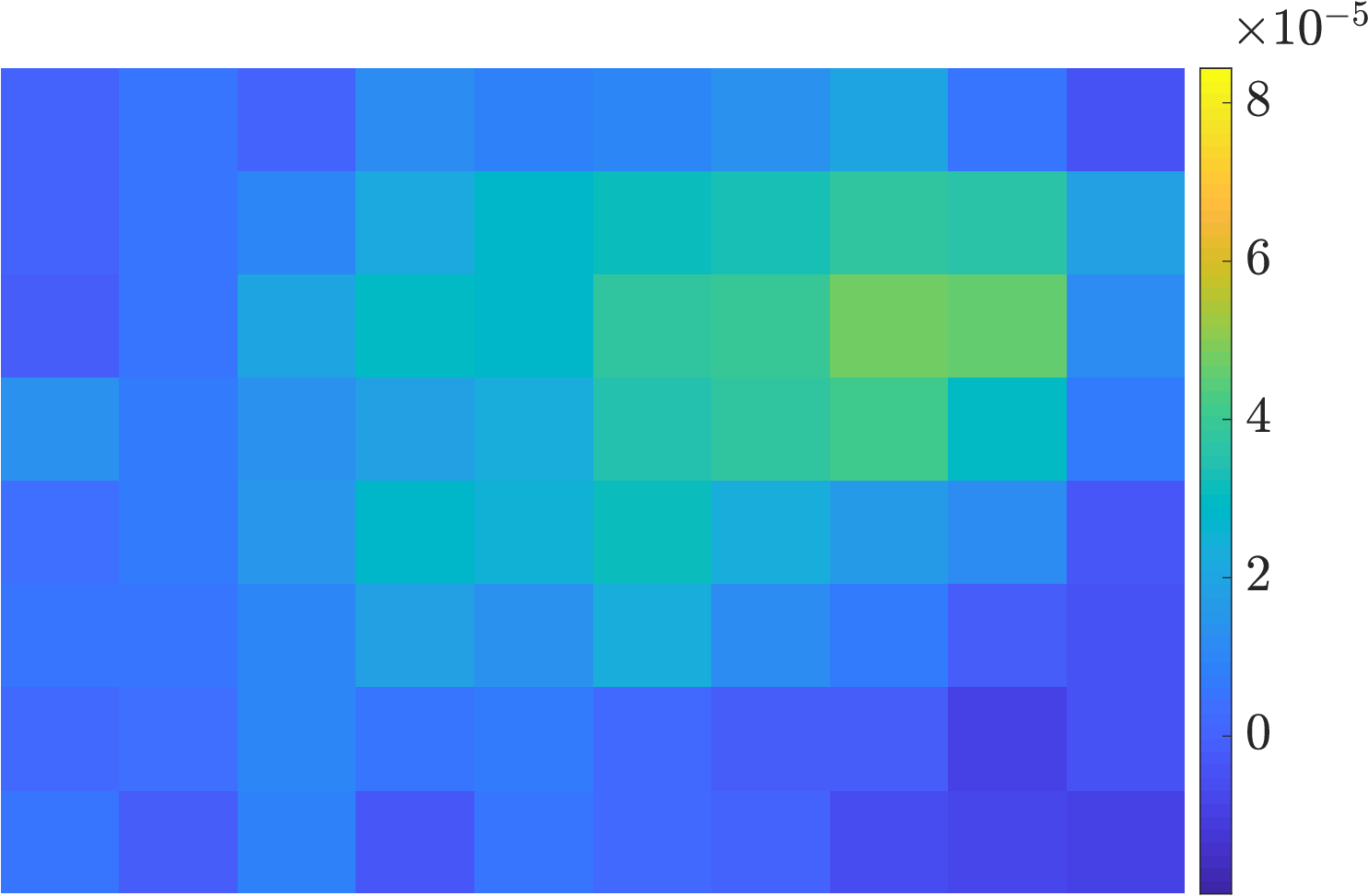}   %[Para su tamaño:anchura con respecto al ancho de la hoja]{Imágenes\Nombre_de_la_imagen}
    \caption{Predicted shear component ($\varepsilon_{xy}$).}
 \end{subfigure}
\caption{\textbf{PGNNIV prediction versus FEM solution of the components of the strain fields for a single test-set example of the linear material.}}
\label{line}
\end{figure}

\begin{figure}[H]
 \centering
 \begin{subfigure}[t]{0.495\linewidth}
     \centering %Para centrar la imagen
    \includegraphics[scale=0.3]{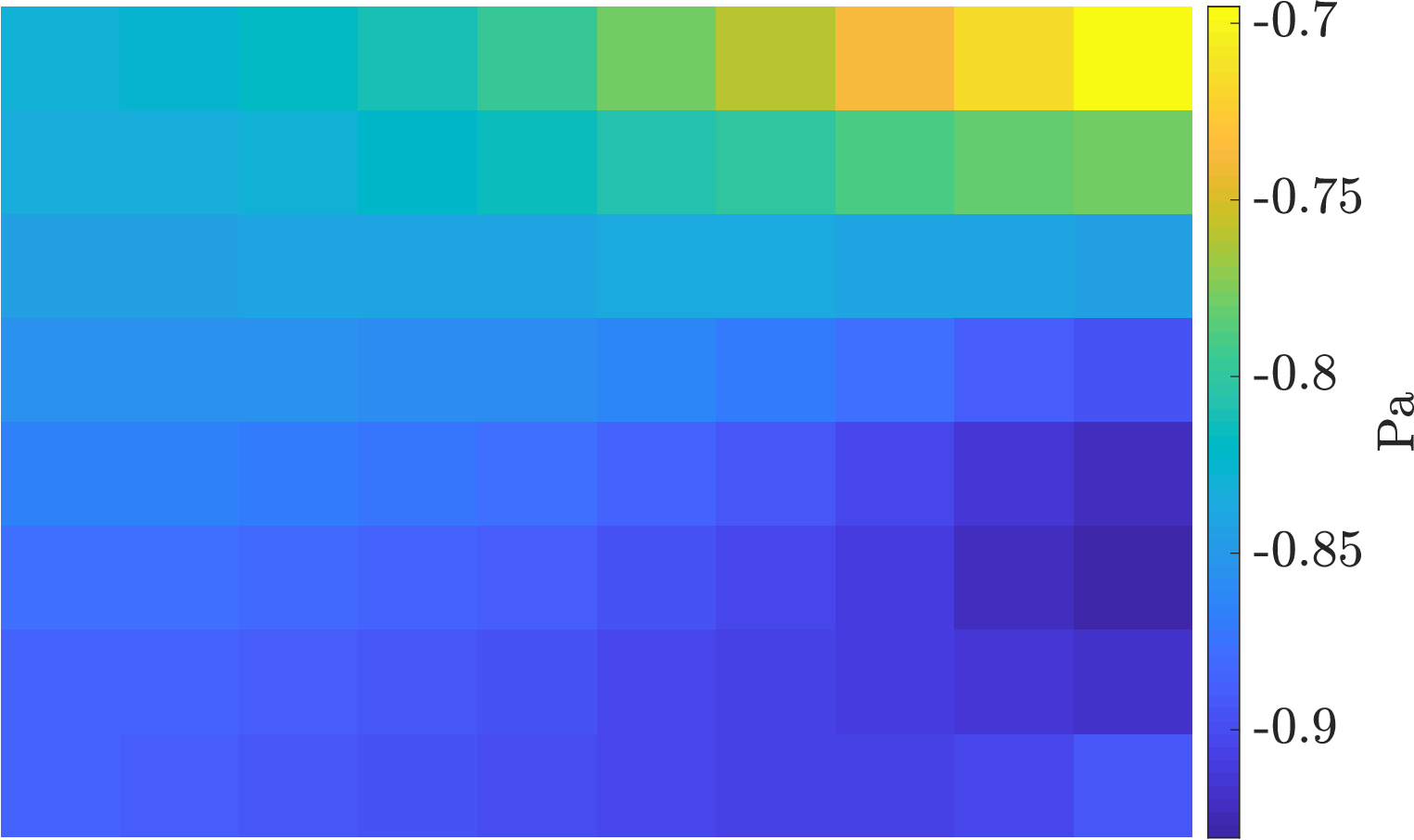}   %[Para su tamaño:anchura con respecto al ancho de la hoja]{Imágenes\Nombre_de_la_imagen}
    \caption{Real normal component ($\sigma_{xx}$).}
 \end{subfigure}
 \hfill
 \begin{subfigure}[t]{0.495\linewidth}
     \centering %Para centrar la imagen
    \includegraphics[scale=0.3]{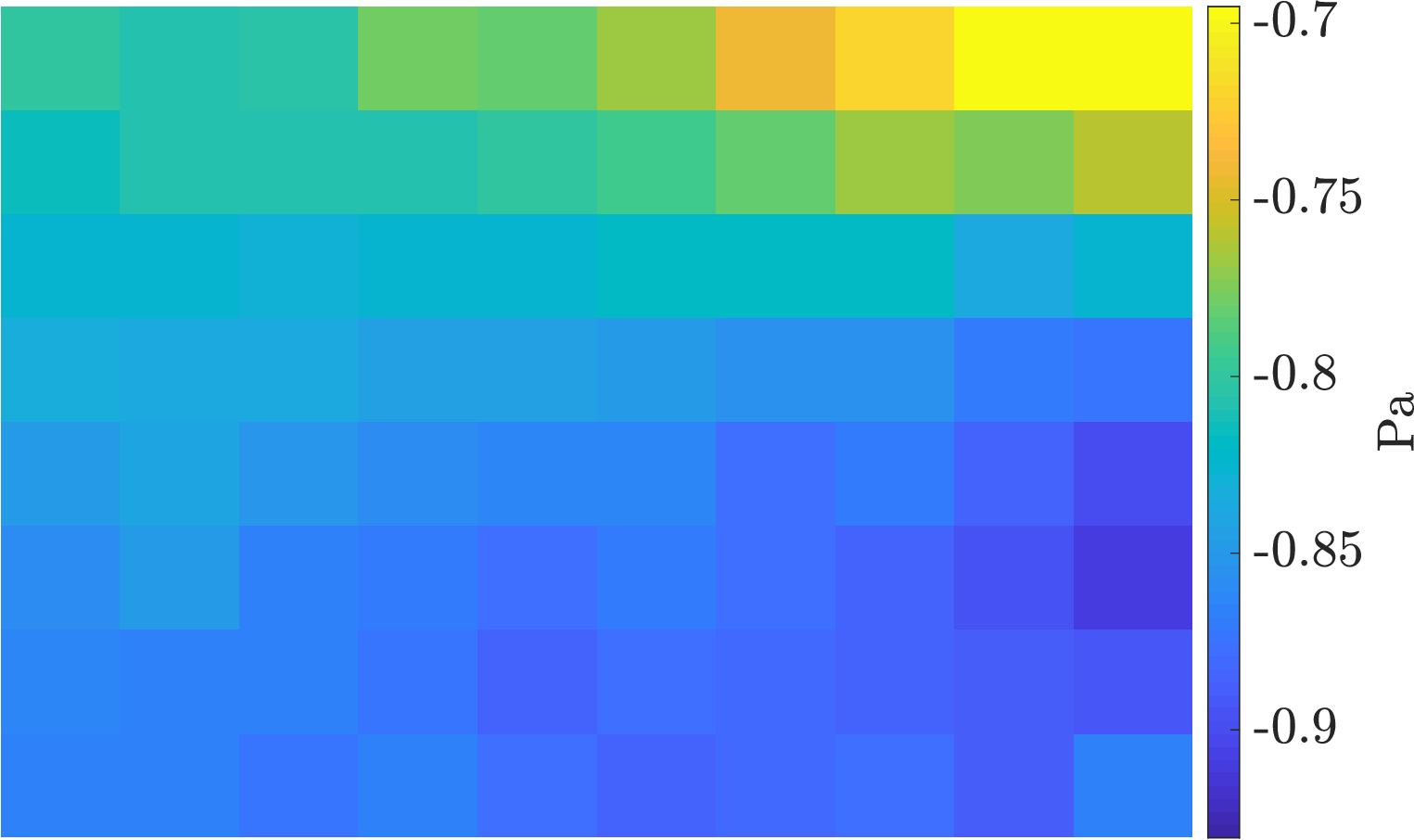}   %[Para su tamaño:anchura con respecto al ancho de la hoja]{Imágenes\Nombre_de_la_imagen}
    \caption{Predicted normal component ($\sigma_{xx}$).}
 \end{subfigure}
 \hfill
 \begin{subfigure}[t]{0.495\linewidth}
    \centering %Para centrar la imagen
    \includegraphics[scale=0.3]{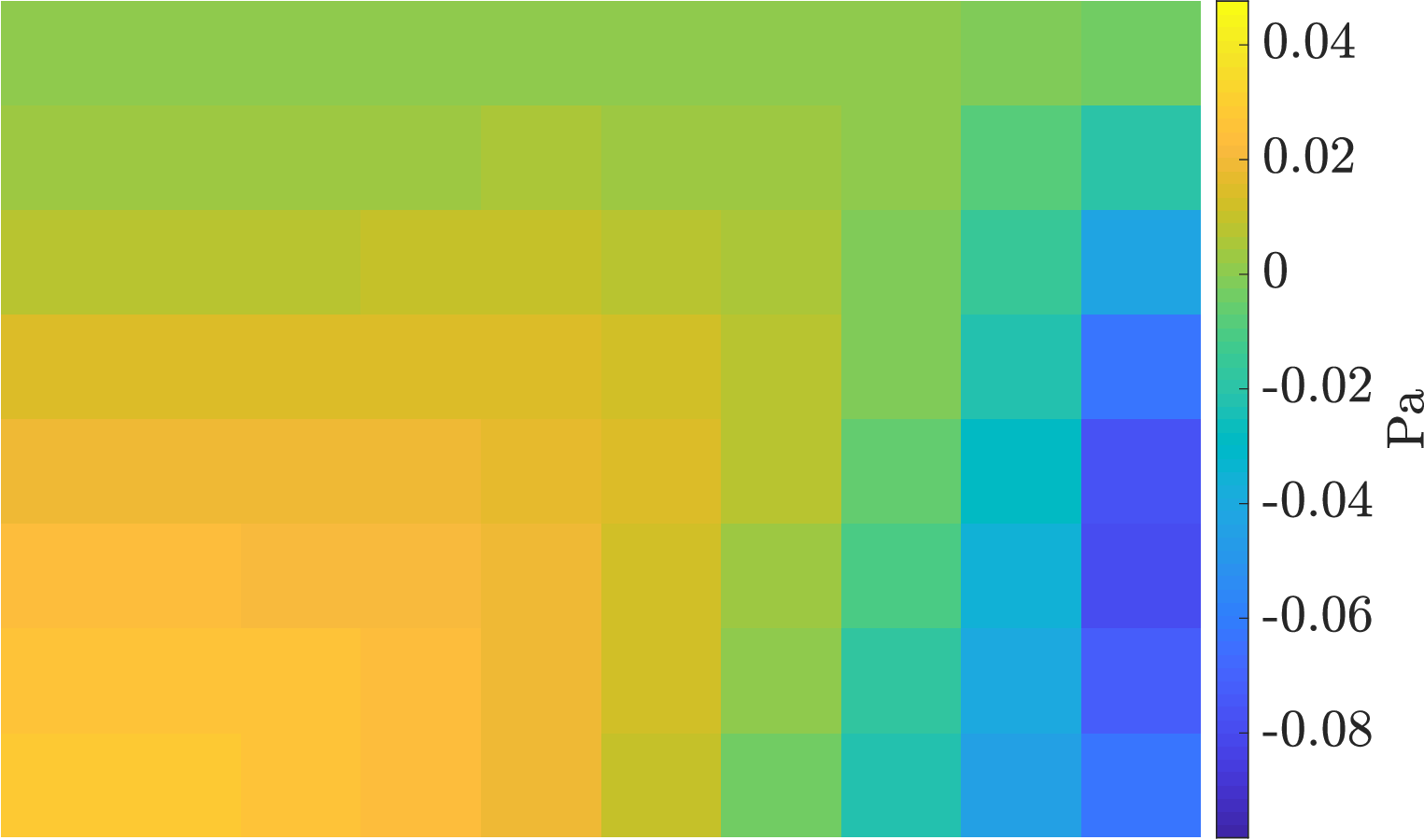}   %[Para su tamaño:anchura con respecto al ancho de la hoja]{Imágenes\Nombre_de_la_imagen}
    \caption{Real normal component ($\sigma_{yy}$).}
 \end{subfigure}
  \begin{subfigure}[t]{0.495\linewidth}
     \centering %Para centrar la imagen
    \includegraphics[scale=0.3]{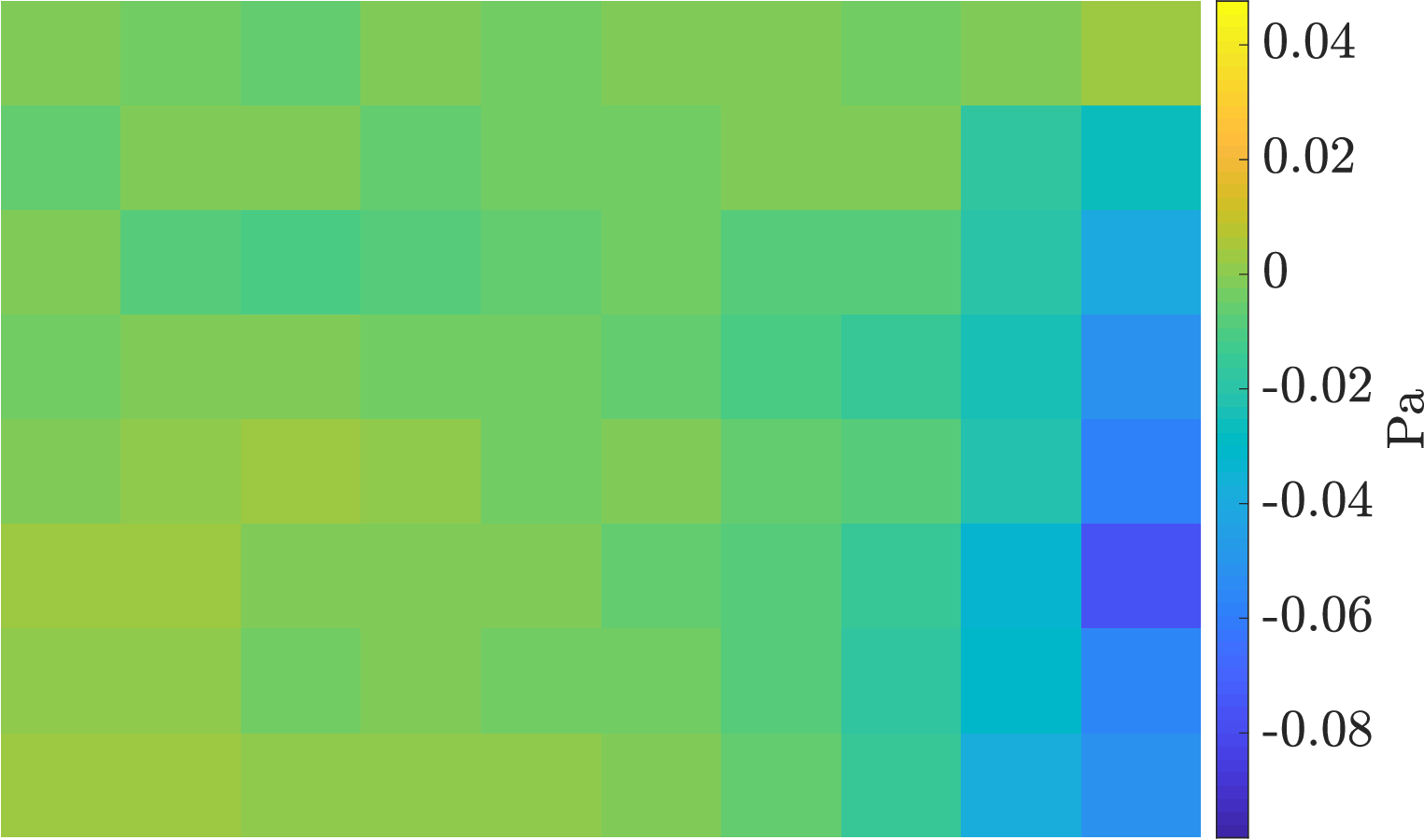}   %[Para su tamaño:anchura con respecto al ancho de la hoja]{Imágenes\Nombre_de_la_imagen}
    \caption{Predicted normal component ($\sigma_{yy}$).}
 \end{subfigure}
 \hfill
 \begin{subfigure}[t]{0.495\linewidth}
     \centering %Para centrar la imagen
    \includegraphics[scale=0.3]{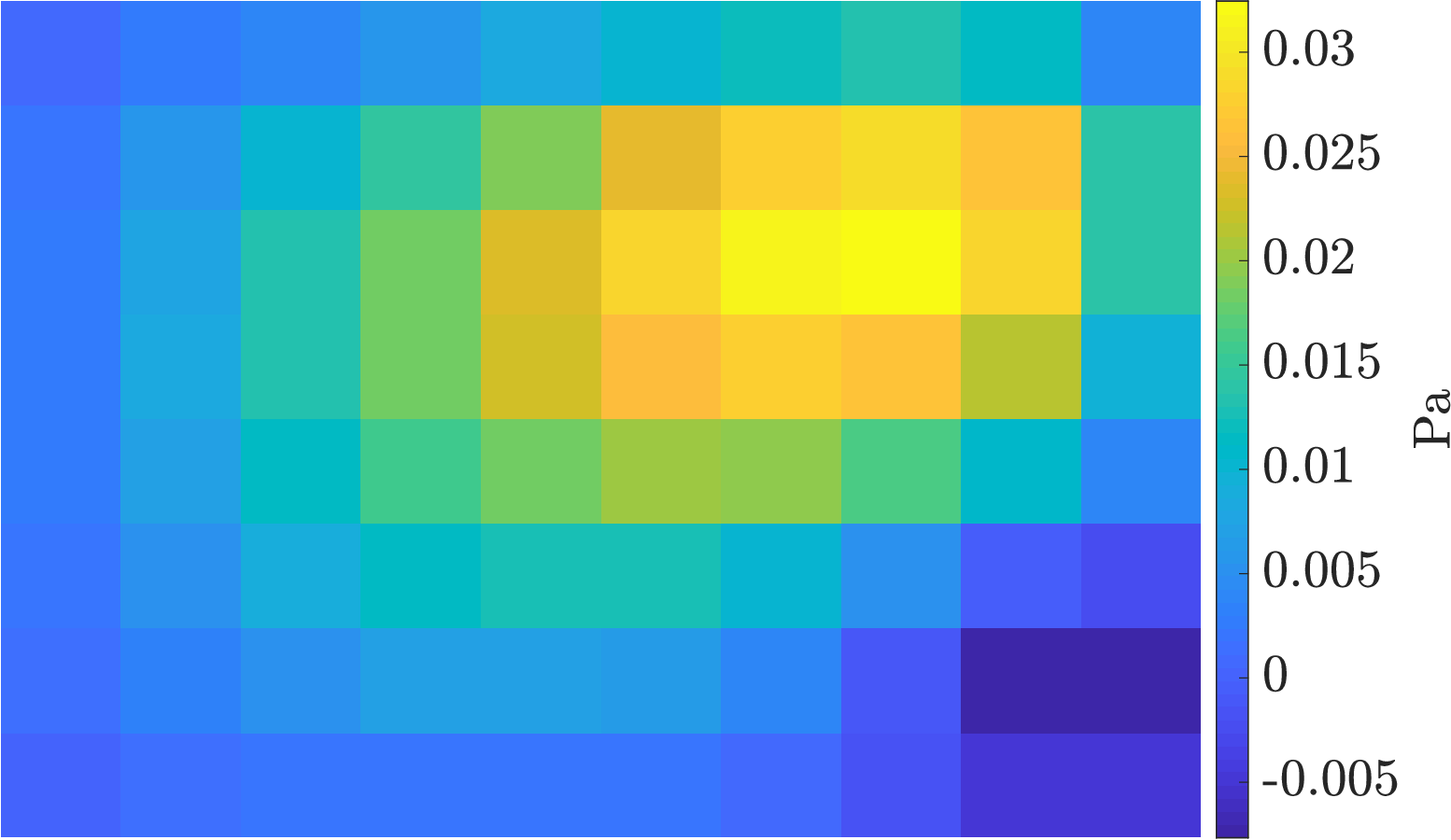}   %[Para su tamaño:anchura con respecto al ancho de la hoja]{Imágenes\Nombre_de_la_imagen}
    \caption{Predicted normal component ($\sigma_{xy}$).}
 \end{subfigure}
 \hfill
 \begin{subfigure}[t]{0.495\linewidth}
    \centering %Para centrar la imagen
    \includegraphics[scale=0.3]{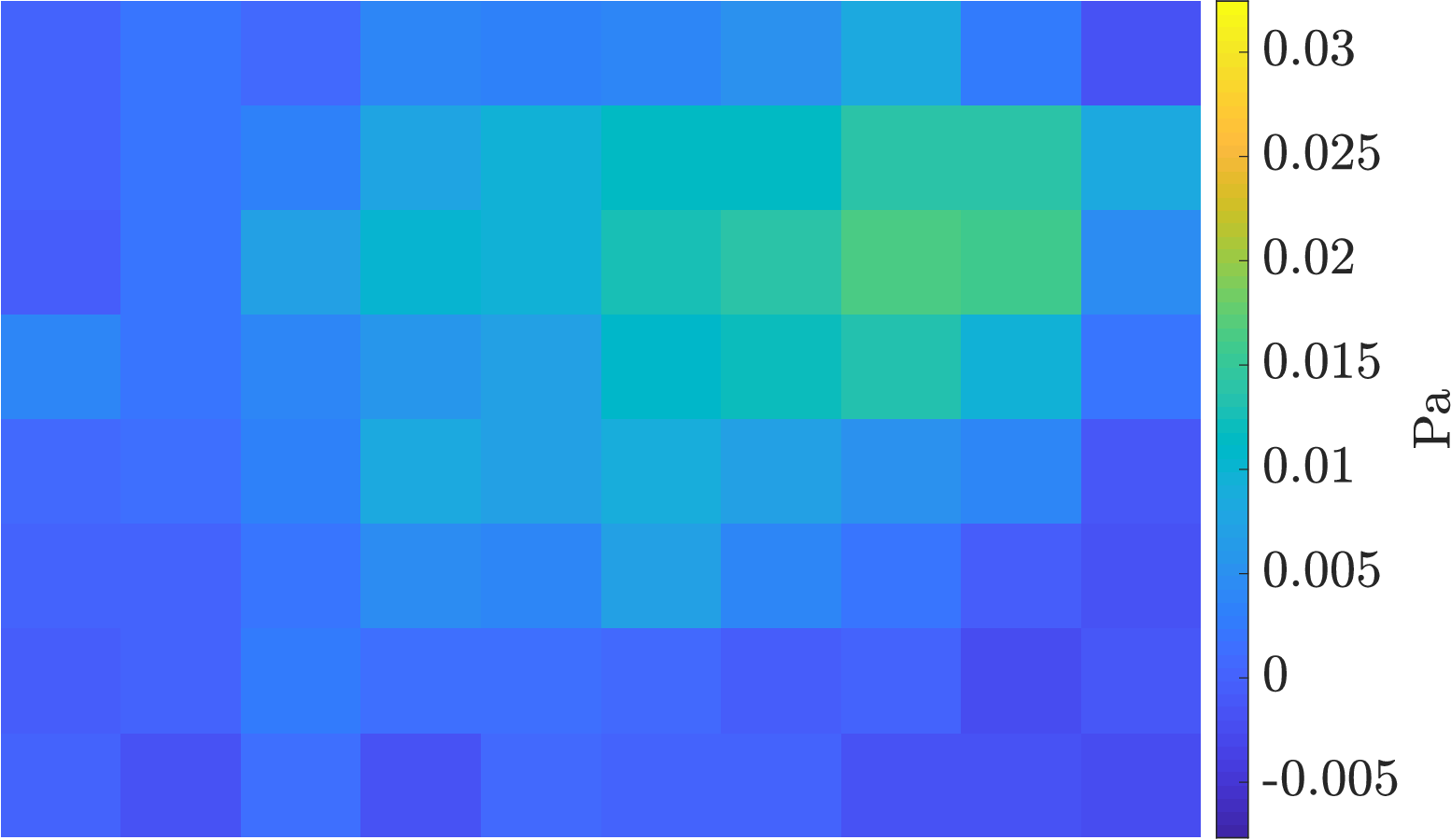}   %[Para su tamaño:anchura con respecto al ancho de la hoja]{Imágenes\Nombre_de_la_imagen}
    \caption{Predicted shear component ($\sigma_{xy}$).}
 \end{subfigure}
\caption{\textbf{PGNNIV prediction versus FEM solution of the components of the stress fields for a single test-set example of the linear material.}}
\label{lins}
\end{figure}

In Fig. \ref{linu} the displacement field (FEM solution versus PGNNIV prediction) is represented for a test example. In Fig. \ref{line} the components of the strain tensor are illustrated and in Fig. \ref{lins} the components of the stress tensor. We observe high similarity between the ground truth and predictive fields despite the coarse discretization.

%\clearpage

\subsubsection{Softening material.}

\begin{figure}[htbp]
 \centering
 \begin{subfigure}[t]{0.495\linewidth}
     \centering %Para centrar la imagen
    \includegraphics[scale=0.3]{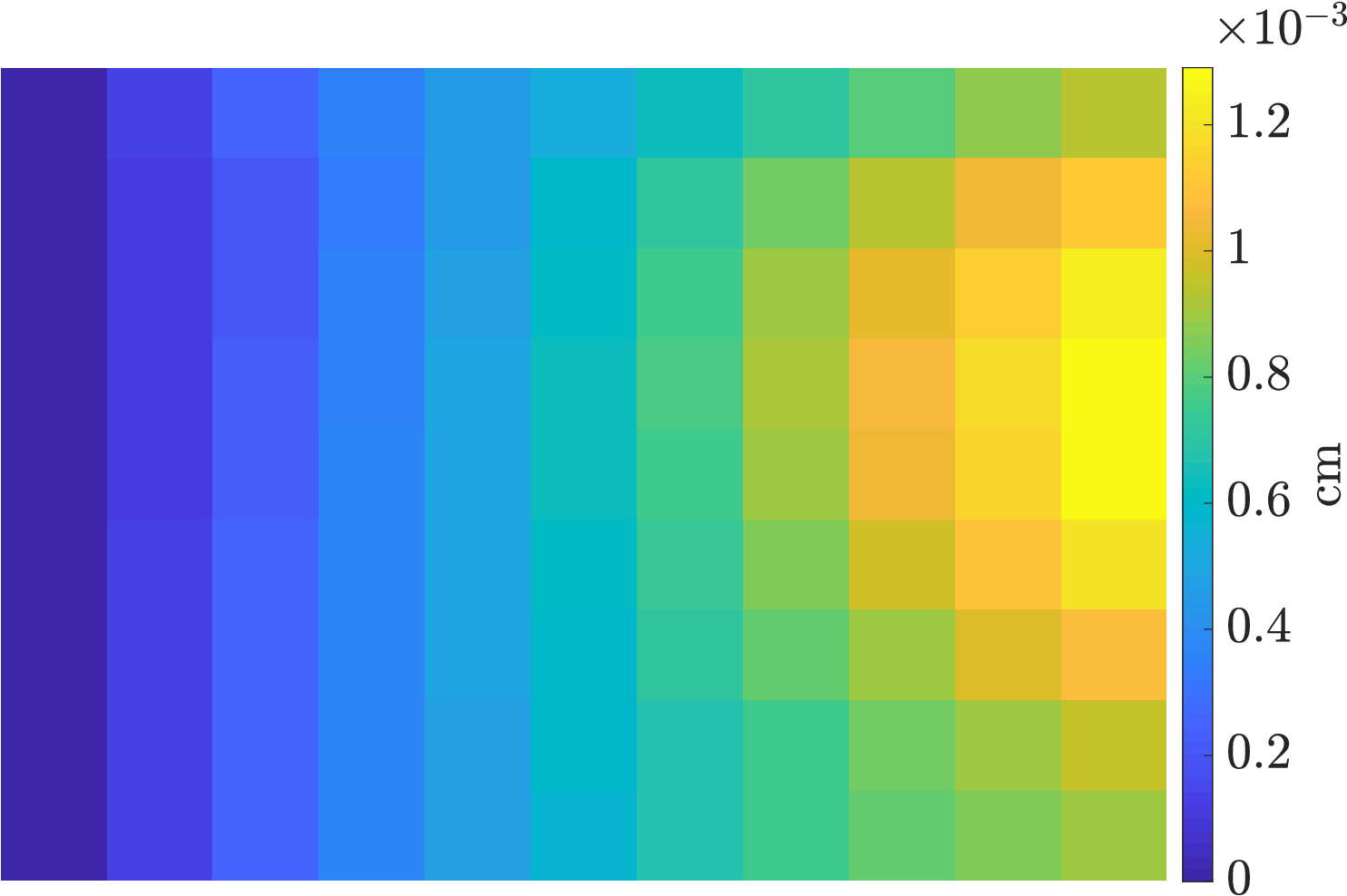}   %[Para su tamaño:anchura con respecto al ancho de la hoja]{Imágenes\Nombre_de_la_imagen}
    \caption{Real horizontal component ($u_x$).}
 \end{subfigure}
 \hfill
 \begin{subfigure}[t]{0.495\linewidth}
    \centering %Para centrar la imagen
    \includegraphics[scale=0.3]{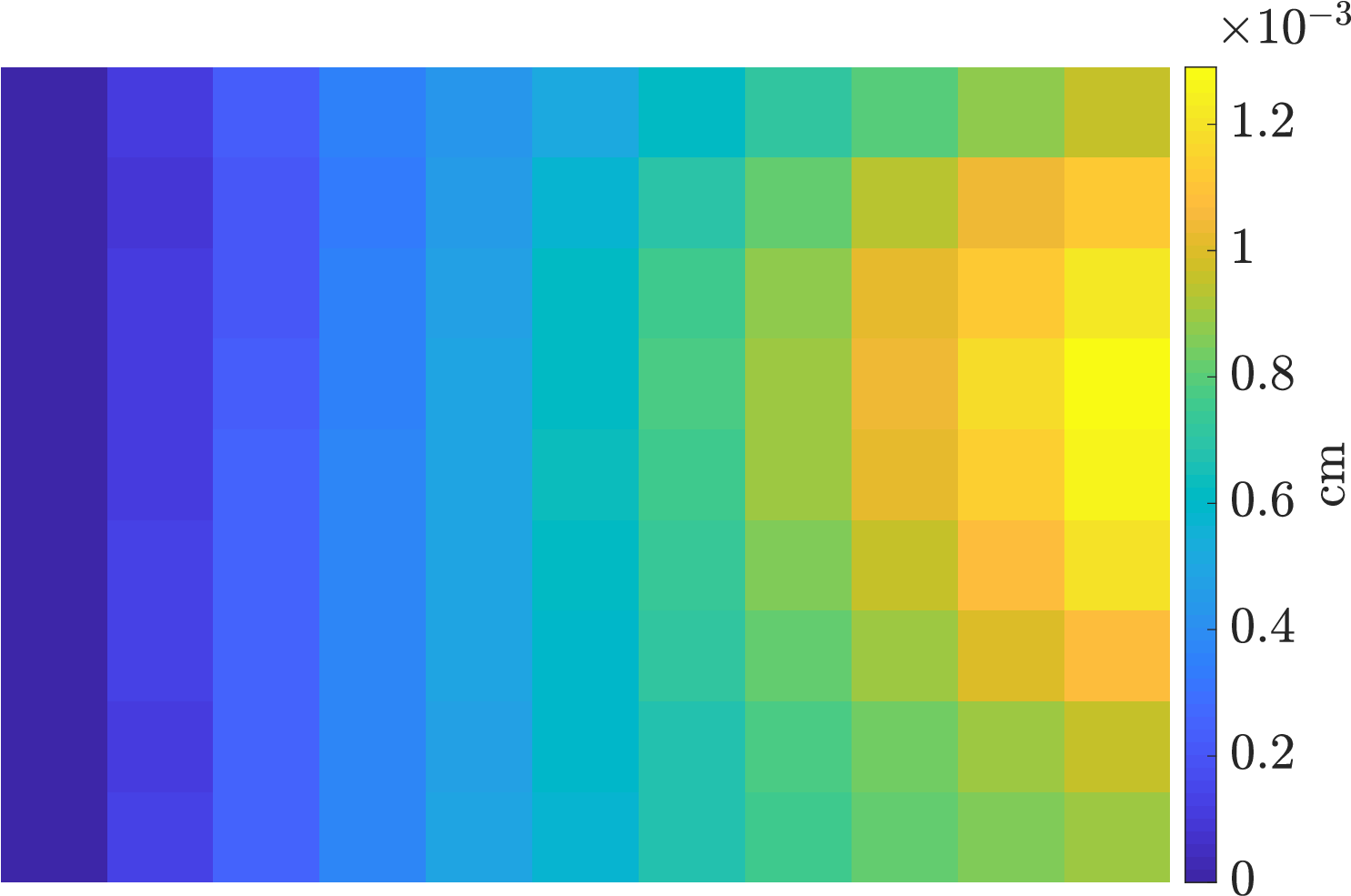}   %[Para su tamaño:anchura con respecto al ancho de la hoja]{Imágenes\Nombre_de_la_imagen}
    \caption{Predicted horizontal component ($u_x$).}
 \end{subfigure}
 \begin{subfigure}[t]{0.495\linewidth}
     \centering %Para centrar la imagen
    \includegraphics[scale=0.3]{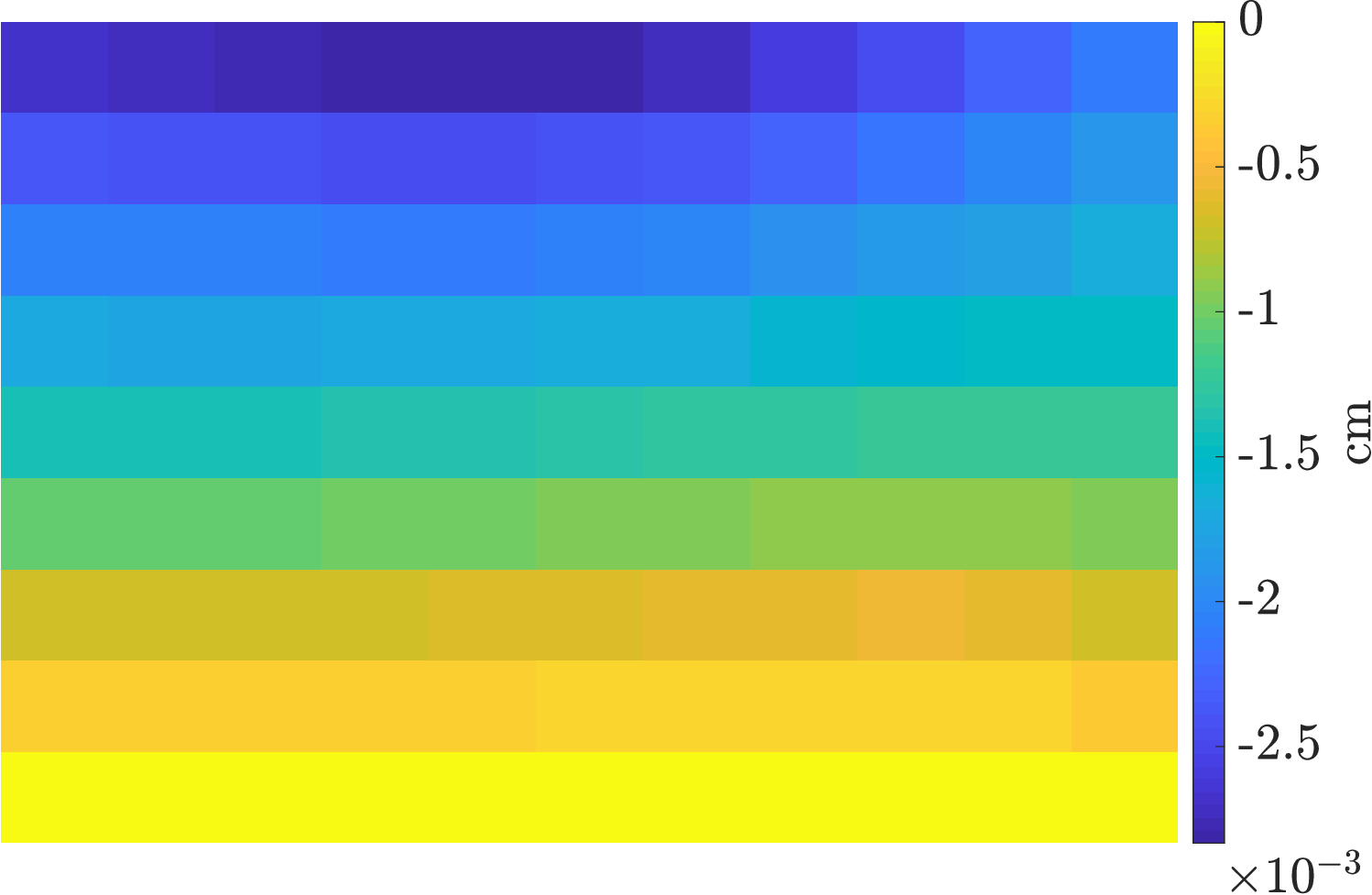}   %[Para su tamaño:anchura con respecto al ancho de la hoja]{Imágenes\Nombre_de_la_imagen}
    \caption{Real vertical component ($u_y$).}
 \end{subfigure}
 \hfill
 \begin{subfigure}[t]{0.495\linewidth}
    \centering %Para centrar la imagen
    \includegraphics[scale=0.3]{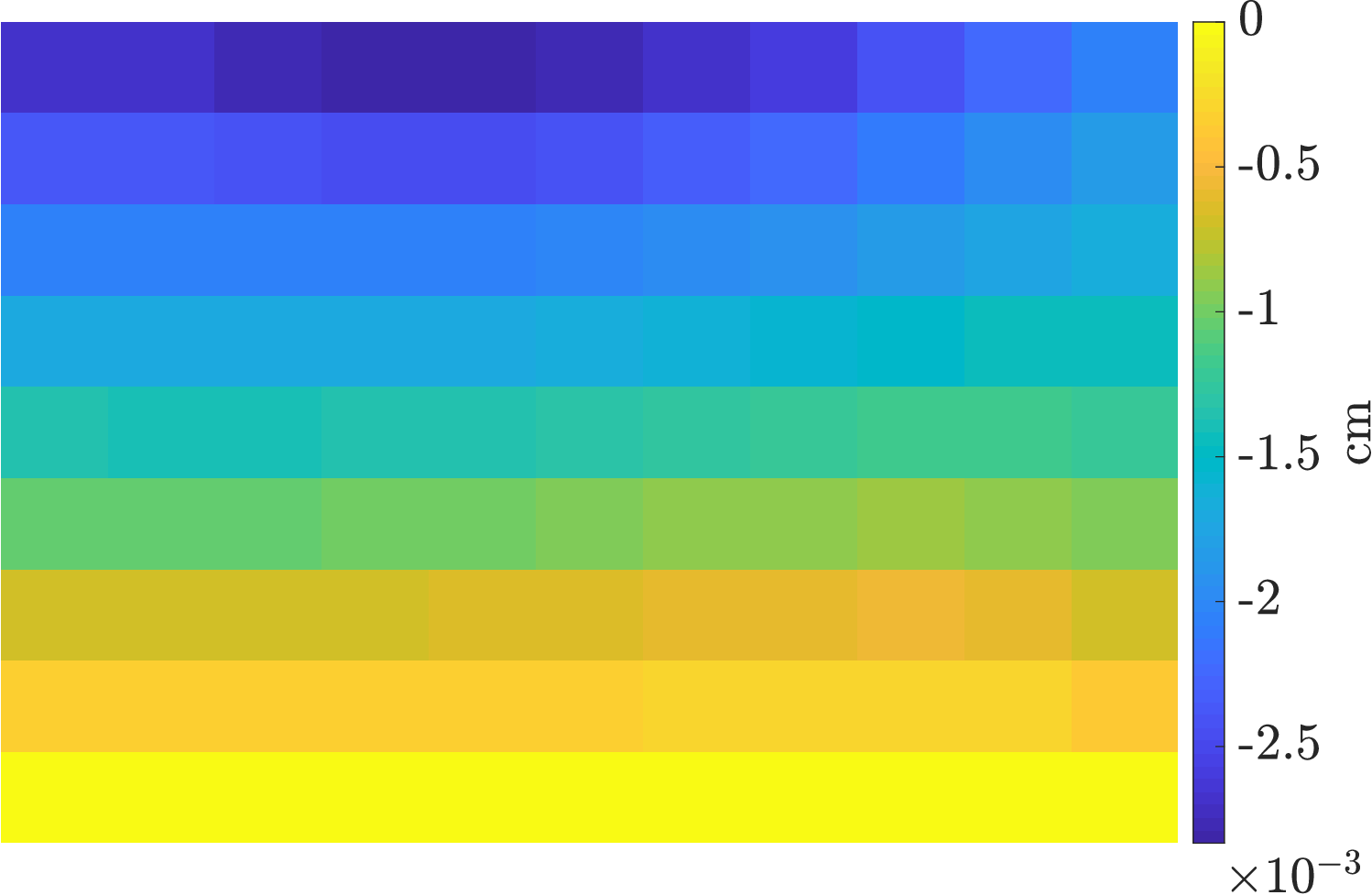}   %[Para su tamaño:anchura con respecto al ancho de la hoja]{Imágenes\Nombre_de_la_imagen}
    \caption{Predicted vertical component ($u_y$).}
 \end{subfigure}
\caption{\textbf{PGNNIV prediction versus FEM solution of the components of the displacement field for a single test-set example of the softening material.}}
\label{softu}
\end{figure}

\begin{figure}[htbp]
 \centering
 \begin{subfigure}[t]{0.495\linewidth}
     \centering %Para centrar la imagen
    \includegraphics[scale=0.30]{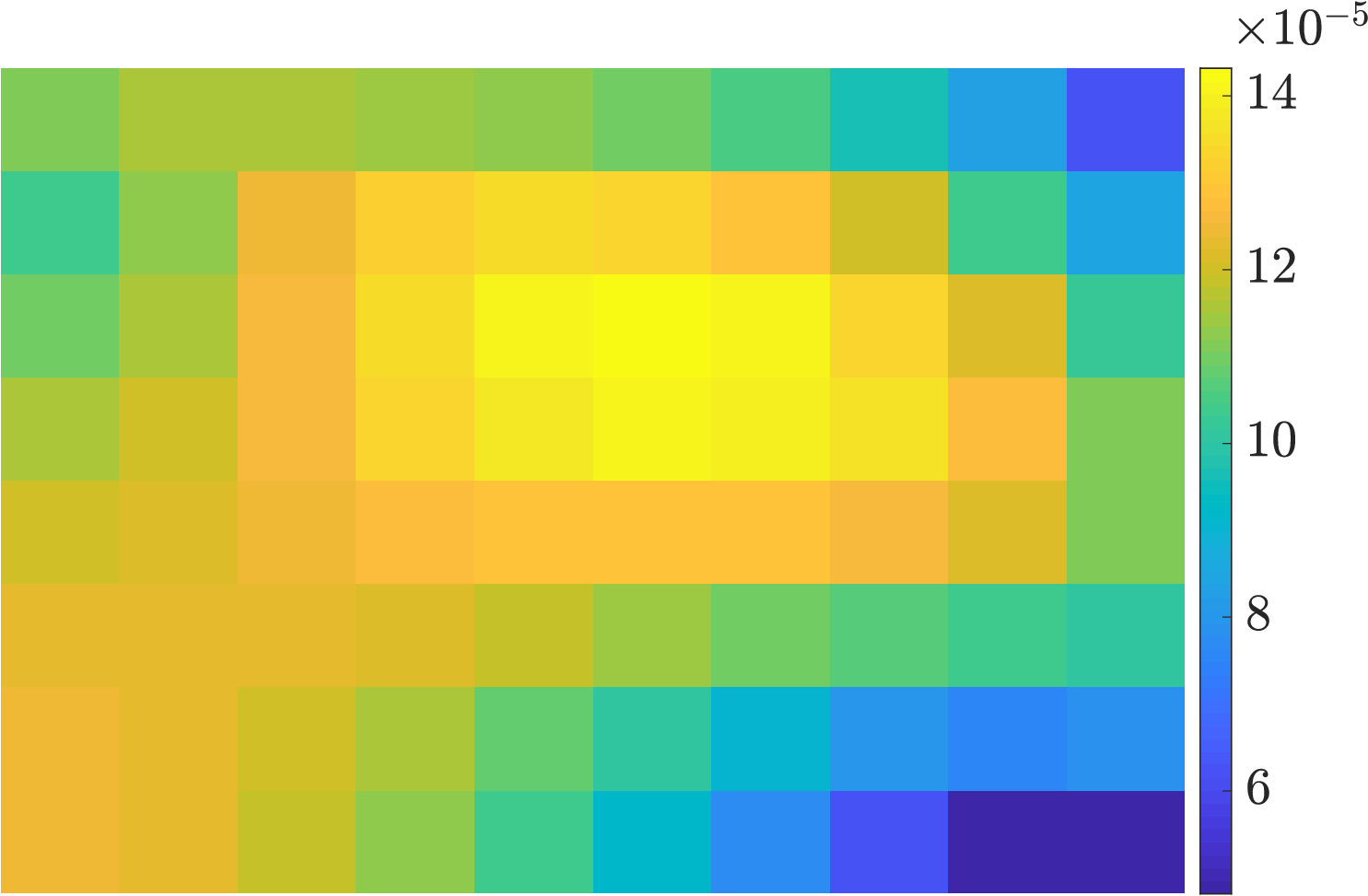}   %[Para su tamaño:anchura con respecto al ancho de la hoja]{Imágenes\Nombre_de_la_imagen}
    \caption{Real normal component ($\varepsilon_{xx}$).}
 \end{subfigure}
 \hfill
 \begin{subfigure}[t]{0.495\linewidth}
     \centering %Para centrar la imagen
    \includegraphics[scale=0.30]{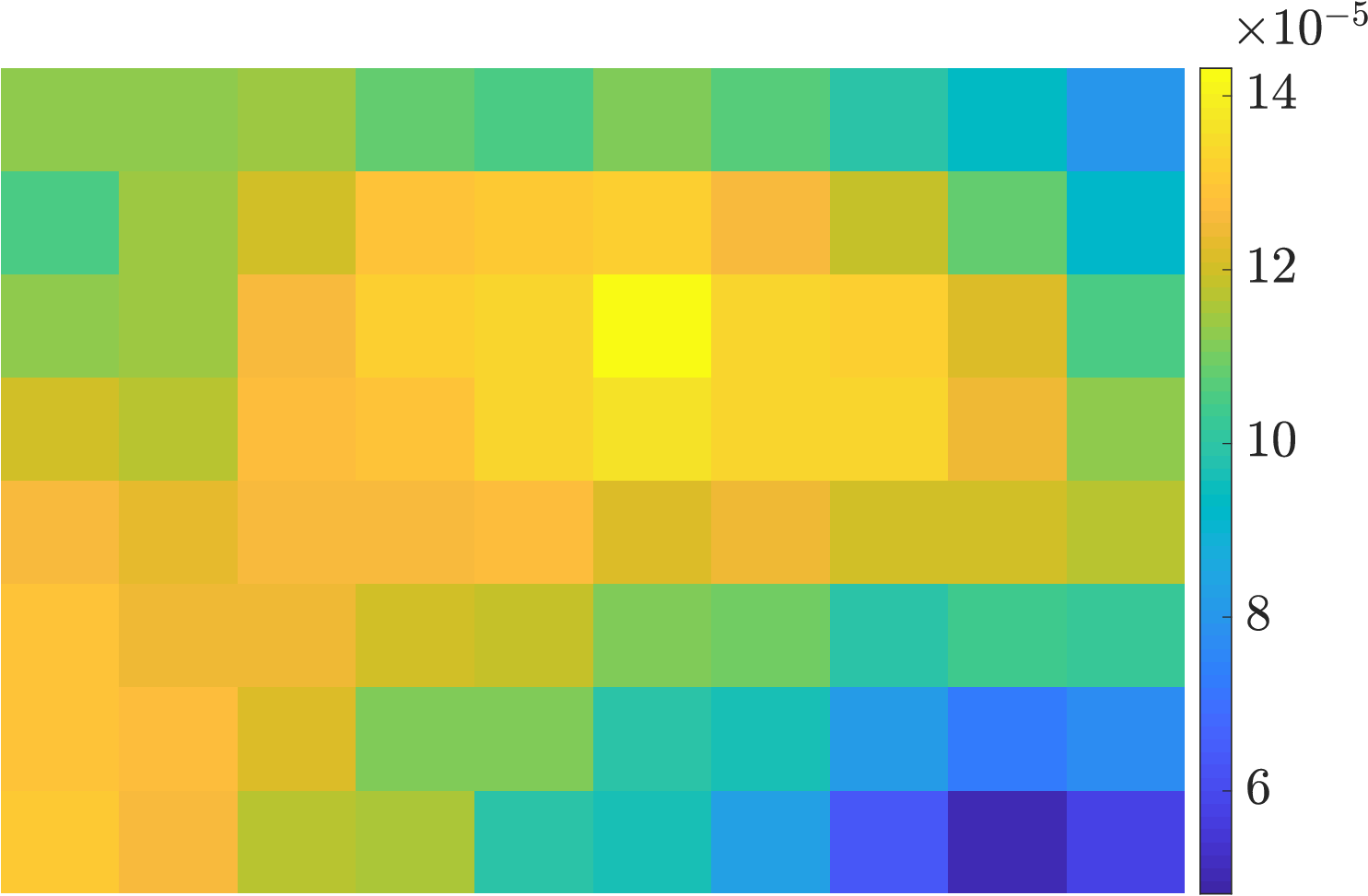}   %[Para su tamaño:anchura con respecto al ancho de la hoja]{Imágenes\Nombre_de_la_imagen}
    \caption{Predicted normal component ($\varepsilon_{xx}$).}
 \end{subfigure}
 \hfill
 \begin{subfigure}[t]{0.495\linewidth}
    \centering %Para centrar la imagen
    \includegraphics[scale=0.30]{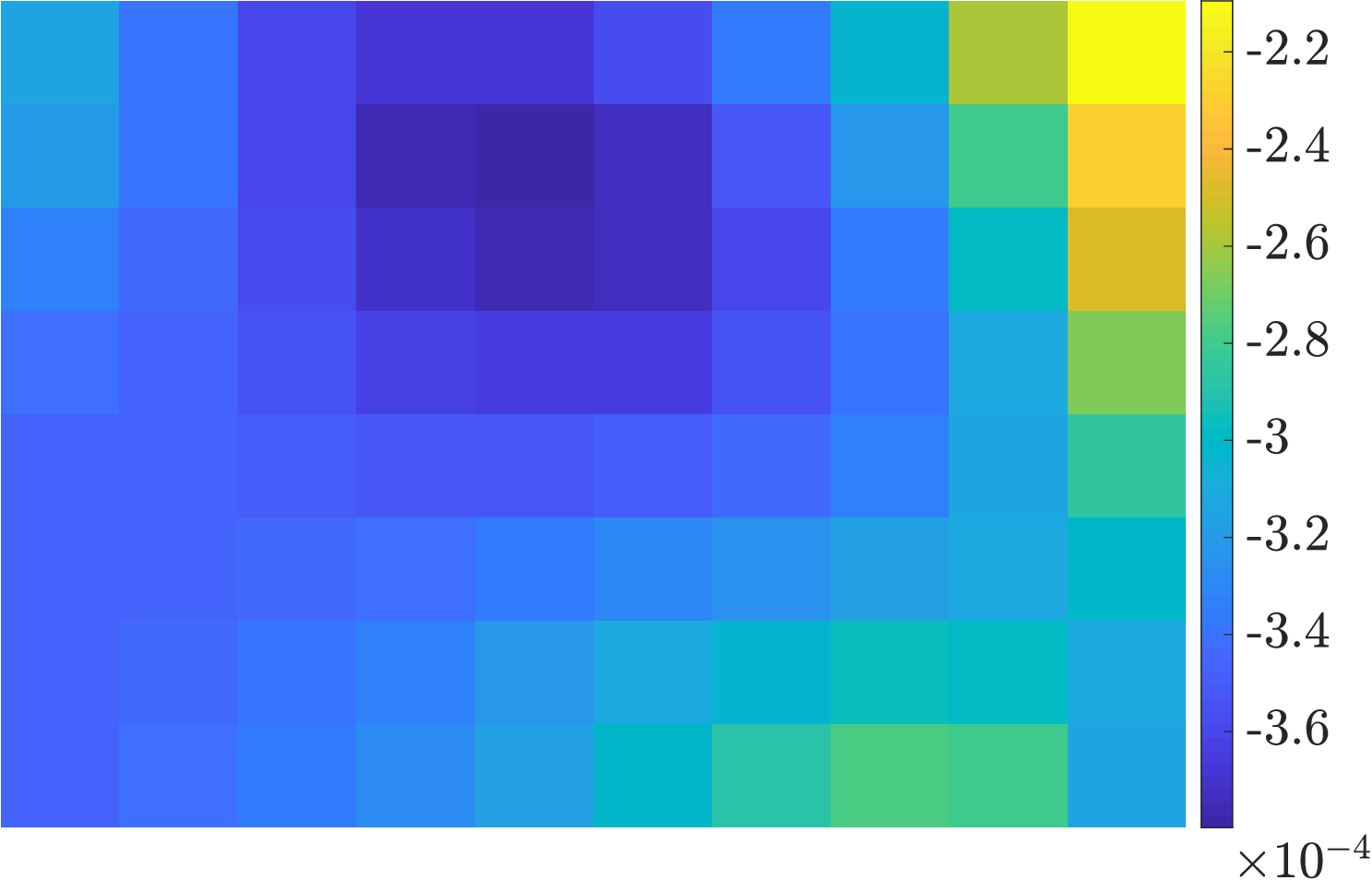}   %[Para su tamaño:anchura con respecto al ancho de la hoja]{Imágenes\Nombre_de_la_imagen}
    \caption{Real normal component ($\varepsilon_{yy}$).}
 \end{subfigure}
  \begin{subfigure}[t]{0.495\linewidth}
     \centering %Para centrar la imagen
    \includegraphics[scale=0.30]{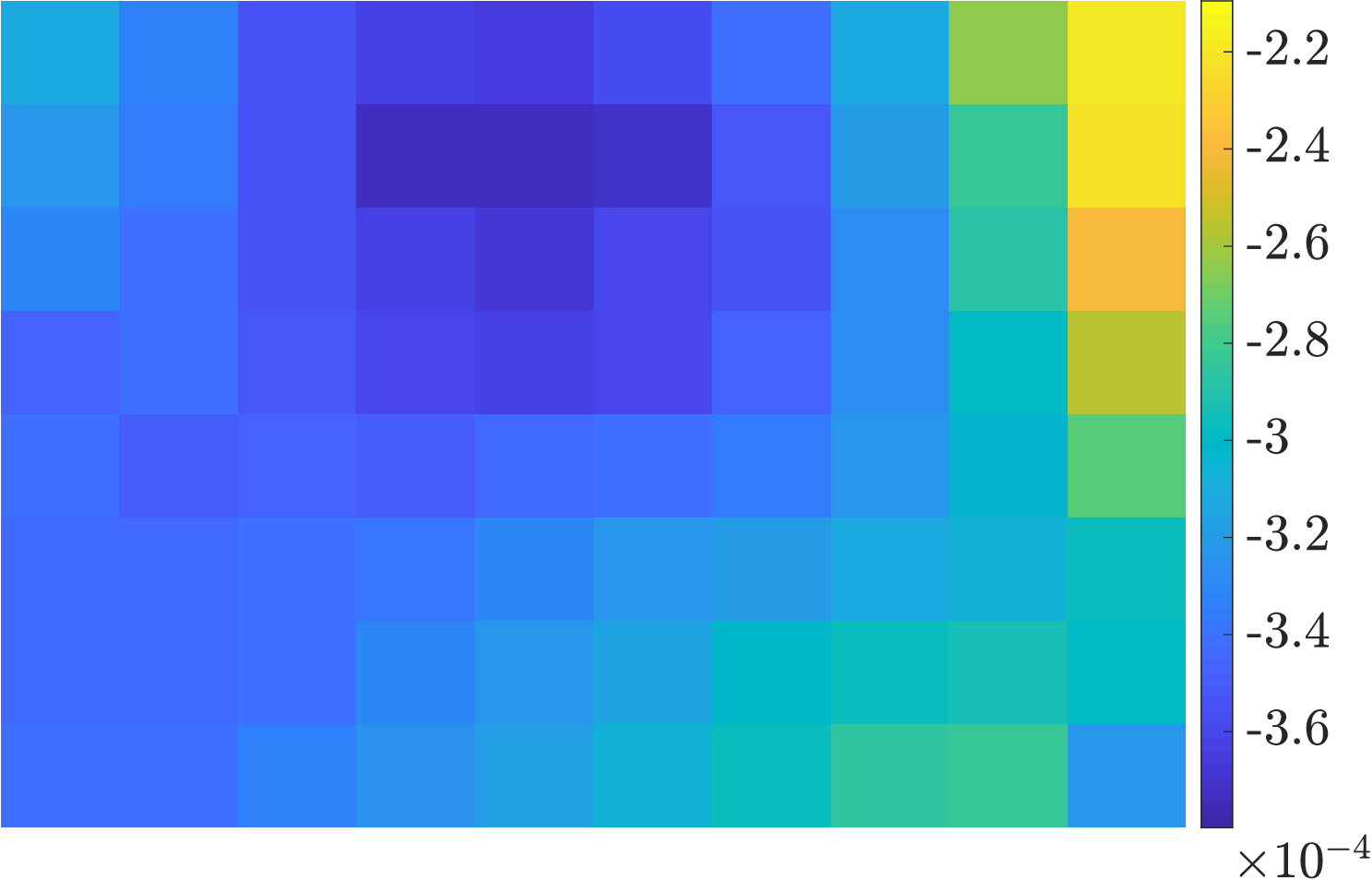}   %[Para su tamaño:anchura con respecto al ancho de la hoja]{Imágenes\Nombre_de_la_imagen}
    \caption{Predicted normal component ($\varepsilon_{yy}$).}
 \end{subfigure}
 \hfill
 \begin{subfigure}[t]{0.495\linewidth}
     \centering %Para centrar la imagen
    \includegraphics[scale=0.30]{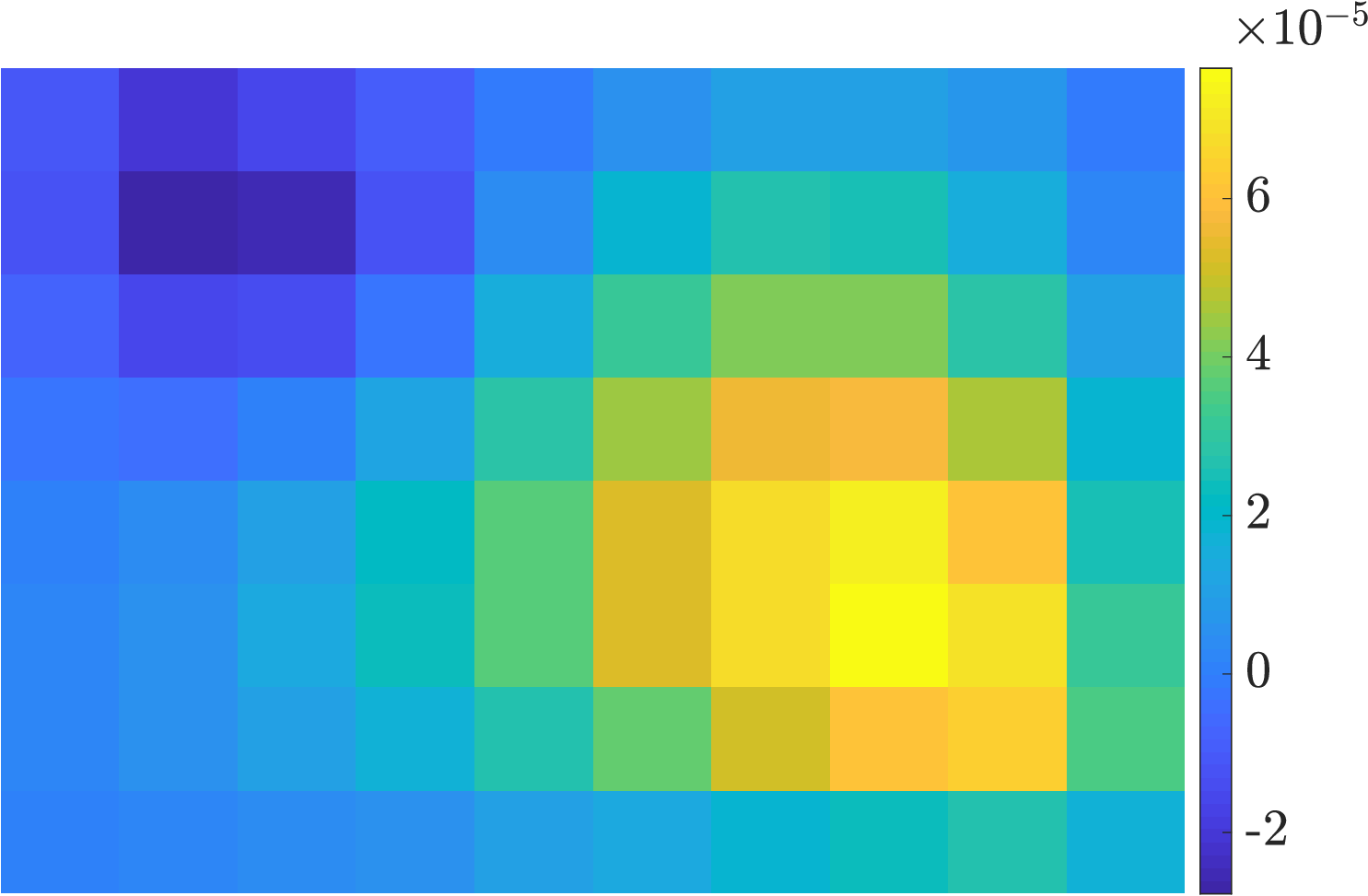}   %[Para su tamaño:anchura con respecto al ancho de la hoja]{Imágenes\Nombre_de_la_imagen}
    \caption{Real shear component ($\varepsilon_{xy}$).}
 \end{subfigure}
 \hfill
 \begin{subfigure}[t]{0.495\linewidth}
    \centering %Para centrar la imagen
    \includegraphics[scale=0.30]{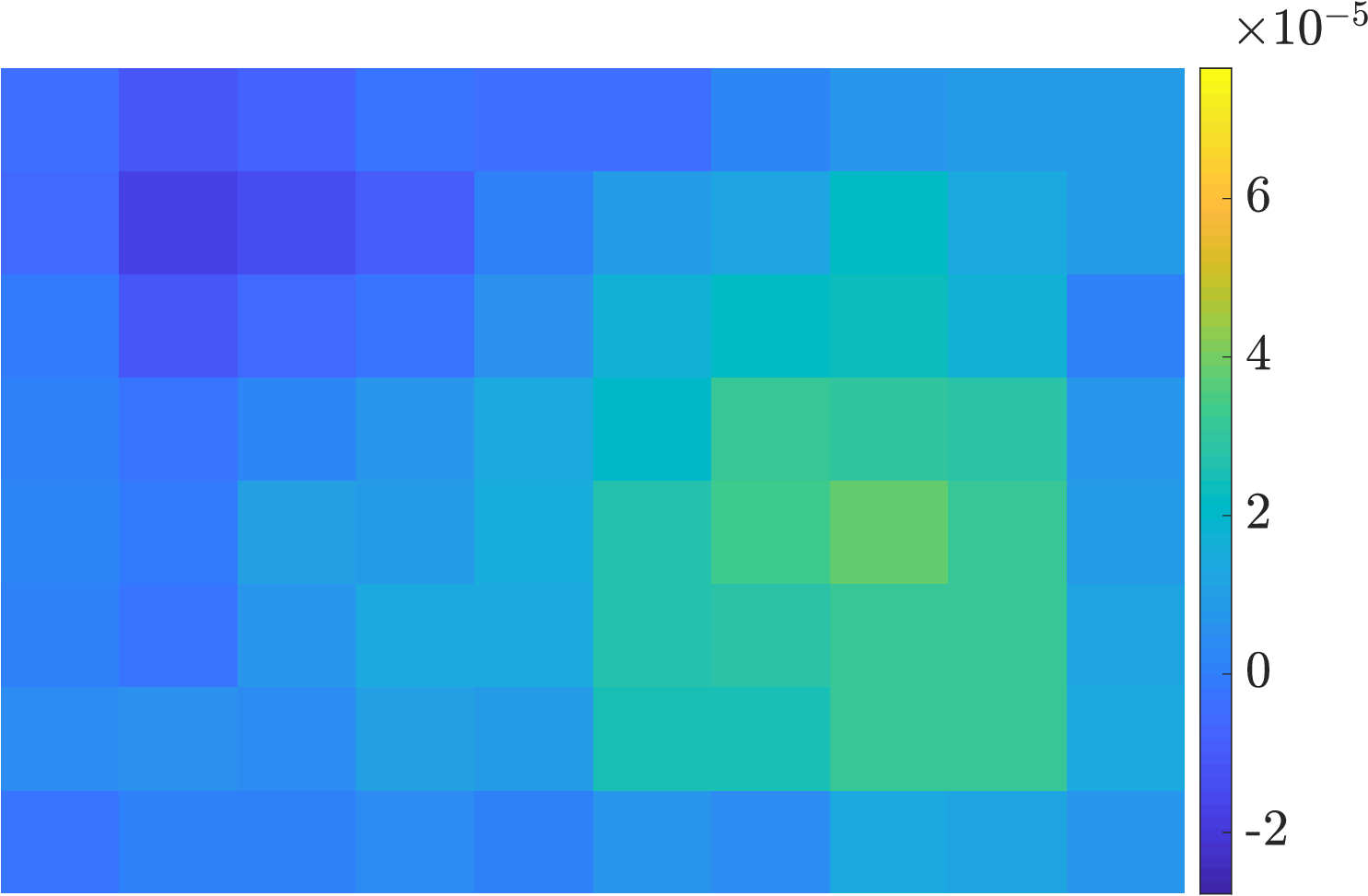}   %[Para su tamaño:anchura con respecto al ancho de la hoja]{Imágenes\Nombre_de_la_imagen}
    \caption{Predicted shear component ($\varepsilon_{xy}$).}
 \end{subfigure}
\caption{\textbf{PGNNIV prediction versus FEM solution of the components of the strain fields for a single test-set example of the softening material.}}
\label{softe}
\end{figure}

\begin{figure}[htbp]
 \centering
 \begin{subfigure}[t]{0.495\linewidth}
     \centering %Para centrar la imagen
    \includegraphics[scale=0.3]{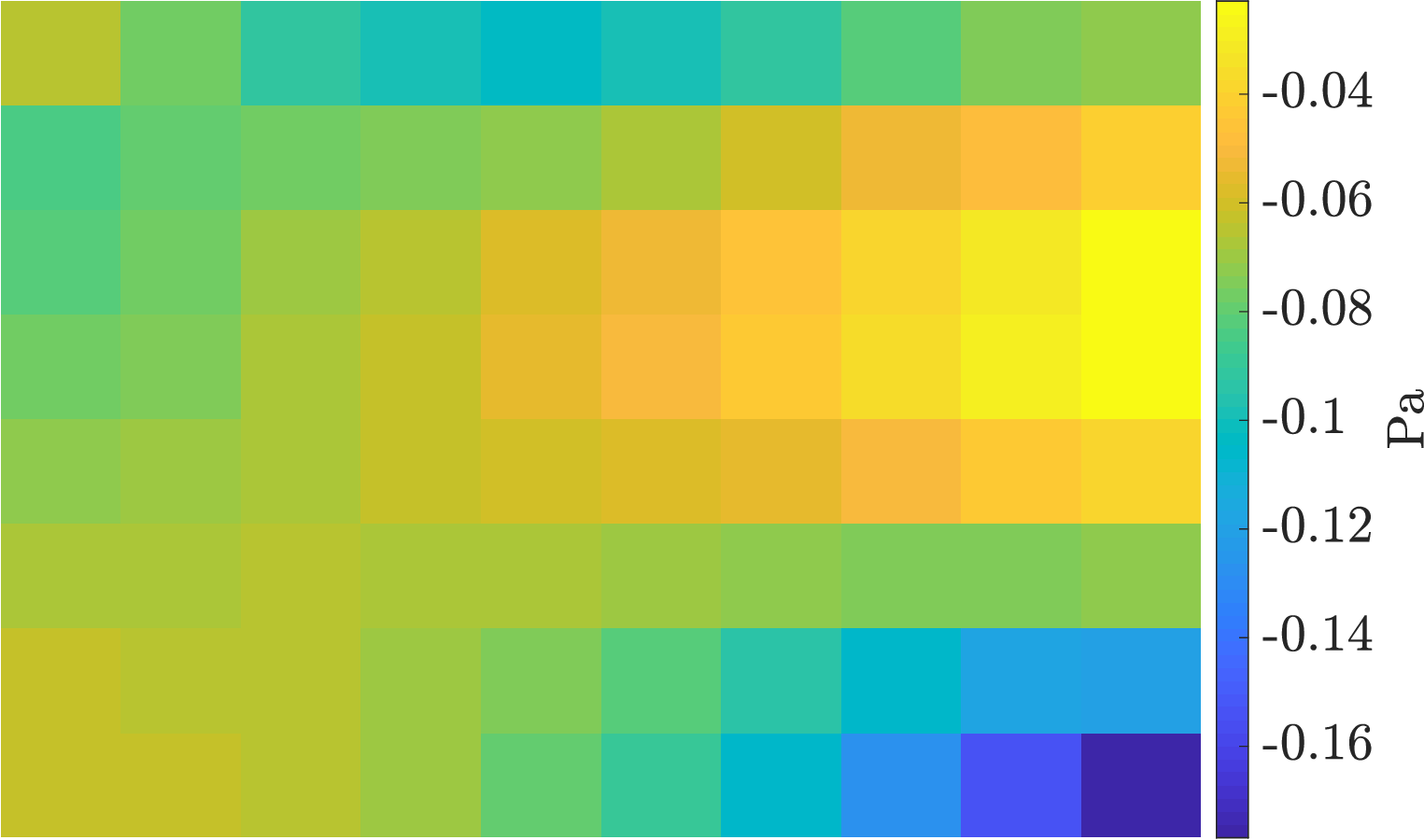}   %[Para su tamaño:anchura con respecto al ancho de la hoja]{Imágenes\Nombre_de_la_imagen}
    \caption{Real normal component ($\sigma_{xx}$).}
 \end{subfigure}
 \hfill
 \begin{subfigure}[t]{0.495\linewidth}
     \centering %Para centrar la imagen
    \includegraphics[scale=0.3]{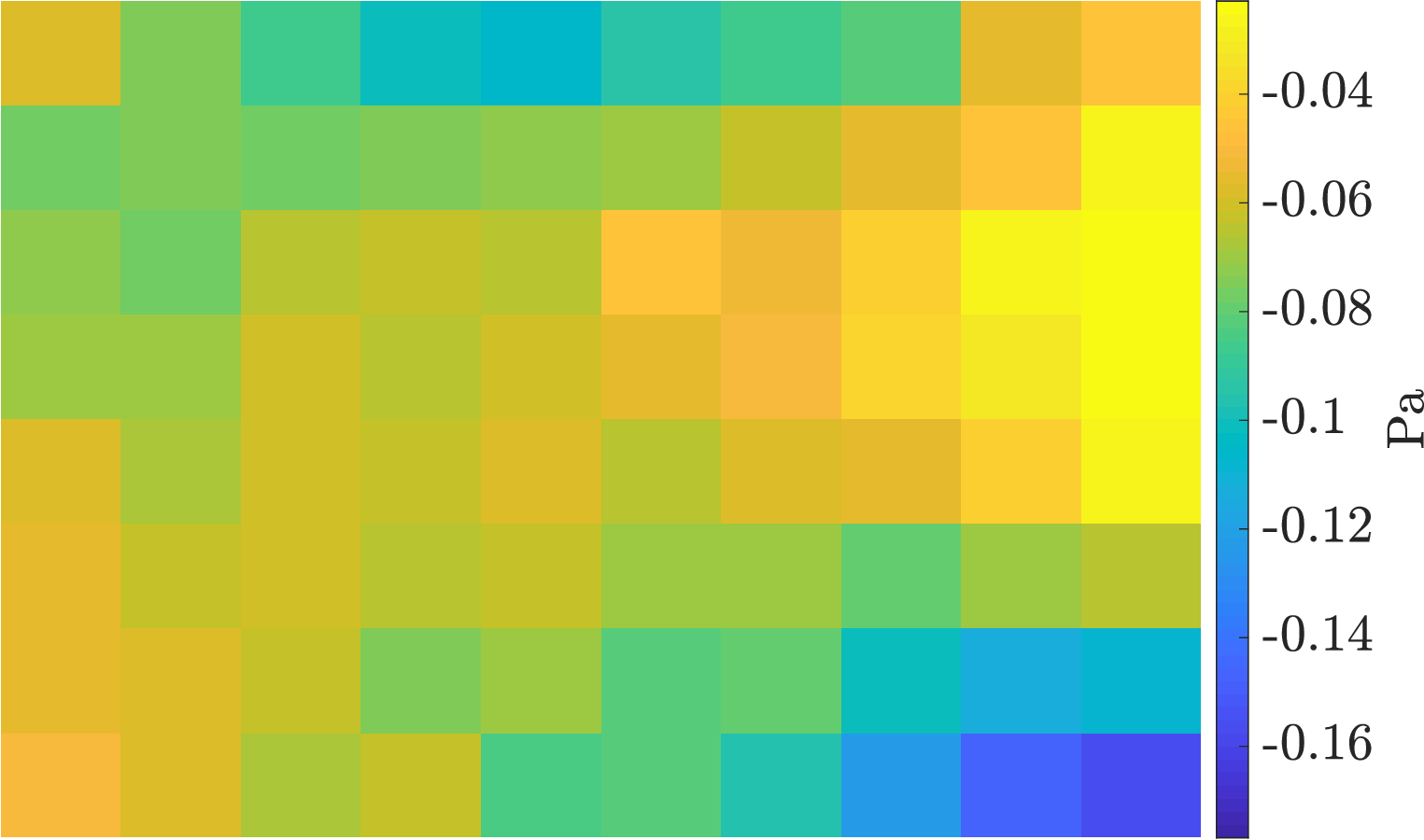}   %[Para su tamaño:anchura con respecto al ancho de la hoja]{Imágenes\Nombre_de_la_imagen}
    \caption{Predicted normal component ($\sigma_{xx}$).}
 \end{subfigure}
 \hfill
 \begin{subfigure}[t]{0.495\linewidth}
    \centering %Para centrar la imagen
    \includegraphics[scale=0.3]{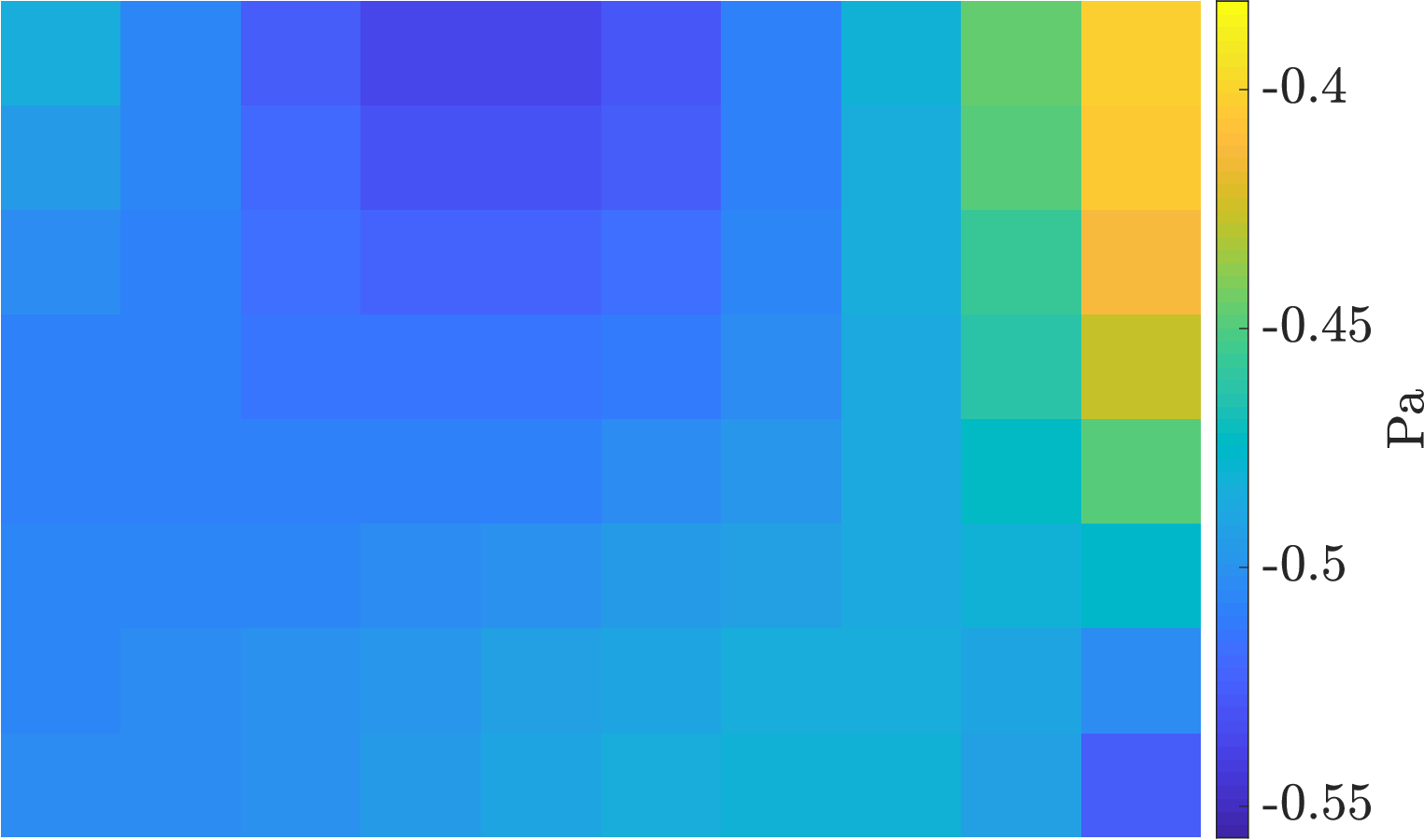}   %[Para su tamaño:anchura con respecto al ancho de la hoja]{Imágenes\Nombre_de_la_imagen}
    \caption{Real normal component ($\sigma_{yy}$).}
 \end{subfigure}
  \begin{subfigure}[t]{0.495\linewidth}
     \centering %Para centrar la imagen
    \includegraphics[scale=0.3]{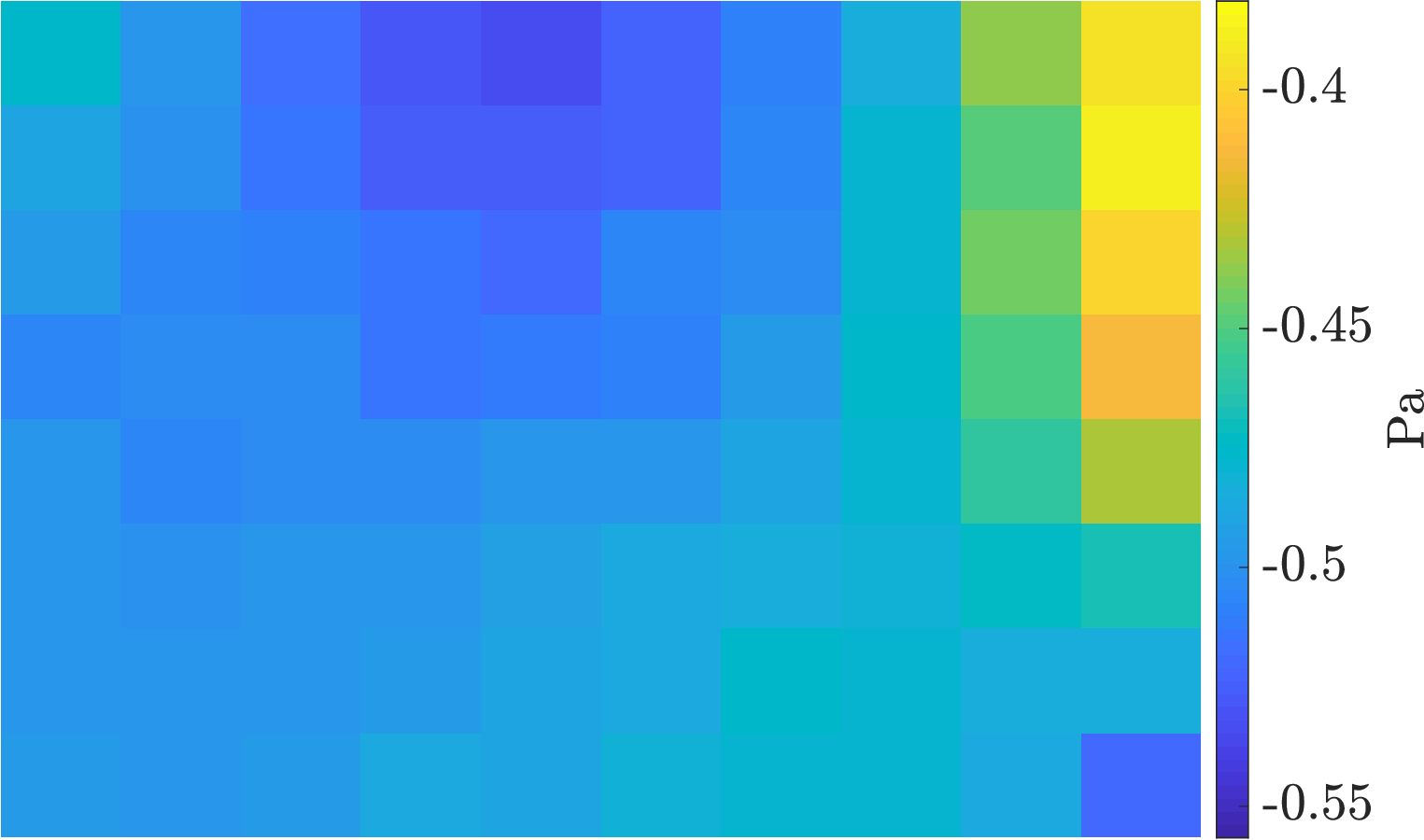}   %[Para su tamaño:anchura con respecto al ancho de la hoja]{Imágenes\Nombre_de_la_imagen}
    \caption{Predicted normal component ($\sigma_{yy}$).}
 \end{subfigure}
 \hfill
 \begin{subfigure}[t]{0.495\linewidth}
     \centering %Para centrar la imagen
    \includegraphics[scale=0.3]{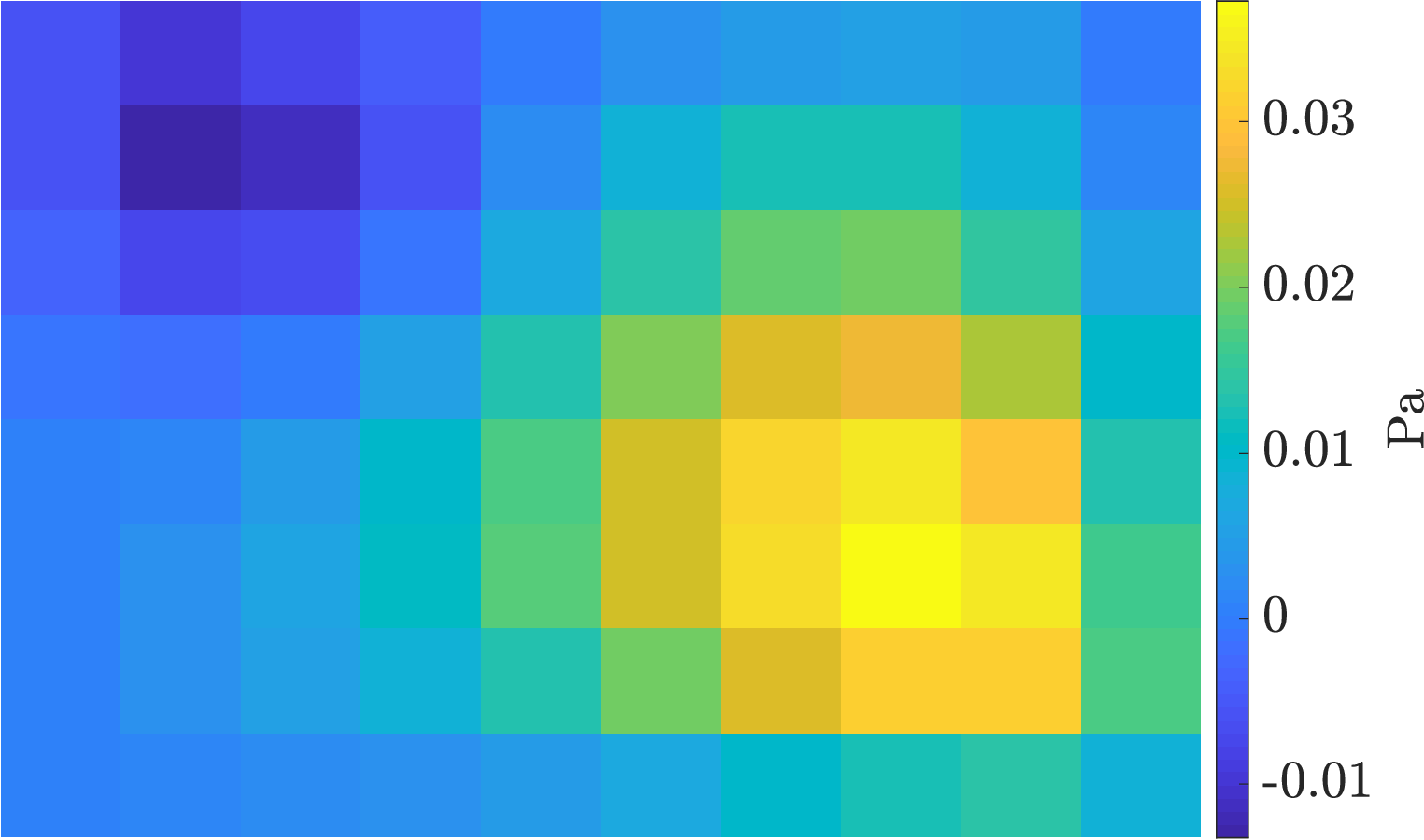}   %[Para su tamaño:anchura con respecto al ancho de la hoja]{Imágenes\Nombre_de_la_imagen}
    \caption{Real shear component ($\sigma_{xy}$).}
 \end{subfigure}
 \hfill
 \begin{subfigure}[t]{0.495\linewidth}
    \centering %Para centrar la imagen
    \includegraphics[scale=0.3]{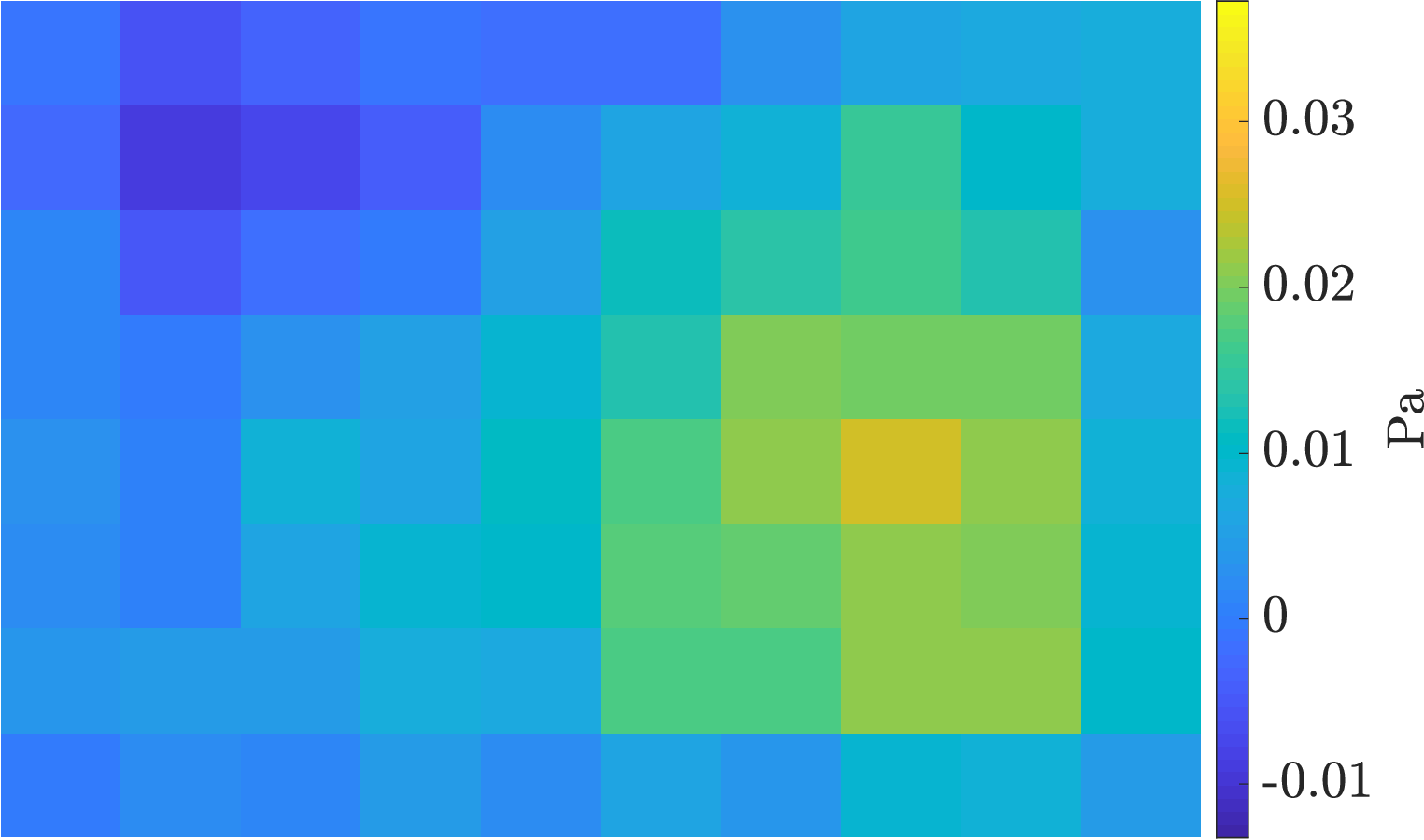}   %[Para su tamaño:anchura con respecto al ancho de la hoja]{Imágenes\Nombre_de_la_imagen}
    \caption{Predicted shear component ($\sigma_{xy}$).}
 \end{subfigure}
\caption{\textbf{PGNNIV prediction versus FEM solution of the components of the stress fields for a single test-set example of the softening material.}}
\label{softs}
\end{figure}

In Fig. \ref{softu} the displacement field (FEM solution versus PGNNIV prediction) is represented for a test example. In Fig. \ref{softe} the components of the strain tensor are illustrated and in Fig. \ref{softs} the components of the stress tensor. We observe high similarity between the ground truth and predictive fields despite the coarse discretization.

%\clearpage

\subsubsection{Hardening material.}

\begin{figure}[htbp]
 \centering
 \begin{subfigure}[t]{0.495\linewidth}
     \centering %Para centrar la imagen
    \includegraphics[scale=0.3]{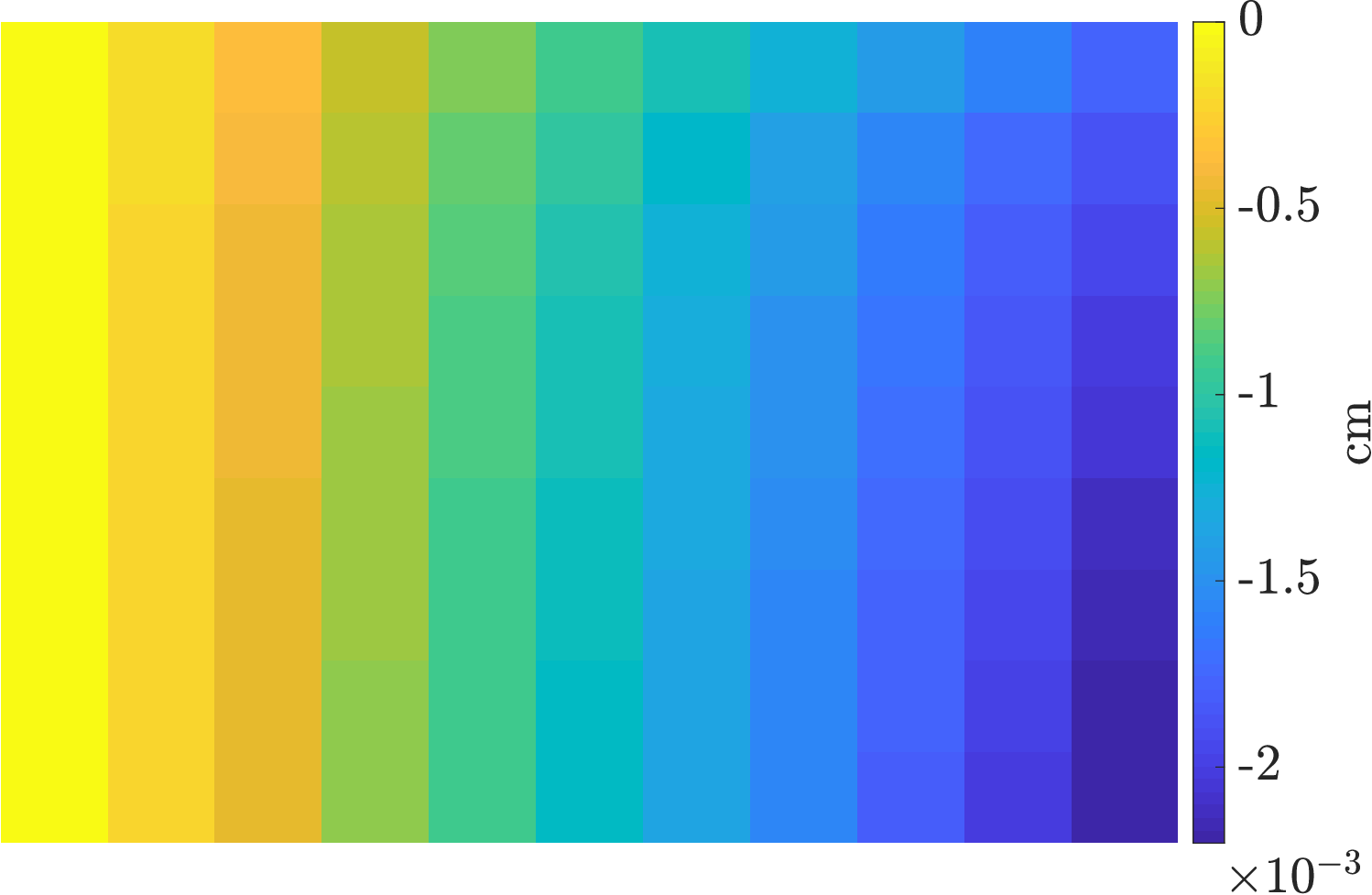}   %[Para su tamaño:anchura con respecto al ancho de la hoja]{Imágenes\Nombre_de_la_imagen}
    \caption{Real horizontal component ($u_x$).}
 \end{subfigure}
 \hfill
 \begin{subfigure}[t]{0.495\linewidth}
    \centering %Para centrar la imagen
    \includegraphics[scale=0.3]{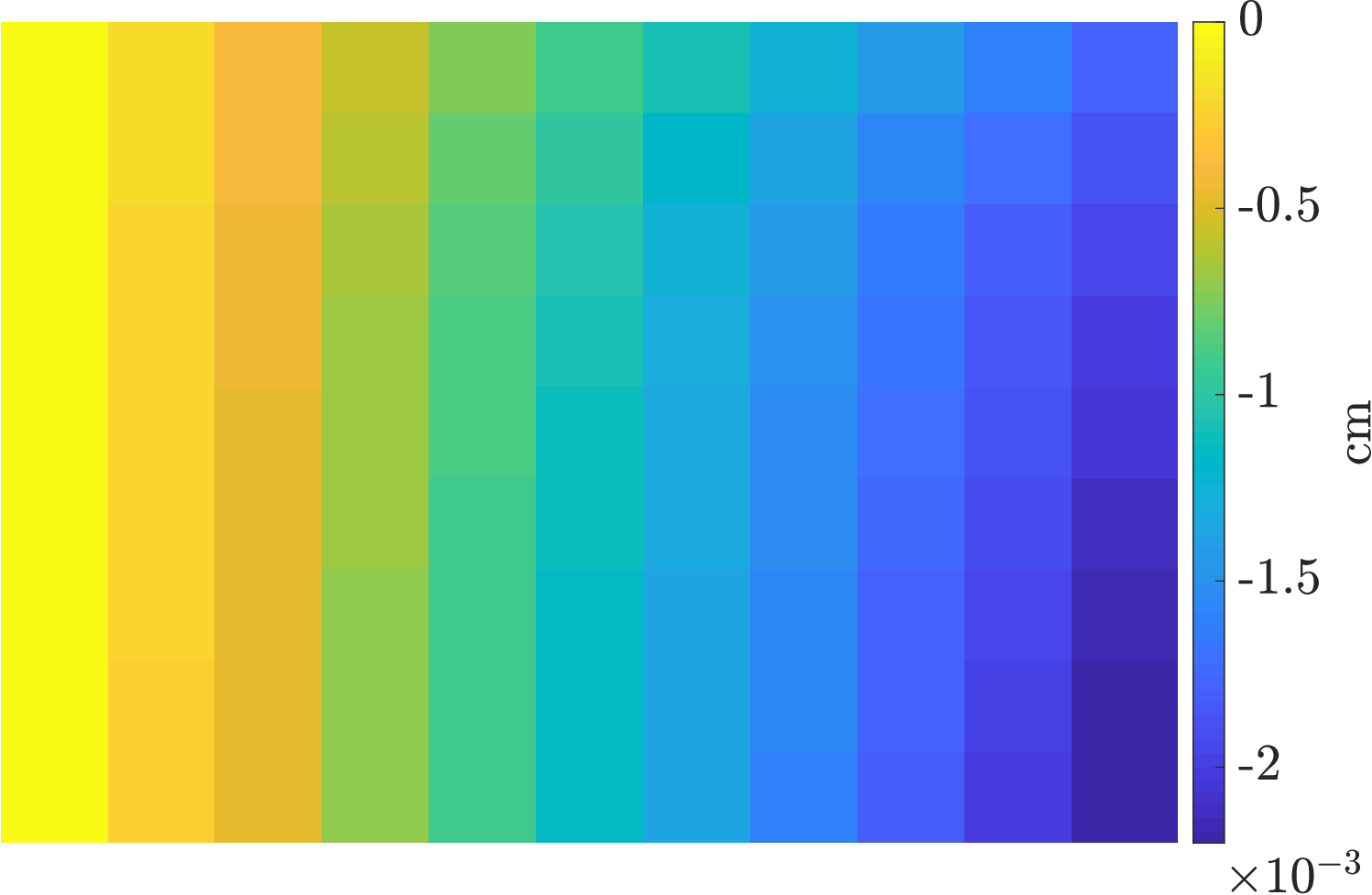}   %[Para su tamaño:anchura con respecto al ancho de la hoja]{Imágenes\Nombre_de_la_imagen}
    \caption{Predicted horizontal component ($u_x$).}
    \label{fig:app_softening_uy}   %Nombre para referirse a la imagen
 \end{subfigure}
 \begin{subfigure}[t]{0.495\linewidth}
     \centering %Para centrar la imagen
    \includegraphics[scale=0.3]{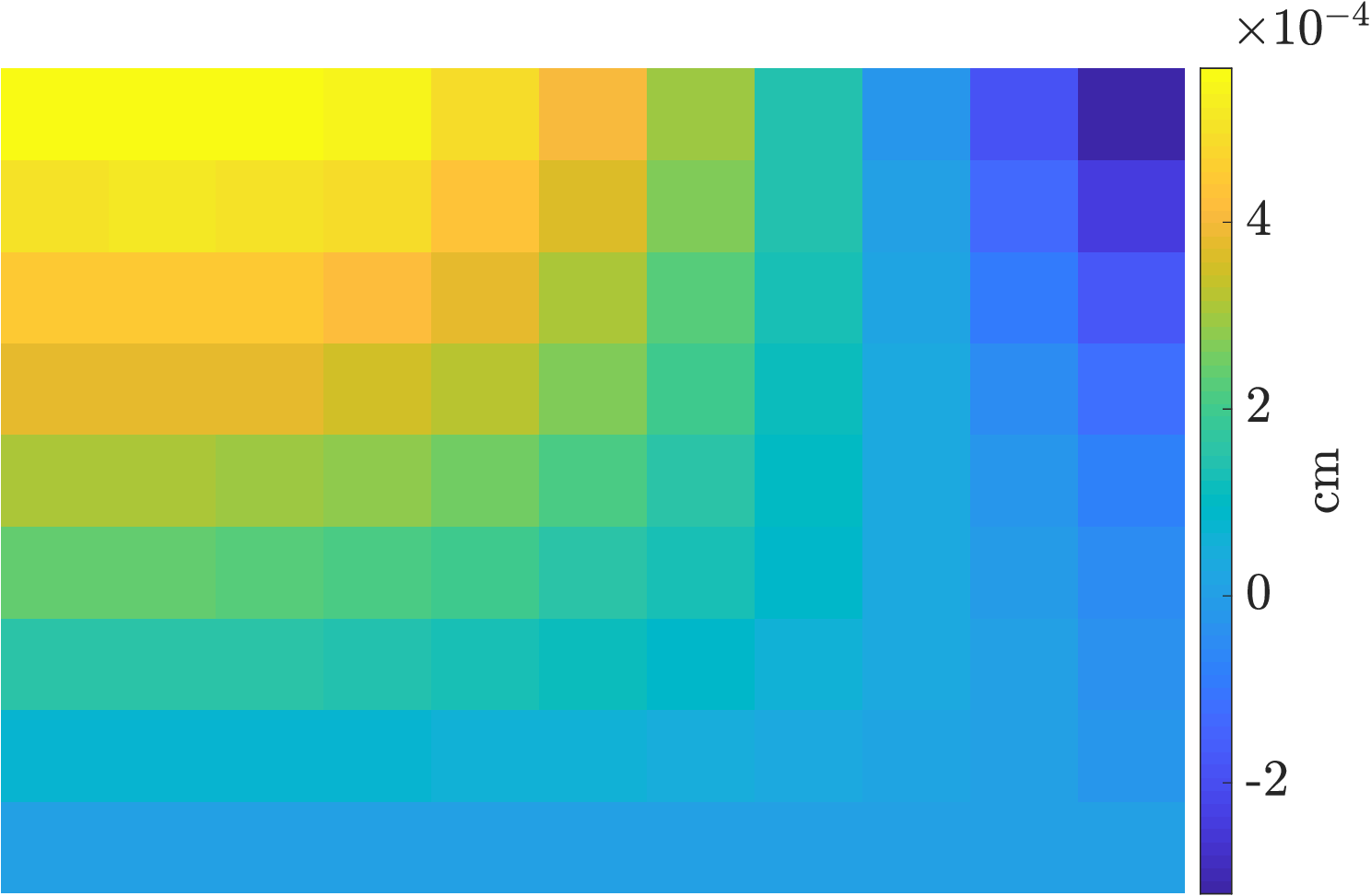}   %[Para su tamaño:anchura con respecto al ancho de la hoja]{Imágenes\Nombre_de_la_imagen}
    \caption{Real vertical component ($u_y$).}
 \end{subfigure}
 \hfill
 \begin{subfigure}[t]{0.495\linewidth}
    \centering %Para centrar la imagen
    \includegraphics[scale=0.3]{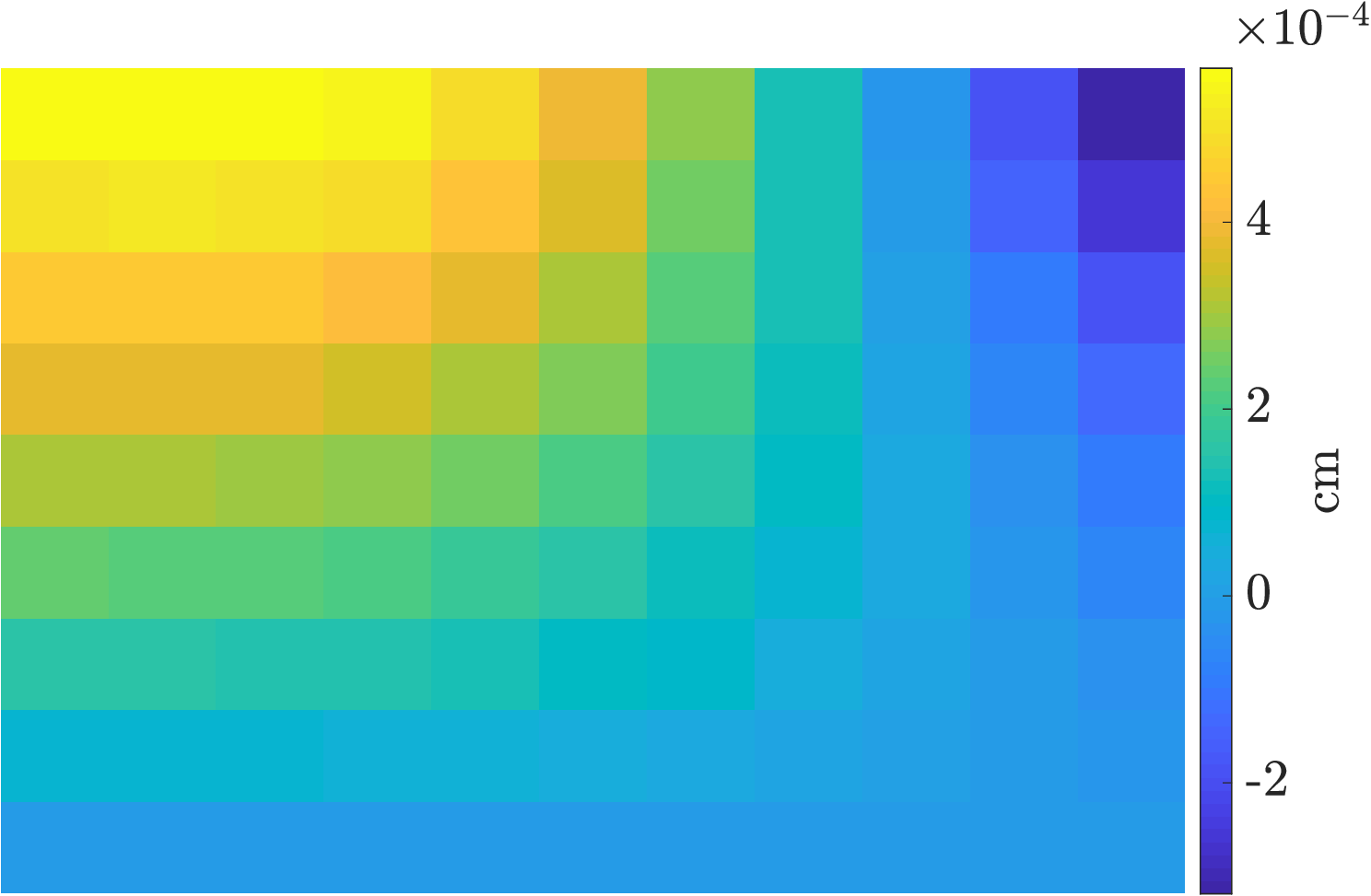}   %[Para su tamaño:anchura con respecto al ancho de la hoja]{Imágenes\Nombre_de_la_imagen}
    \caption{Predicted vertical component ($u_y$).}
 \end{subfigure}
\caption{\textbf{PGNNIV prediction versus FEM solution of the components of the displacement field for a single test-set example of the hardening material.}}
\label{hardu}
\end{figure}

\begin{figure}[htbp]
 \centering
 \begin{subfigure}[t]{0.495\linewidth}
     \centering %Para centrar la imagen
    \includegraphics[scale=0.30]{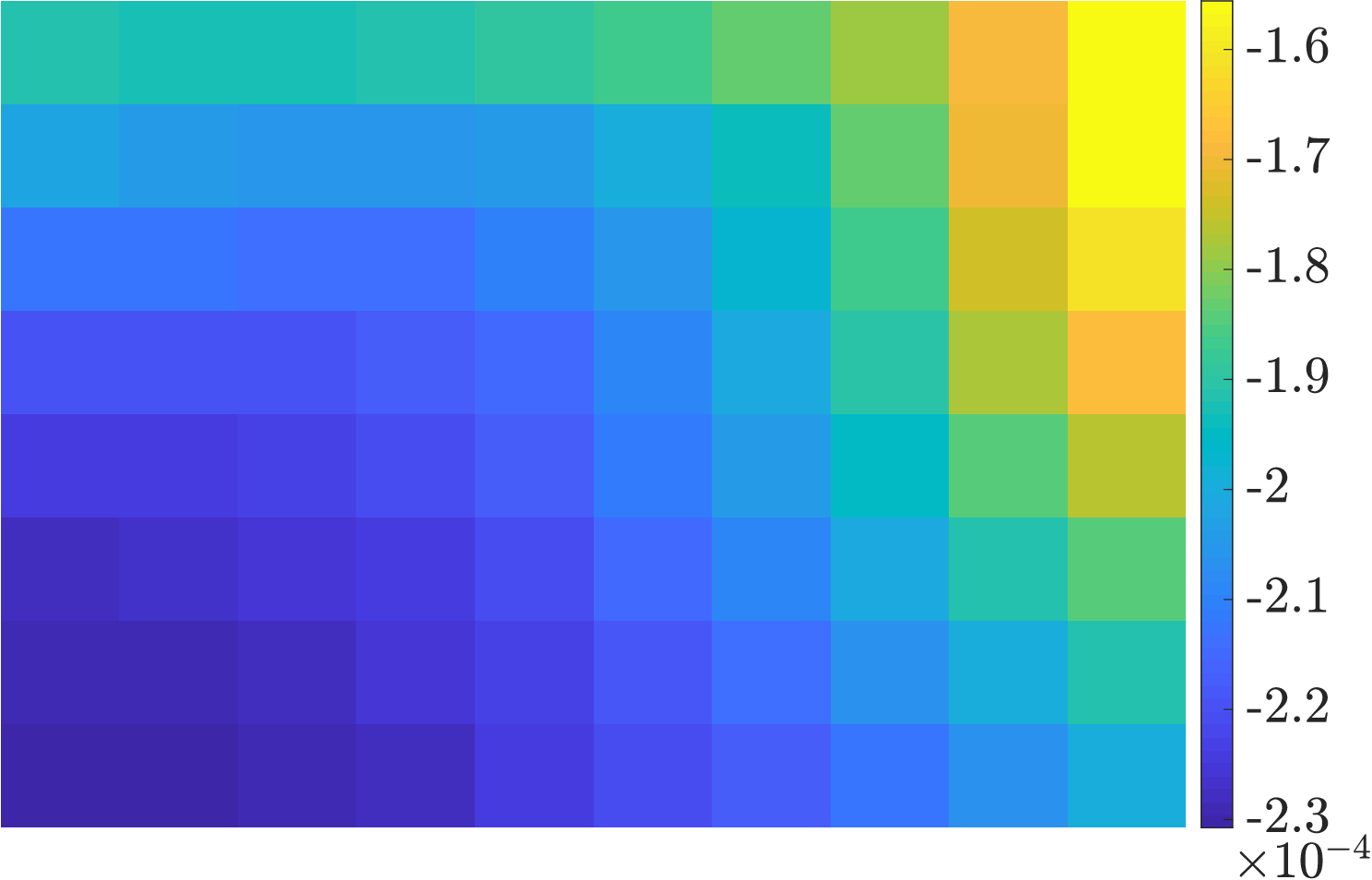}   %[Para su tamaño:anchura con respecto al ancho de la hoja]{Imágenes\Nombre_de_la_imagen}
    \caption{Real normal component ($\varepsilon_{xx}$).}
 \end{subfigure}
 \hfill
 \begin{subfigure}[t]{0.495\linewidth}
     \centering %Para centrar la imagen
    \includegraphics[scale=0.30]{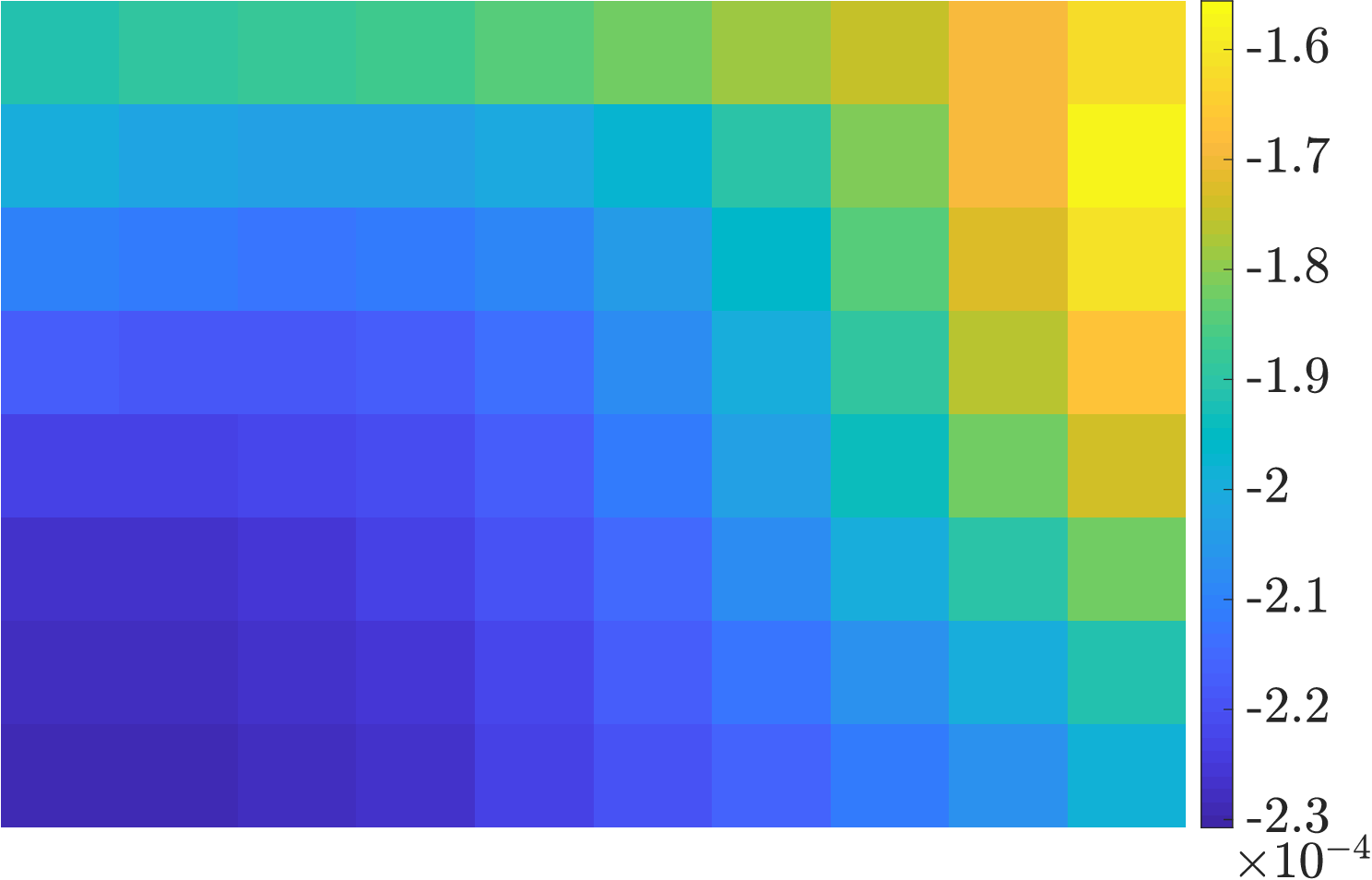}   %[Para su tamaño:anchura con respecto al ancho de la hoja]{Imágenes\Nombre_de_la_imagen}
    \caption{Predicted normal component ($\varepsilon_{xx}$).}
 \end{subfigure}
 \hfill
 \begin{subfigure}[t]{0.495\linewidth}
    \centering %Para centrar la imagen
    \includegraphics[scale=0.30]{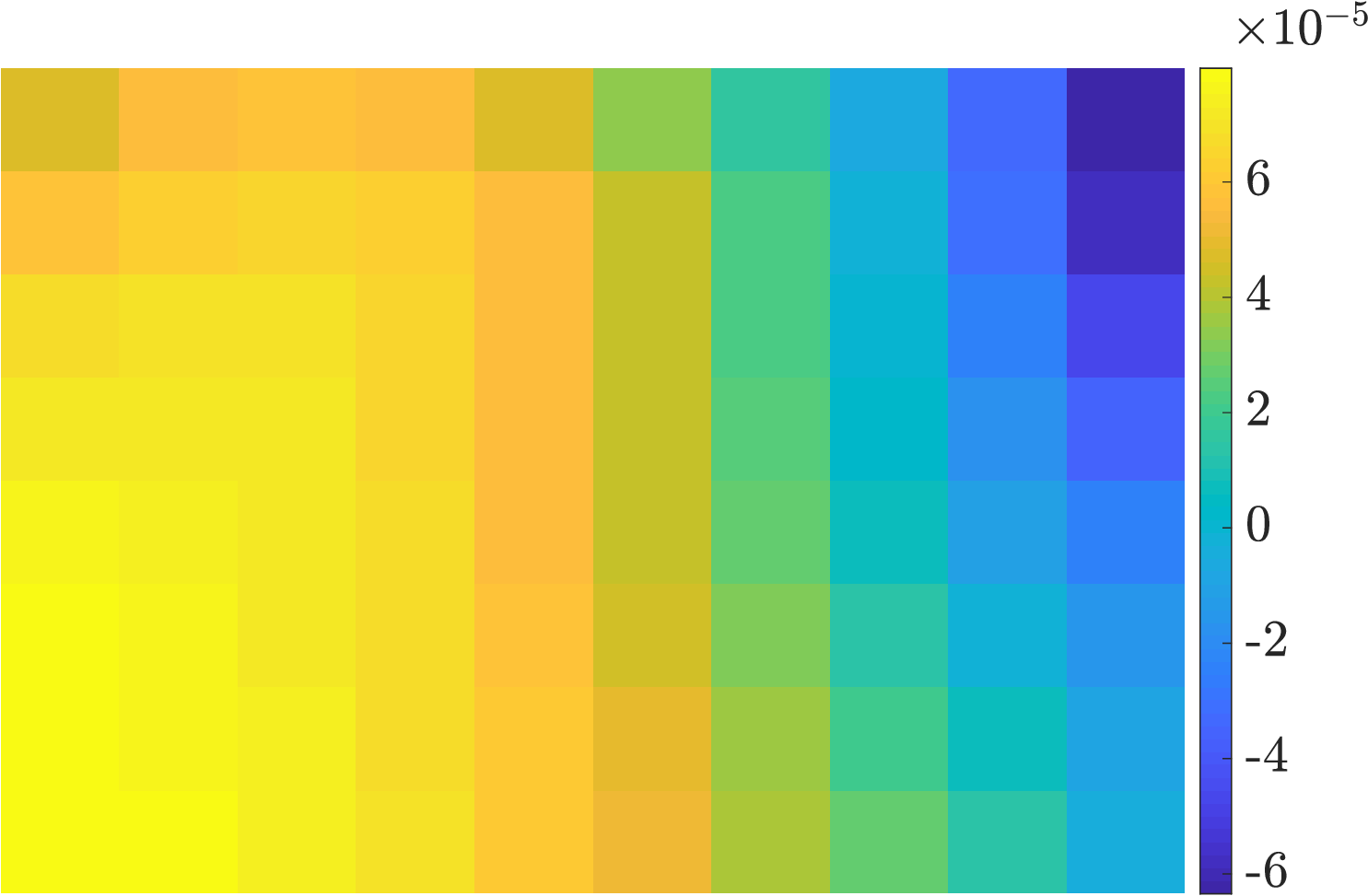}   %[Para su tamaño:anchura con respecto al ancho de la hoja]{Imágenes\Nombre_de_la_imagen}
    \caption{Real normal component ($\varepsilon_{yy}$).}
 \end{subfigure}
  \begin{subfigure}[t]{0.495\linewidth}
     \centering %Para centrar la imagen
    \includegraphics[scale=0.30]{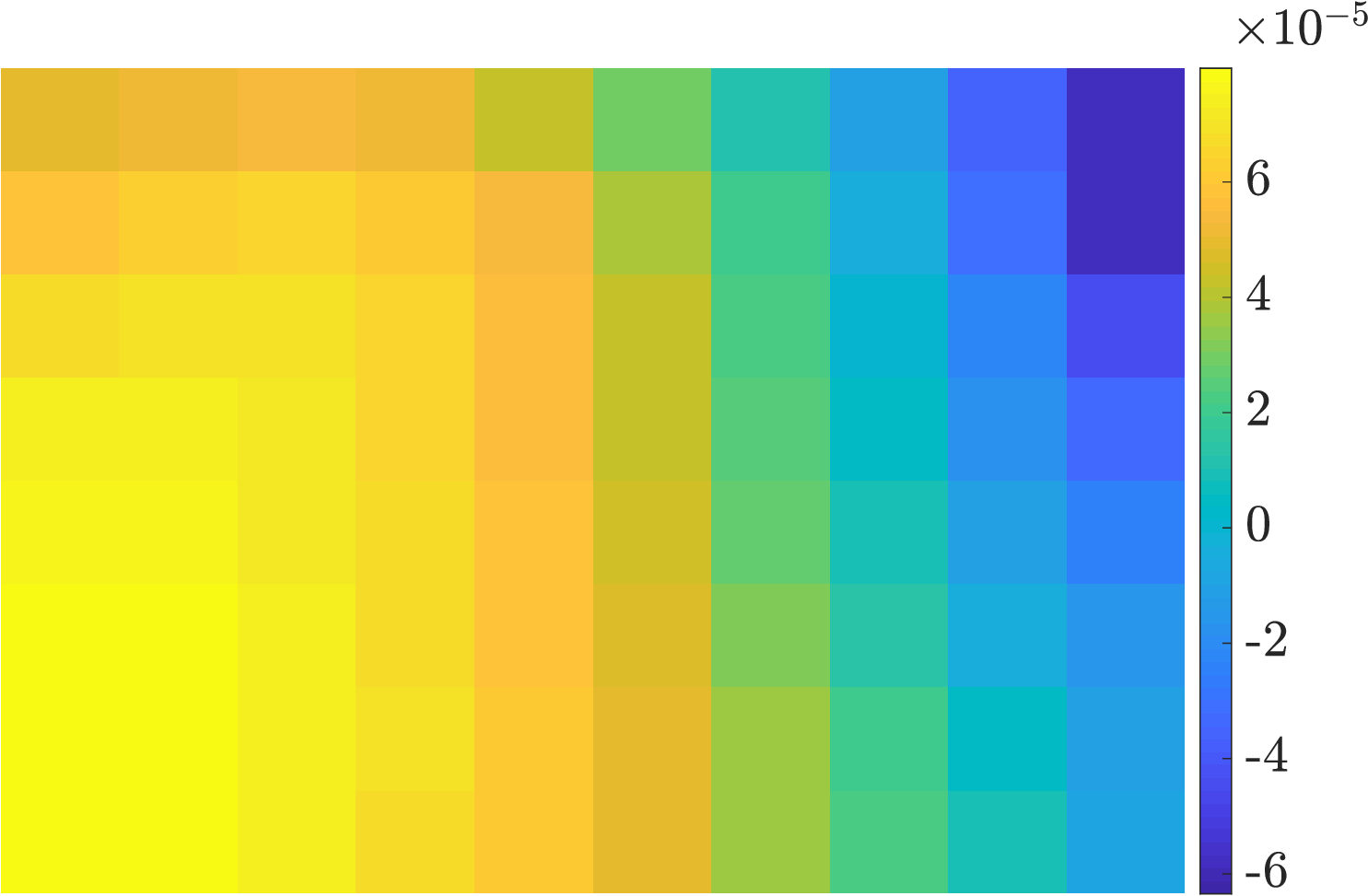}   %[Para su tamaño:anchura con respecto al ancho de la hoja]{Imágenes\Nombre_de_la_imagen}
    \caption{Predicted normal component ($\varepsilon_{yy}$).}
 \end{subfigure}
 \hfill
 \begin{subfigure}[t]{0.495\linewidth}
     \centering %Para centrar la imagen
    \includegraphics[scale=0.30]{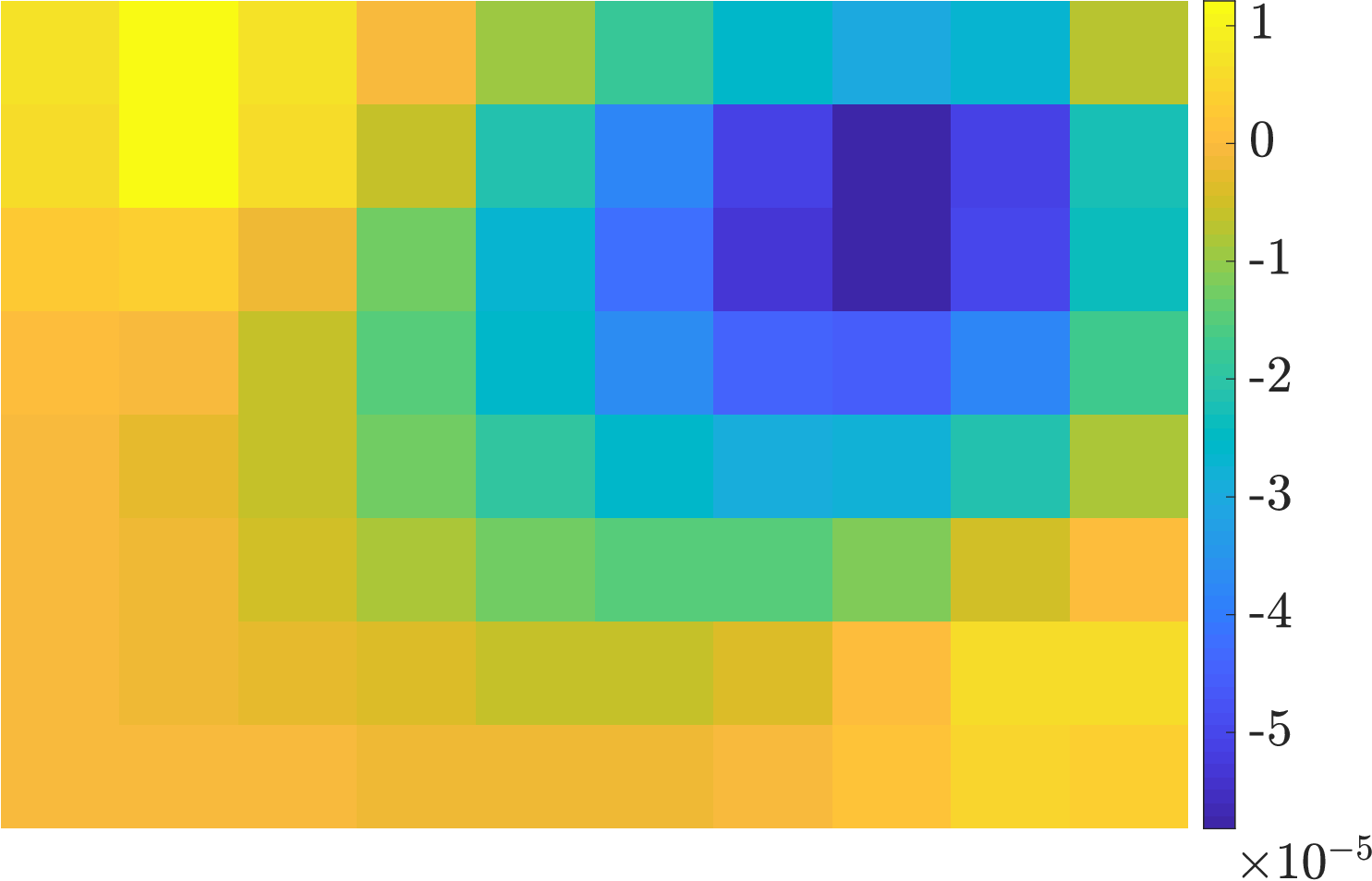}   %[Para su tamaño:anchura con respecto al ancho de la hoja]{Imágenes\Nombre_de_la_imagen}
    \caption{Real shear component ($\varepsilon_{xy}$).}
 \end{subfigure}
 \hfill
 \begin{subfigure}[t]{0.495\linewidth}
    \centering %Para centrar la imagen
    \includegraphics[scale=0.30]{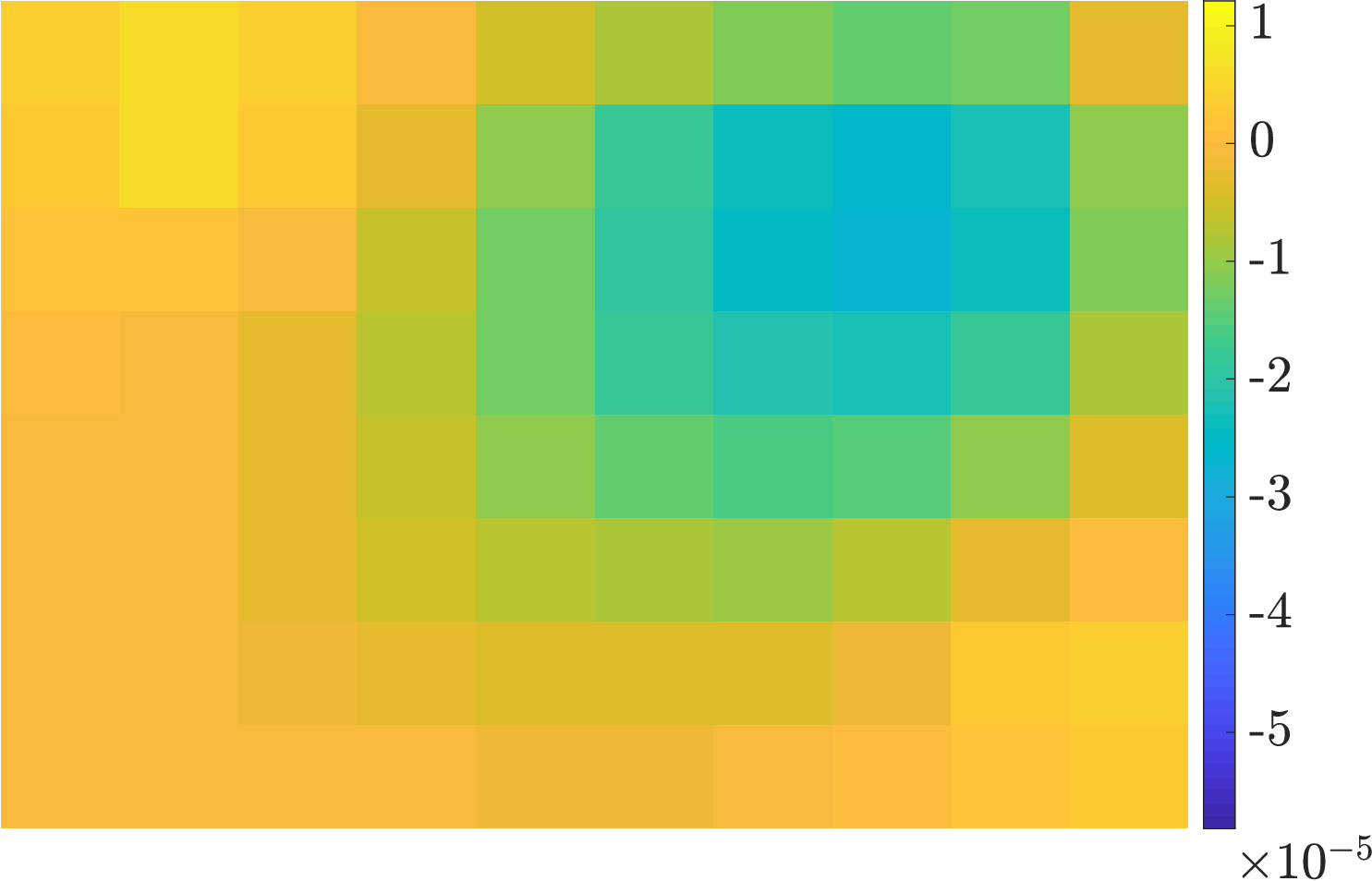}   %[Para su tamaño:anchura con respecto al ancho de la hoja]{Imágenes\Nombre_de_la_imagen}
    \caption{Predicted shear component ($\varepsilon_{xy}$).}
 \end{subfigure}
\caption{\textbf{PGNNIV prediction versus FEM solution of the components of the strain fields for a single test-set example of the hardening material.}}
\label{harde}
\end{figure}

\begin{figure}[htbp]
 \centering
 \begin{subfigure}[t]{0.495\linewidth}
     \centering %Para centrar la imagen
    \includegraphics[scale=0.3]{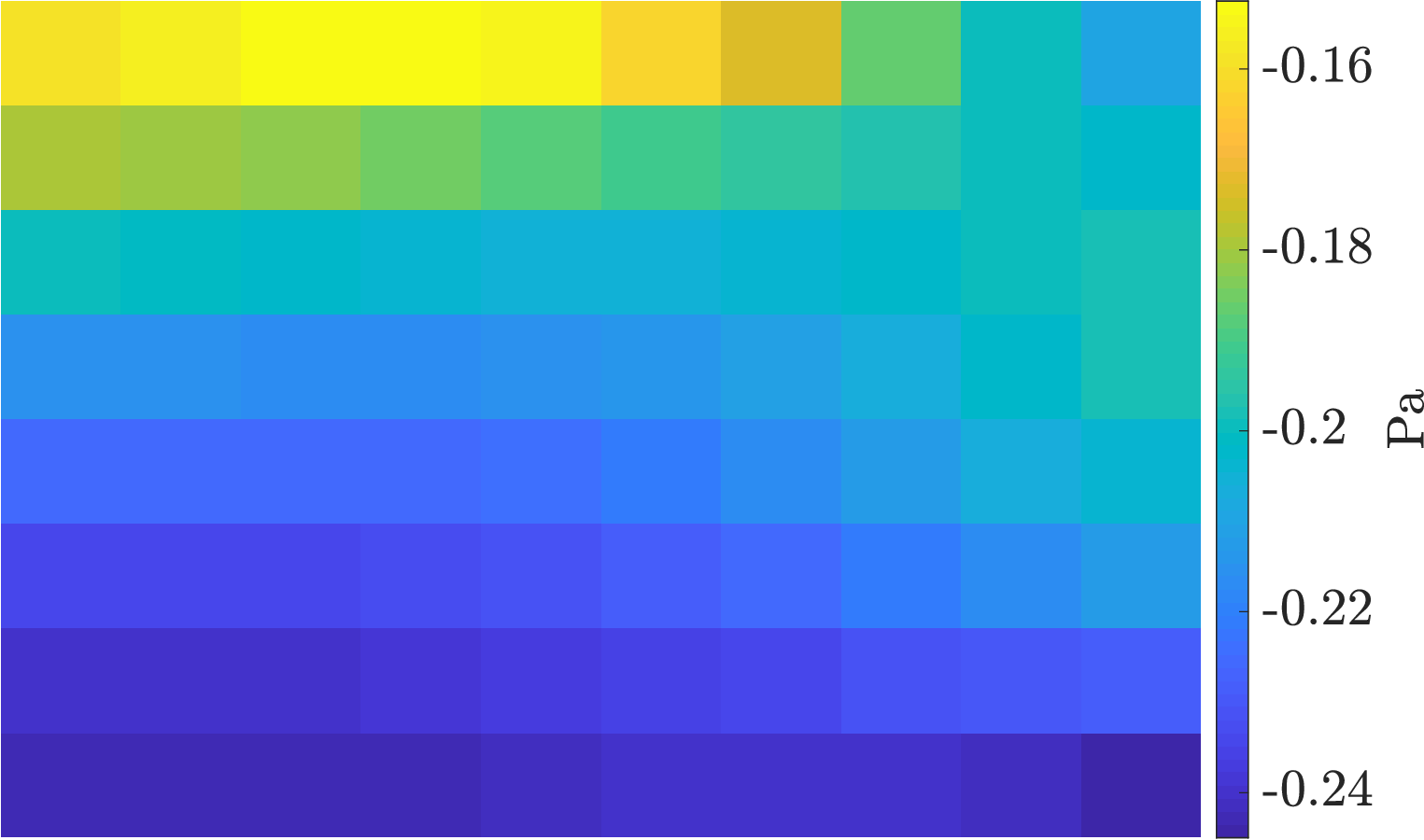}   %[Para su tamaño:anchura con respecto al ancho de la hoja]{Imágenes\Nombre_de_la_imagen}
    \caption{Real normal component ($\sigma_{xx}$).}
 \end{subfigure}
 \hfill
 \begin{subfigure}[t]{0.495\linewidth}
     \centering %Para centrar la imagen
    \includegraphics[scale=0.3]{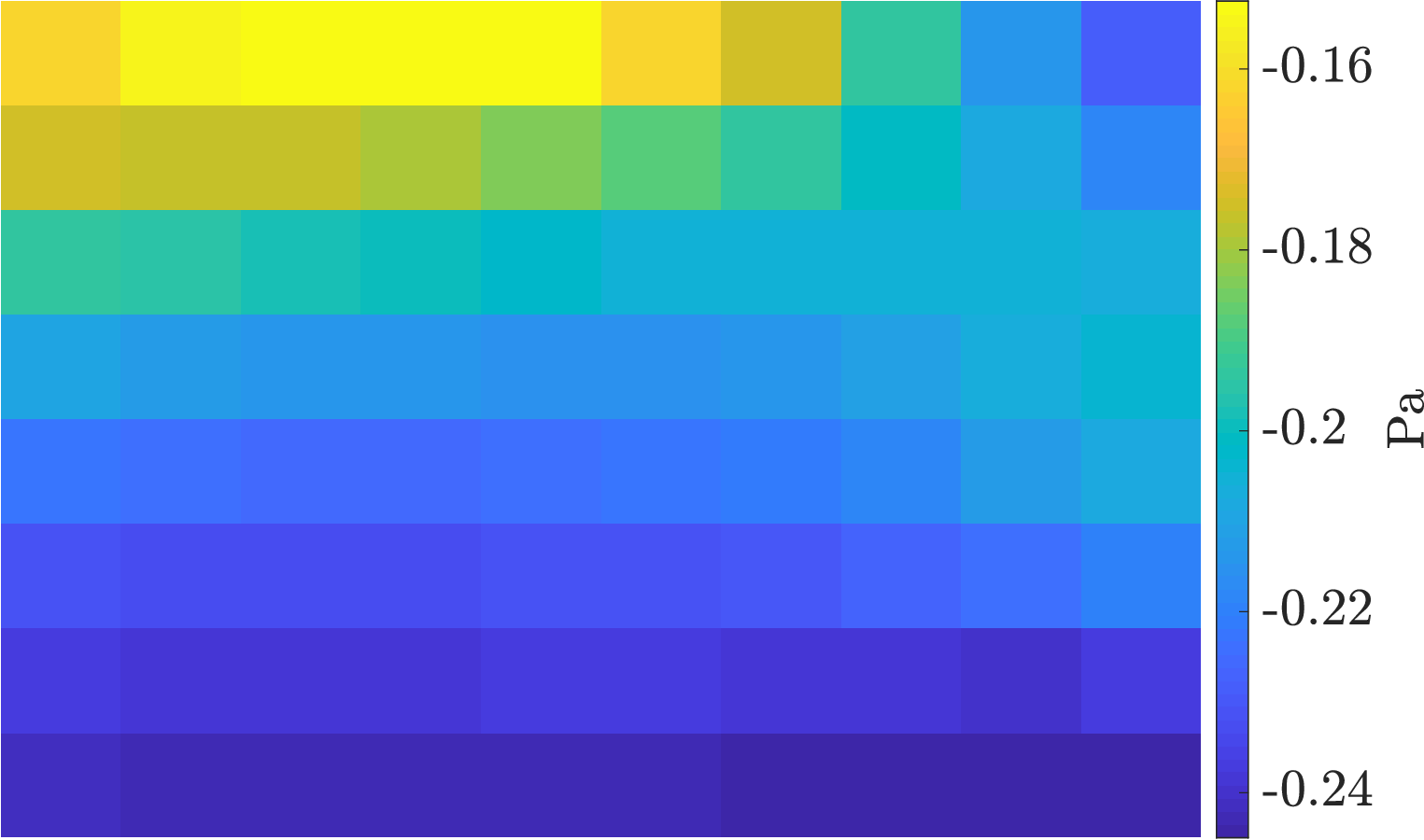}   %[Para su tamaño:anchura con respecto al ancho de la hoja]{Imágenes\Nombre_de_la_imagen}
    \caption{Predicted normal component ($\sigma_{xx}$).}
 \end{subfigure}
 \hfill
 \begin{subfigure}[t]{0.495\linewidth}
    \centering %Para centrar la imagen
    \includegraphics[scale=0.3]{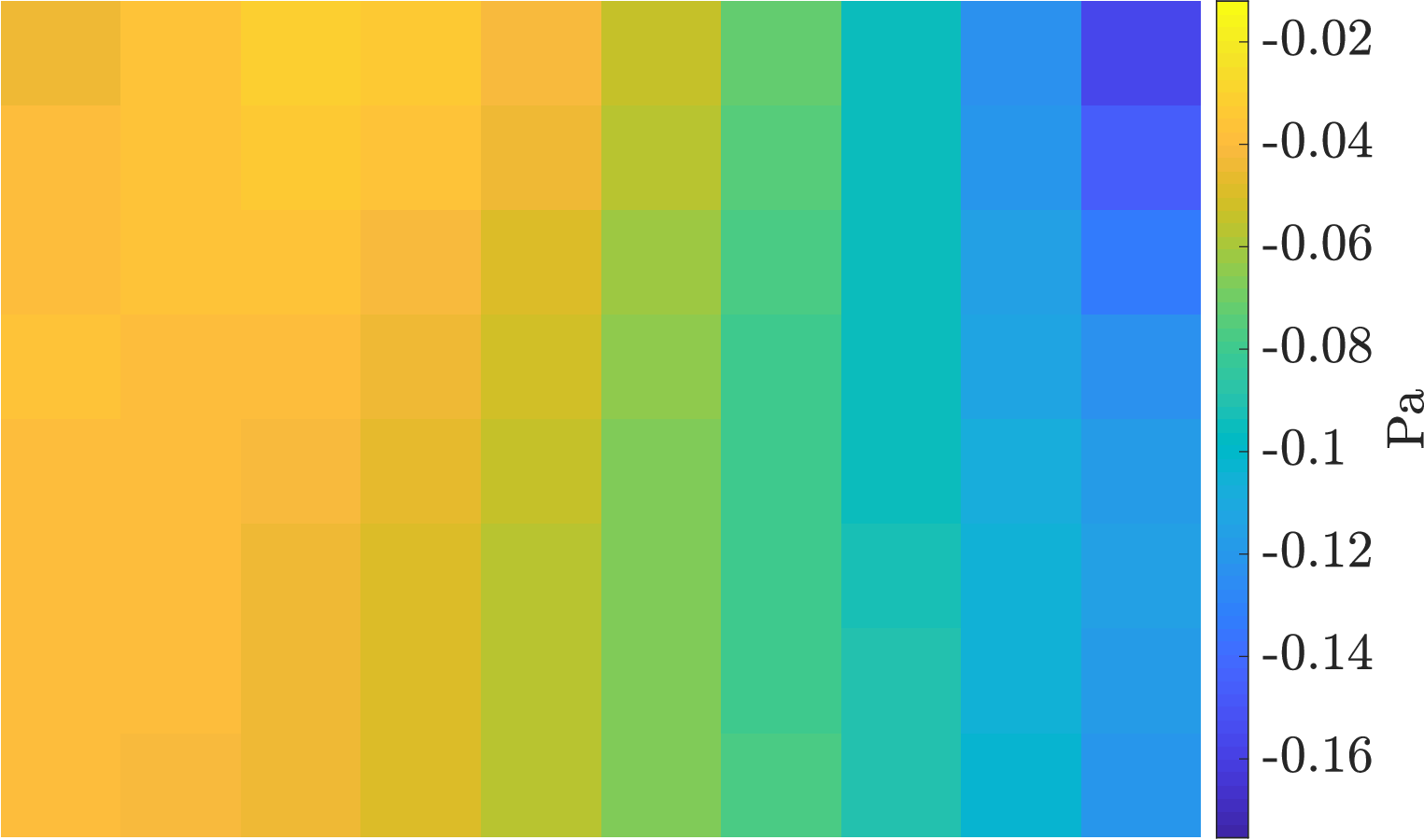}   %[Para su tamaño:anchura con respecto al ancho de la hoja]{Imágenes\Nombre_de_la_imagen}
    \caption{Real normal component ($\sigma_{yy}$).}
 \end{subfigure}
  \begin{subfigure}[t]{0.495\linewidth}
     \centering %Para centrar la imagen
    \includegraphics[scale=0.3]{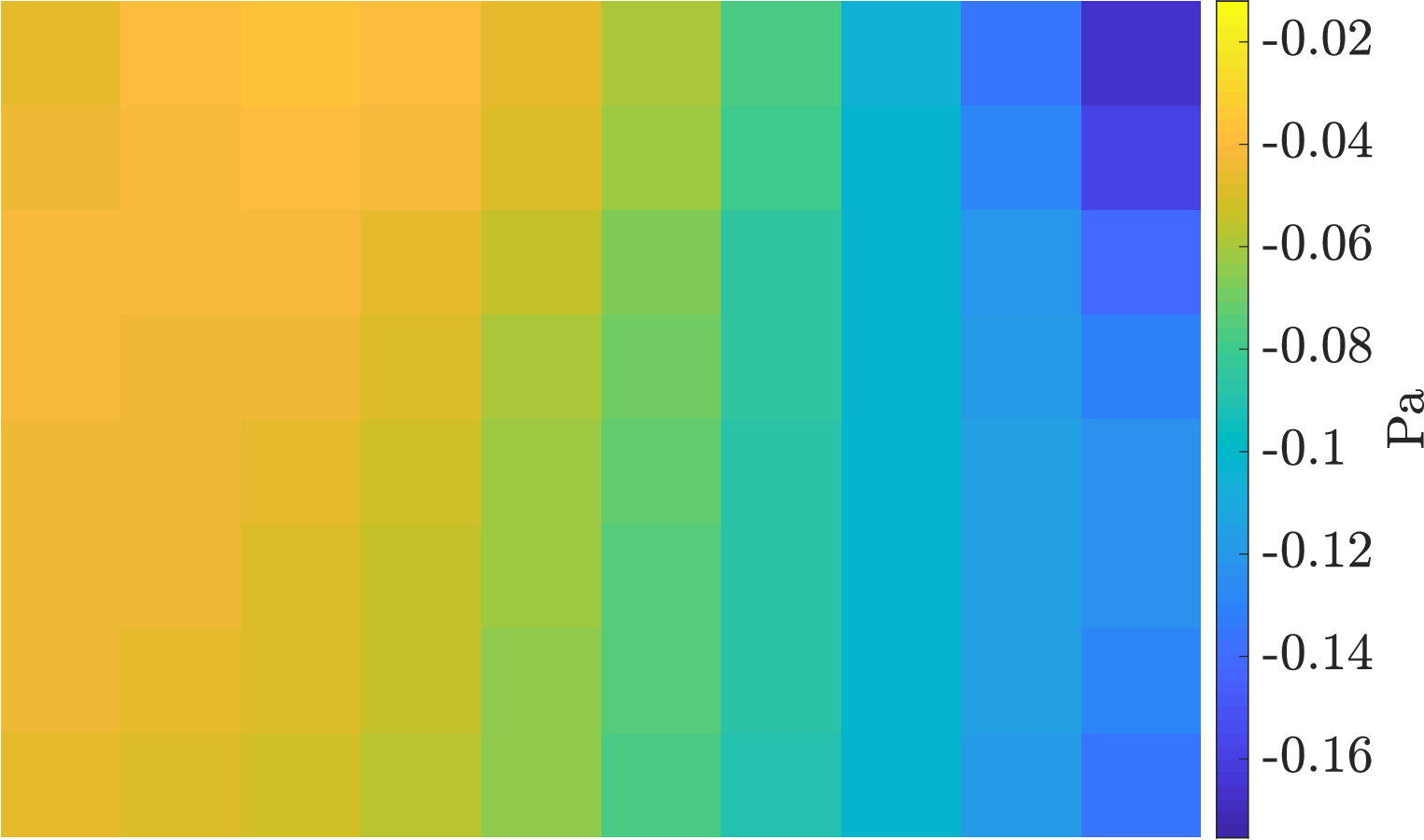}   %[Para su tamaño:anchura con respecto al ancho de la hoja]{Imágenes\Nombre_de_la_imagen}
    \caption{Predicted normal component ($\sigma_{yy}$).}
 \end{subfigure}
 \hfill
 \begin{subfigure}[t]{0.495\linewidth}
     \centering %Para centrar la imagen
    \includegraphics[scale=0.3]{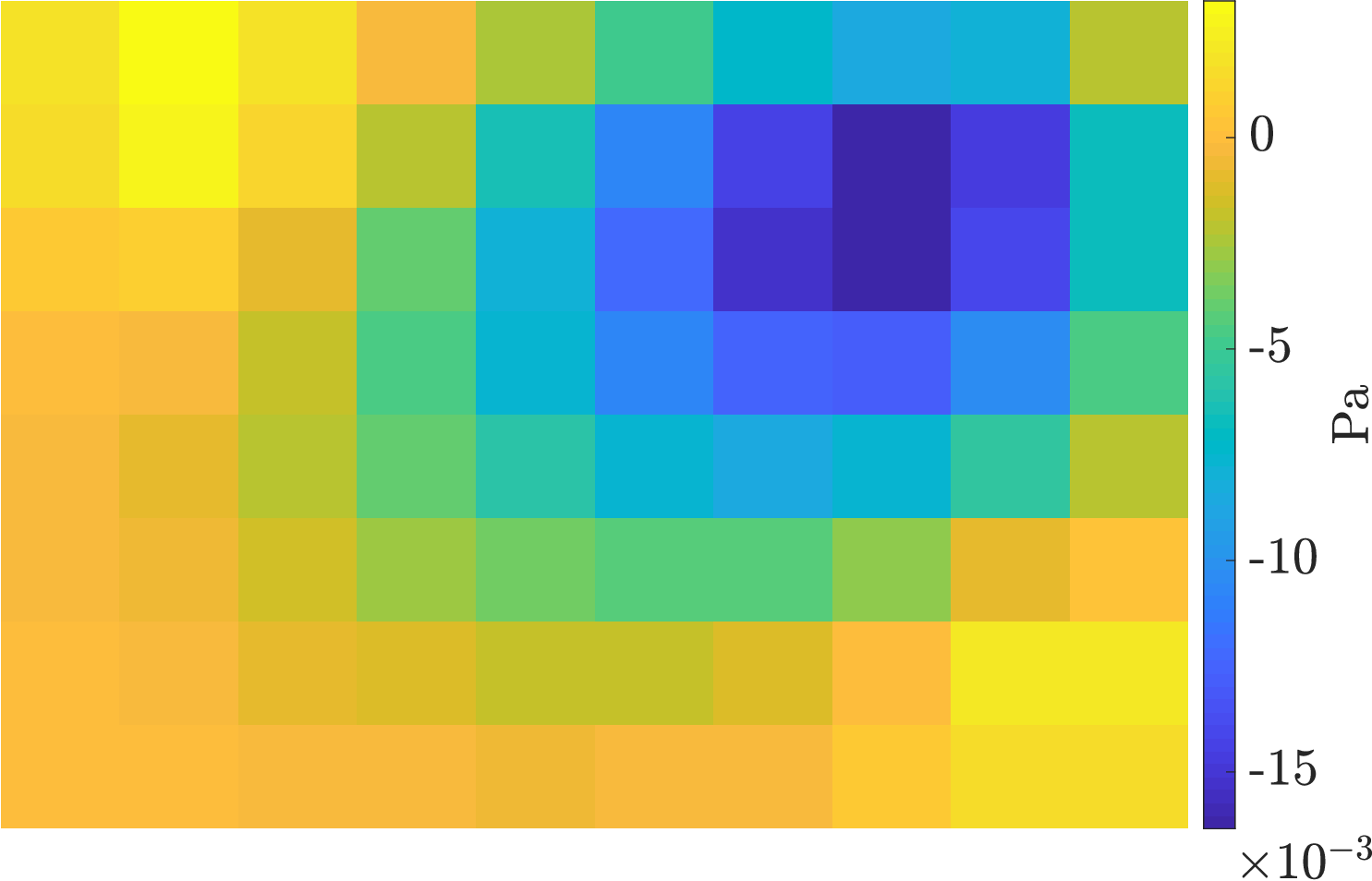}   %[Para su tamaño:anchura con respecto al ancho de la hoja]{Imágenes\Nombre_de_la_imagen}
    \caption{Real shear component ($\sigma_{xy}$).}
 \end{subfigure}
 \hfill
 \begin{subfigure}[t]{0.495\linewidth}
    \centering %Para centrar la imagen
    \includegraphics[scale=0.3]{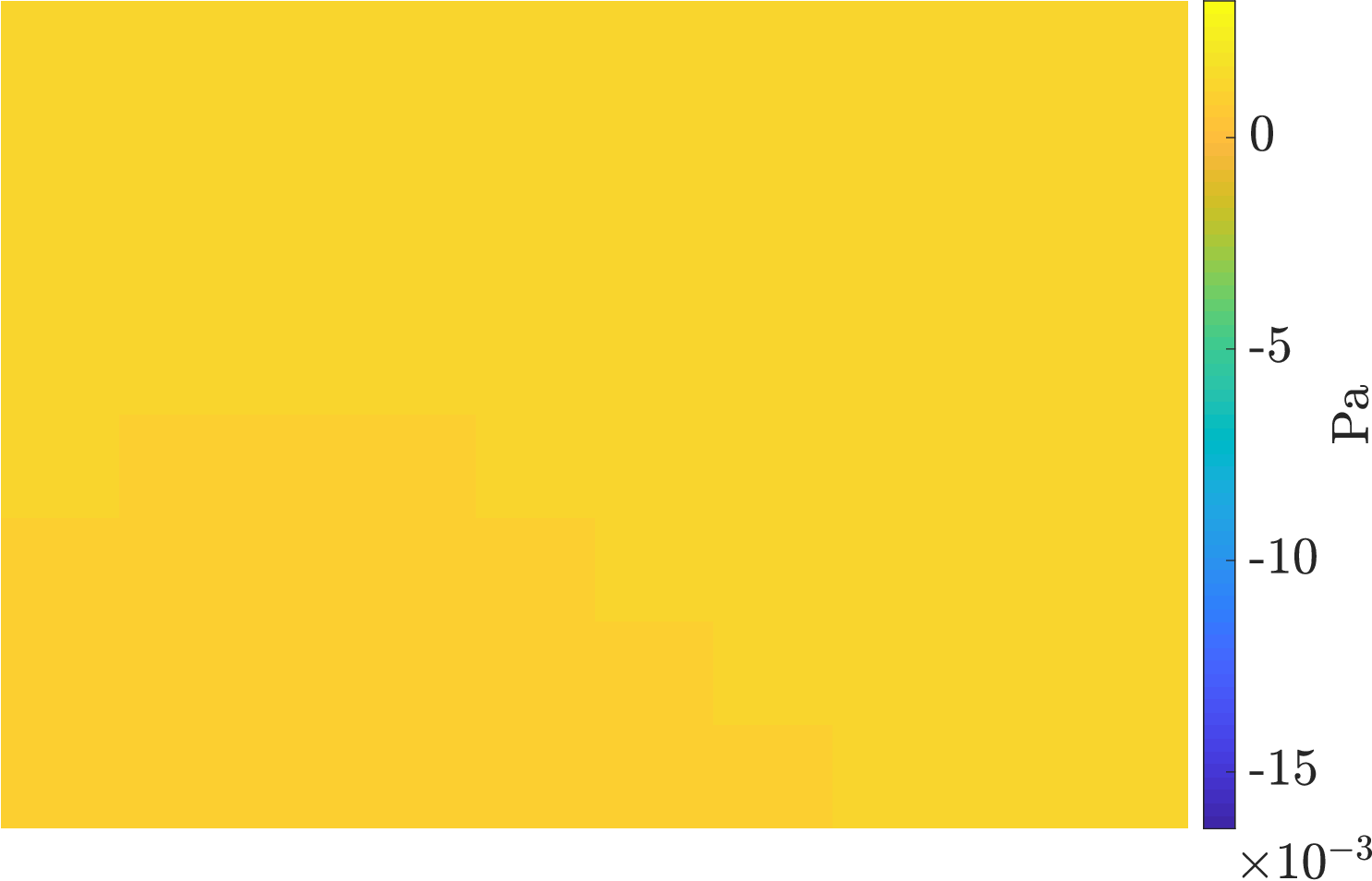}   %[Para su tamaño:anchura con respecto al ancho de la hoja]{Imágenes\Nombre_de_la_imagen}
    \caption{Predicted shear component ($\sigma_{xy}$).}
 \end{subfigure}
\caption{\textbf{PGNNIV prediction versus FEM solution of the components of the stress fields for a single test-set example of the hardening material.}}
\label{hards}
\end{figure}

In Fig. \ref{hardu} the displacement field (FEM solution versus PGNNIV prediction) is represented for a test example. In Fig. \ref{harde} the components of the strain tensor are illustrated and in Fig. \ref{hards} the components of the stress tensor. We observe high similarity between the ground truth and predictive fields despite the coarse discretization.

%\clearpage

\subsection{Finite strains.}

\begin{figure}[htbp]
 \centering
 \begin{subfigure}[t]{0.495\linewidth}
     \centering %Para centrar la imagen
    \includegraphics[scale=0.3]{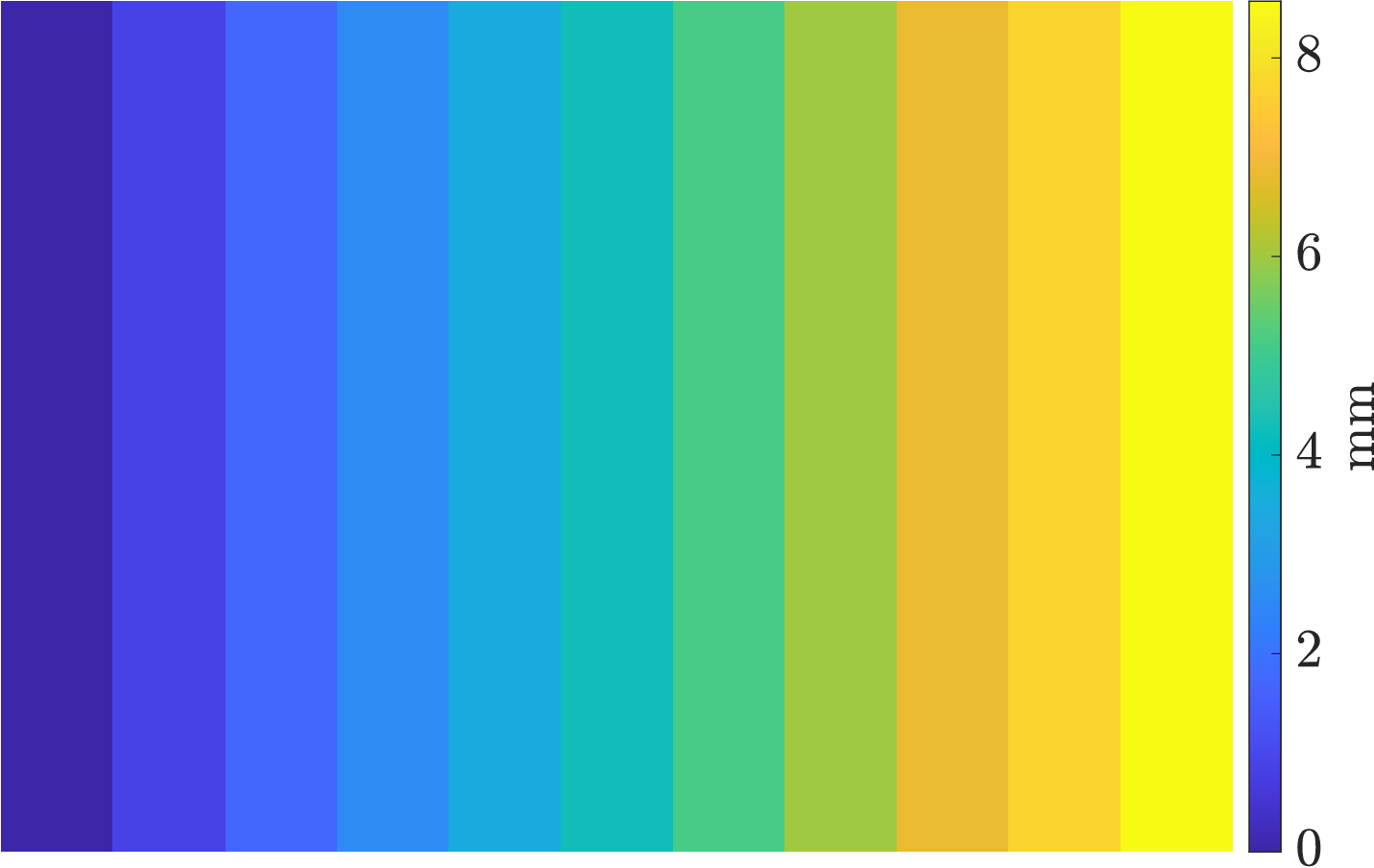}   %[Para su tamaño:anchura con respecto al ancho de la hoja]{Imágenes\Nombre_de_la_imagen}
    \caption{Real horizontal component ($u_x$).}
 \end{subfigure}
 \hfill
 \begin{subfigure}[t]{0.495\linewidth}
    \centering %Para centrar la imagen
    \includegraphics[scale=0.3]{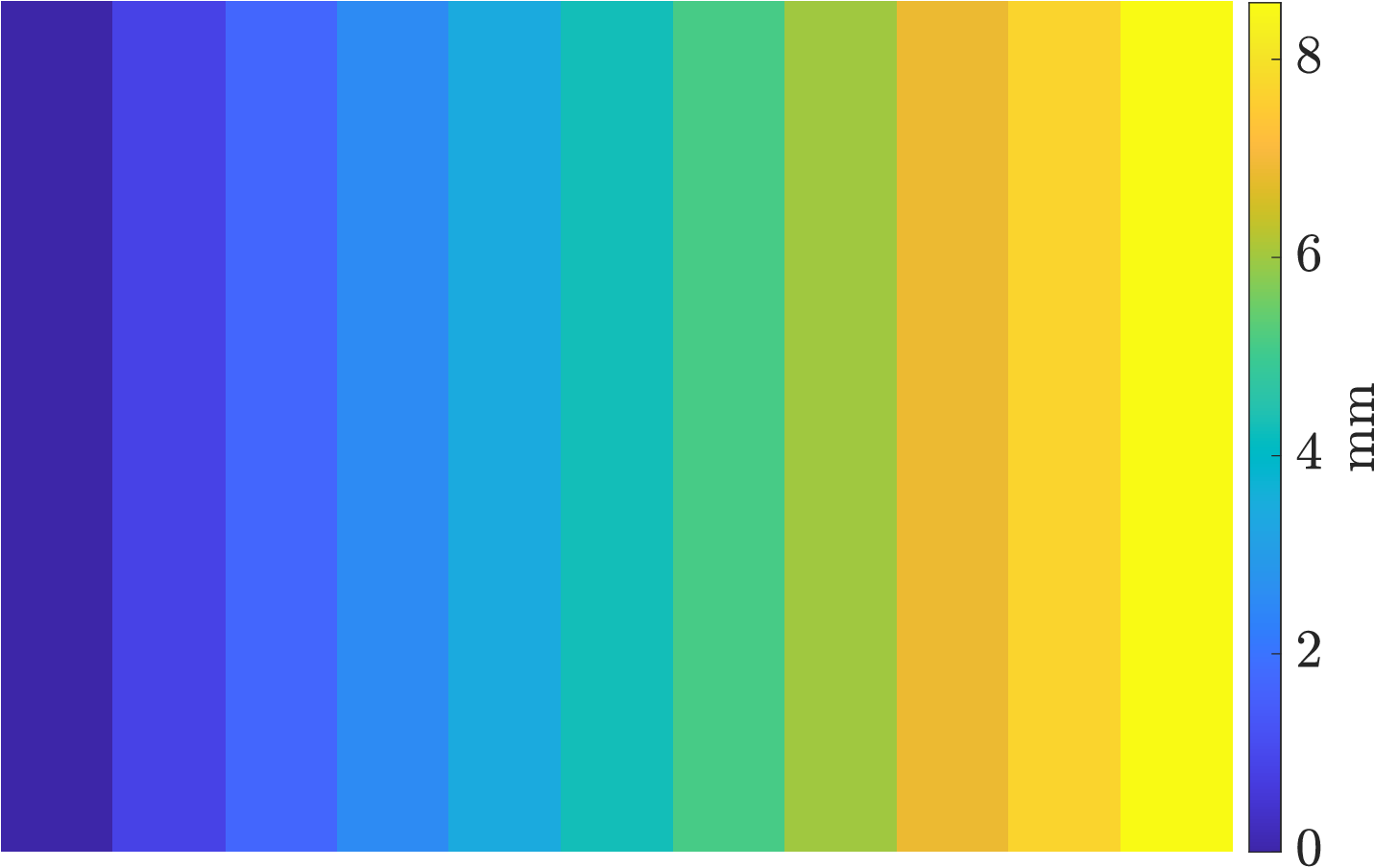}   %[Para su tamaño:anchura con respecto al ancho de la hoja]{Imágenes\Nombre_de_la_imagen}
    \caption{Predicted horizontal component ($u_x$).}
 \end{subfigure}
 \begin{subfigure}[t]{0.495\linewidth}
     \centering %Para centrar la imagen
    \includegraphics[scale=0.3]{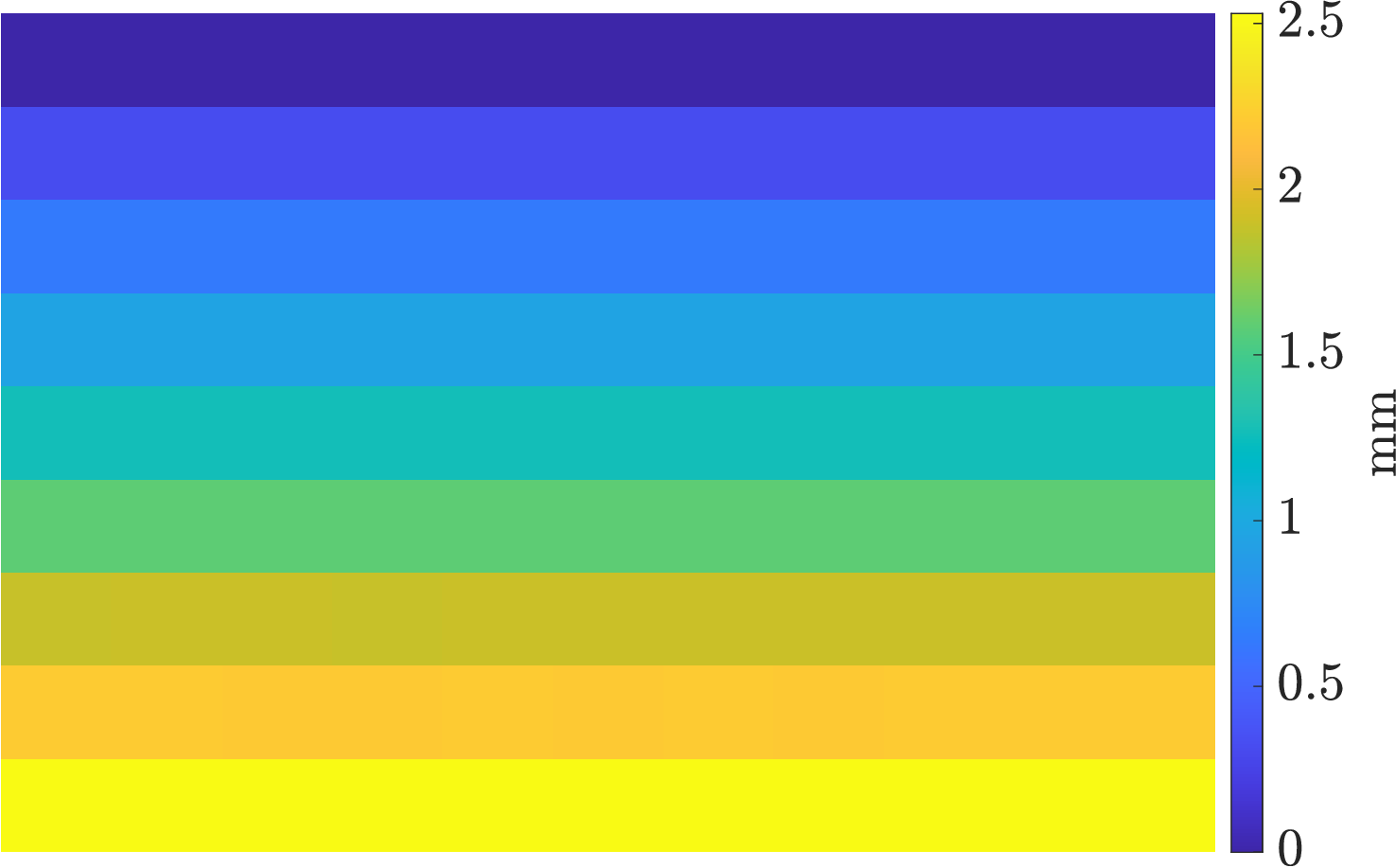}   %[Para su tamaño:anchura con respecto al ancho de la hoja]{Imágenes\Nombre_de_la_imagen}
    \caption{Real vertical component ($u_y$).}
 \end{subfigure}
 \hfill
 \begin{subfigure}[t]{0.495\linewidth}
    \centering %Para centrar la imagen
    \includegraphics[scale=0.3]{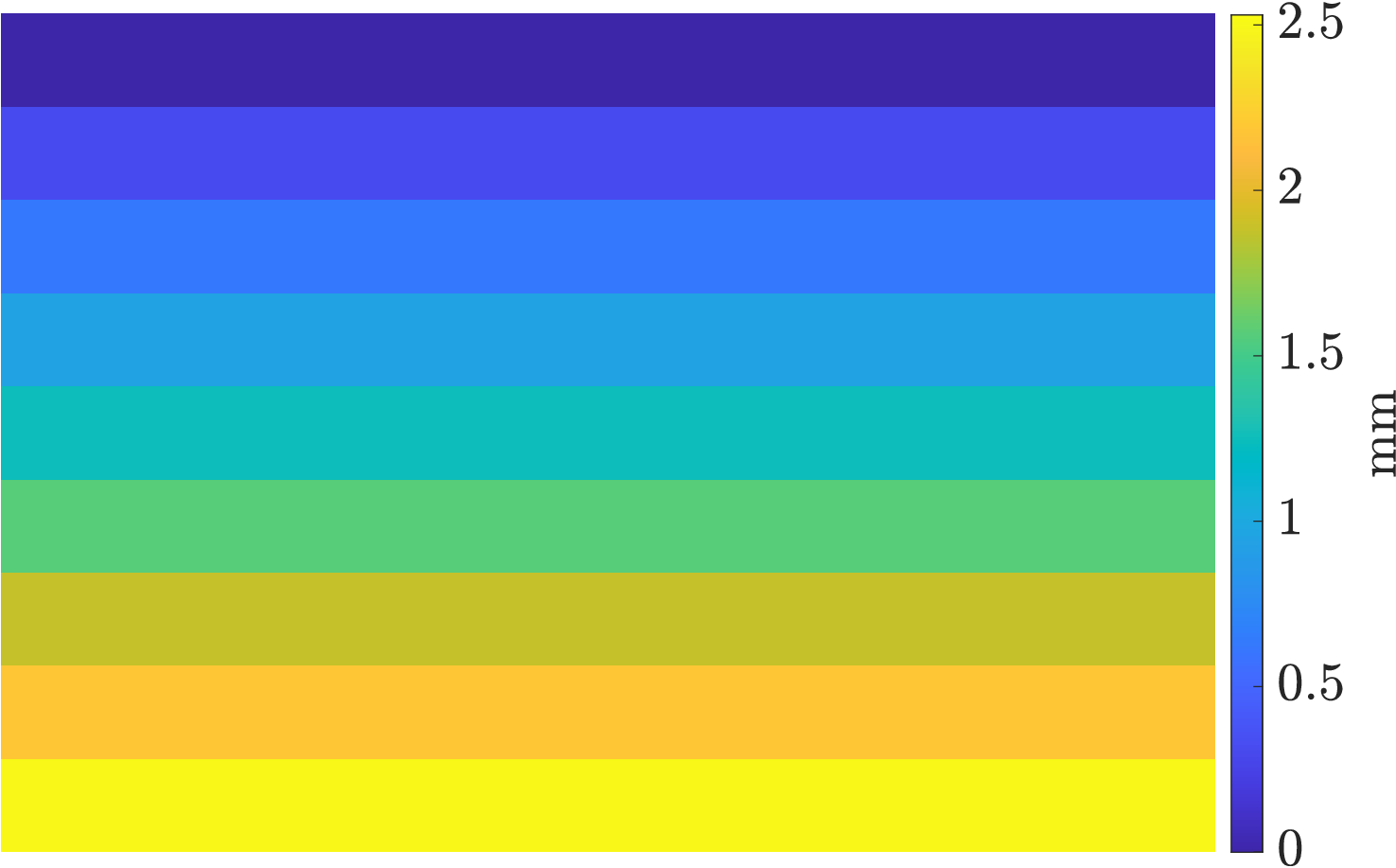}   %[Para su tamaño:anchura con respecto al ancho de la hoja]{Imágenes\Nombre_de_la_imagen}
    \caption{Predicted vertical component ($u_y$).}
 \end{subfigure}
\caption{\textbf{Nonlinear PGNNIV prediction versus FEM solution of the components of the displacement field for a single test-set example of the Ogden material.}}
\label{ogdenu}
\end{figure}

\begin{figure}[htbp]
 \centering
 \begin{subfigure}[t]{0.495\linewidth}
     \centering %Para centrar la imagen
    \includegraphics[scale=0.30]{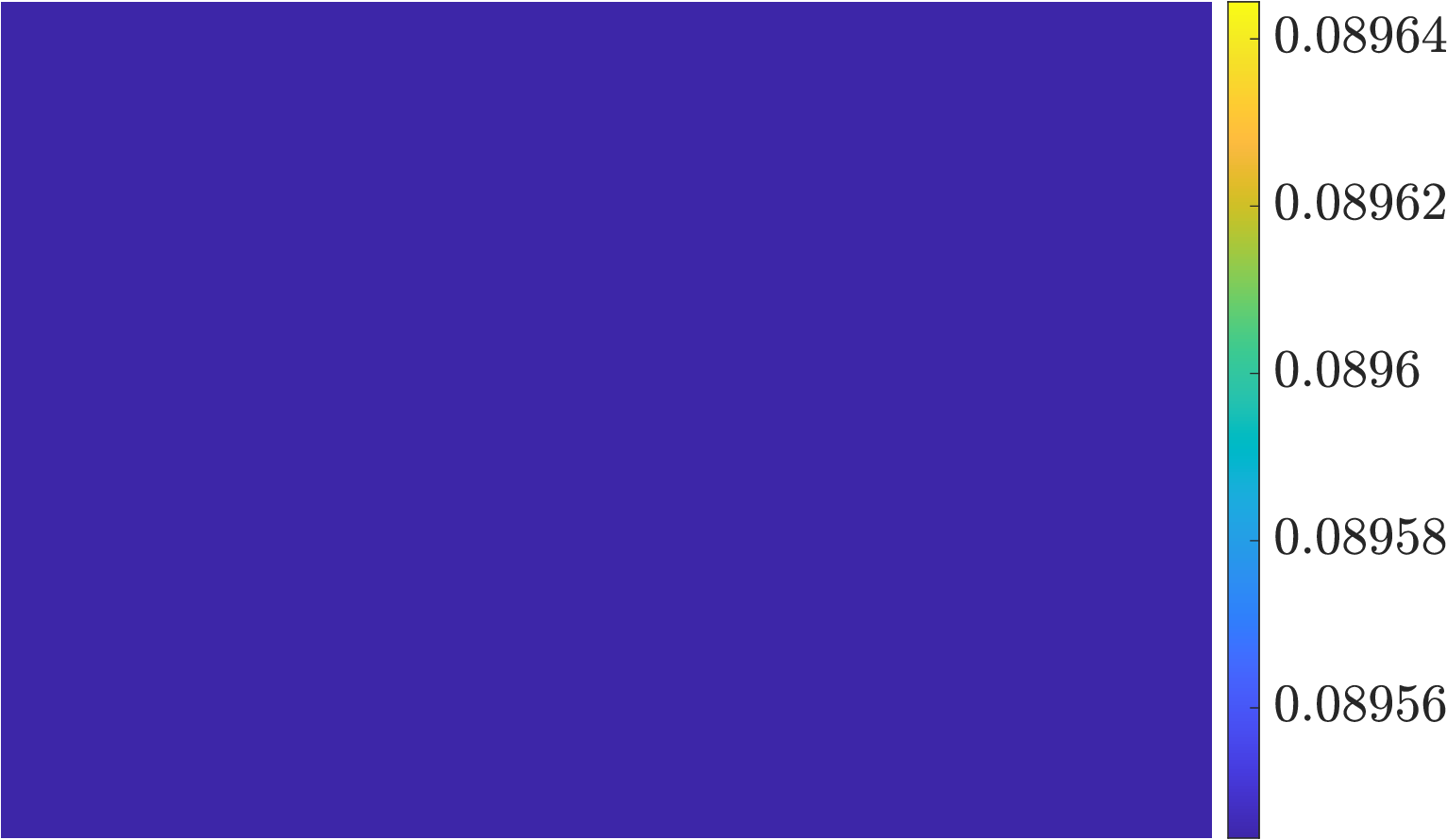}   %[Para su tamaño:anchura con respecto al ancho de la hoja]{Imágenes\Nombre_de_la_imagen}
    \caption{Real normal component ($E_{xx}$).}
 \end{subfigure}
 \hfill
 \begin{subfigure}[t]{0.495\linewidth}
     \centering %Para centrar la imagen
    \includegraphics[scale=0.30]{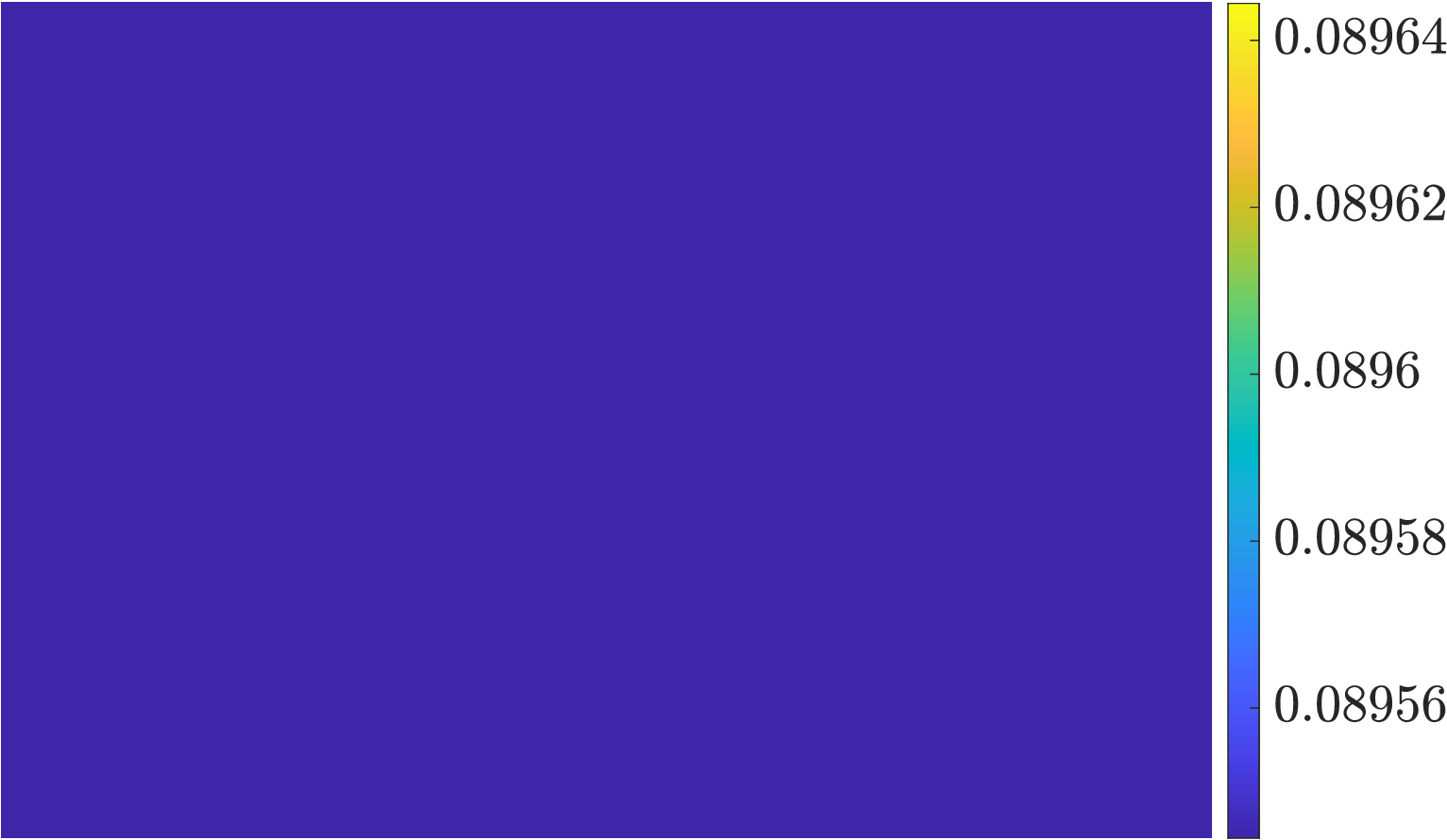}   %[Para su tamaño:anchura con respecto al ancho de la hoja]{Imágenes\Nombre_de_la_imagen}
    \caption{Predicted normal component ($E_{xx}$).}
    \label{fig:app_l_st2}   %Nombre para referirse a la imagen
 \end{subfigure}
 \hfill
 \begin{subfigure}[t]{0.495\linewidth}
    \centering %Para centrar la imagen
    \includegraphics[scale=0.30]{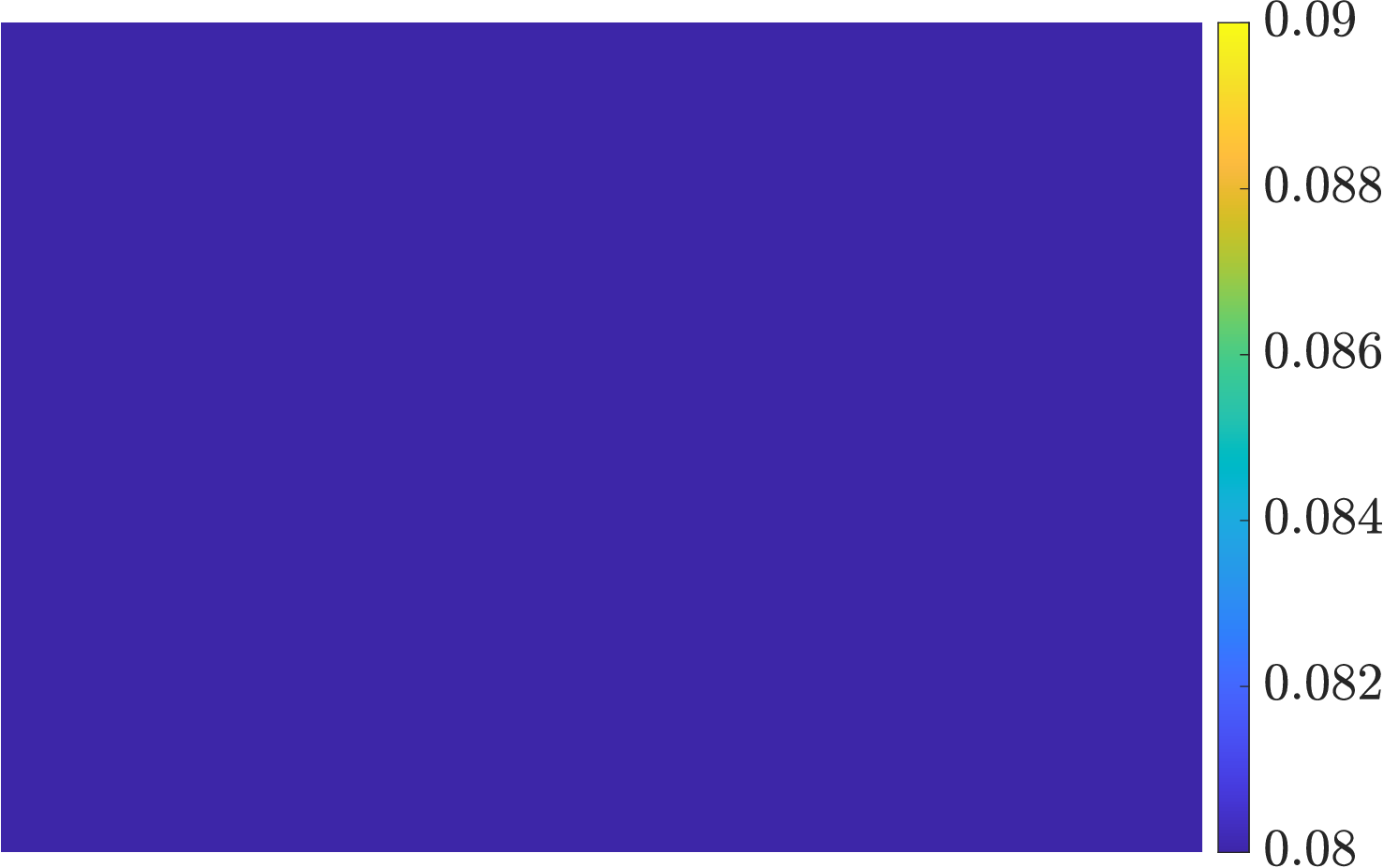}   %[Para su tamaño:anchura con respecto al ancho de la hoja]{Imágenes\Nombre_de_la_imagen}
    \caption{Real normal component ($E_{yy}$).}
 \end{subfigure}
  \begin{subfigure}[t]{0.495\linewidth}
     \centering %Para centrar la imagen
    \includegraphics[scale=0.30]{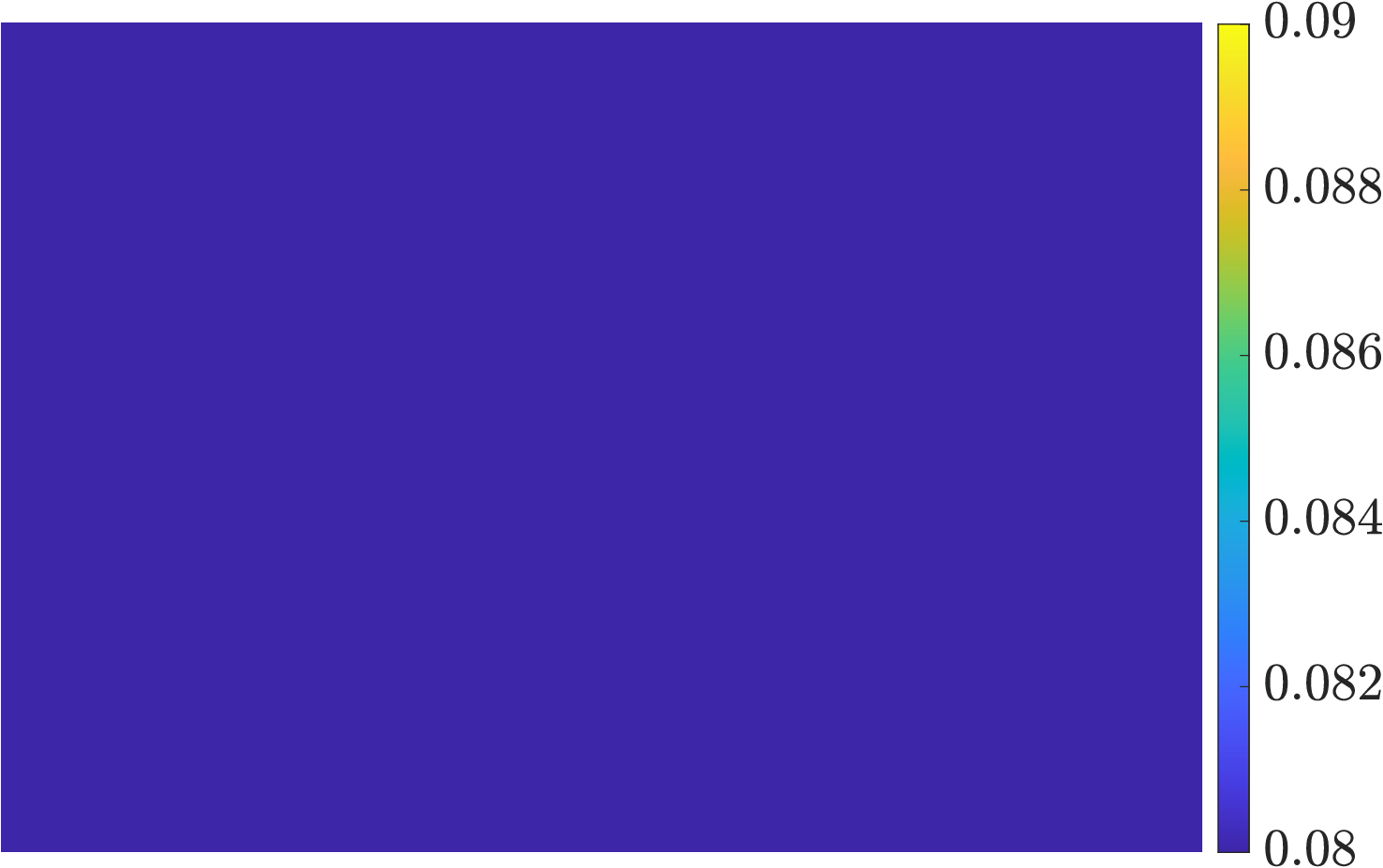}   %[Para su tamaño:anchura con respecto al ancho de la hoja]{Imágenes\Nombre_de_la_imagen}
    \caption{Predicted normal component ($E_{yy}$).}
 \end{subfigure}
 \hfill
 \begin{subfigure}[t]{0.495\linewidth}
     \centering %Para centrar la imagen
    \includegraphics[scale=0.30]{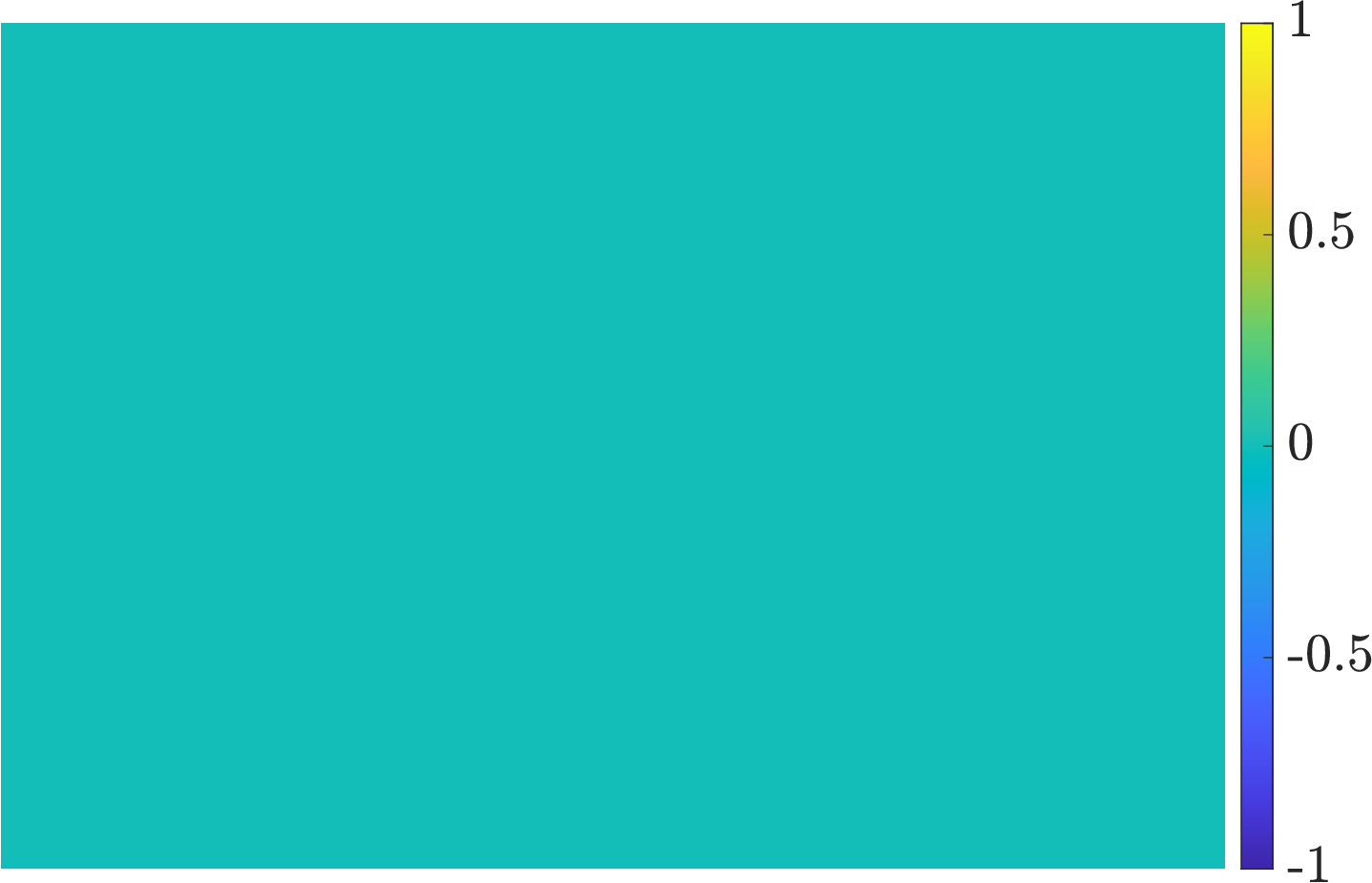}   %[Para su tamaño:anchura con respecto al ancho de la hoja]{Imágenes\Nombre_de_la_imagen}
    \caption{Real shear component ($E_{xy}$).}
 \end{subfigure}
 \hfill
 \begin{subfigure}[t]{0.495\linewidth}
    \centering %Para centrar la imagen
    \includegraphics[scale=0.30]{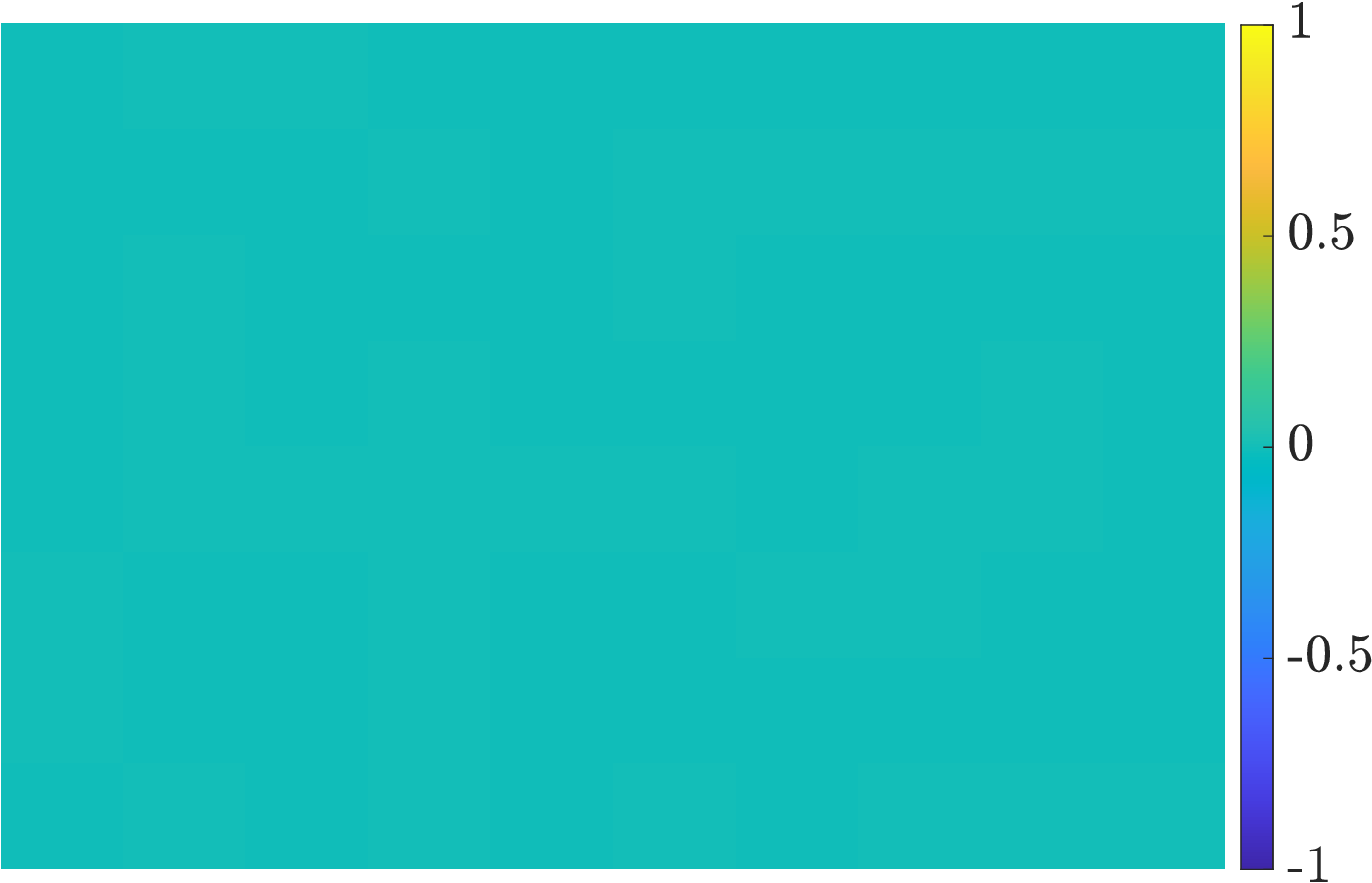}   %[Para su tamaño:anchura con respecto al ancho de la hoja]{Imágenes\Nombre_de_la_imagen}
    \caption{Predicted shear component ($E_{xy}$).}
 \end{subfigure}
\caption{\textbf{PGNNIV prediction versus FEM solution of the components of the strain fields for a single test-set example of the Ogden material.}}
\label{ogdene}
\end{figure}

\begin{figure}[htbp]
 \centering
 \begin{subfigure}[t]{0.495\linewidth}
     \centering %Para centrar la imagen
    \includegraphics[scale=0.3]{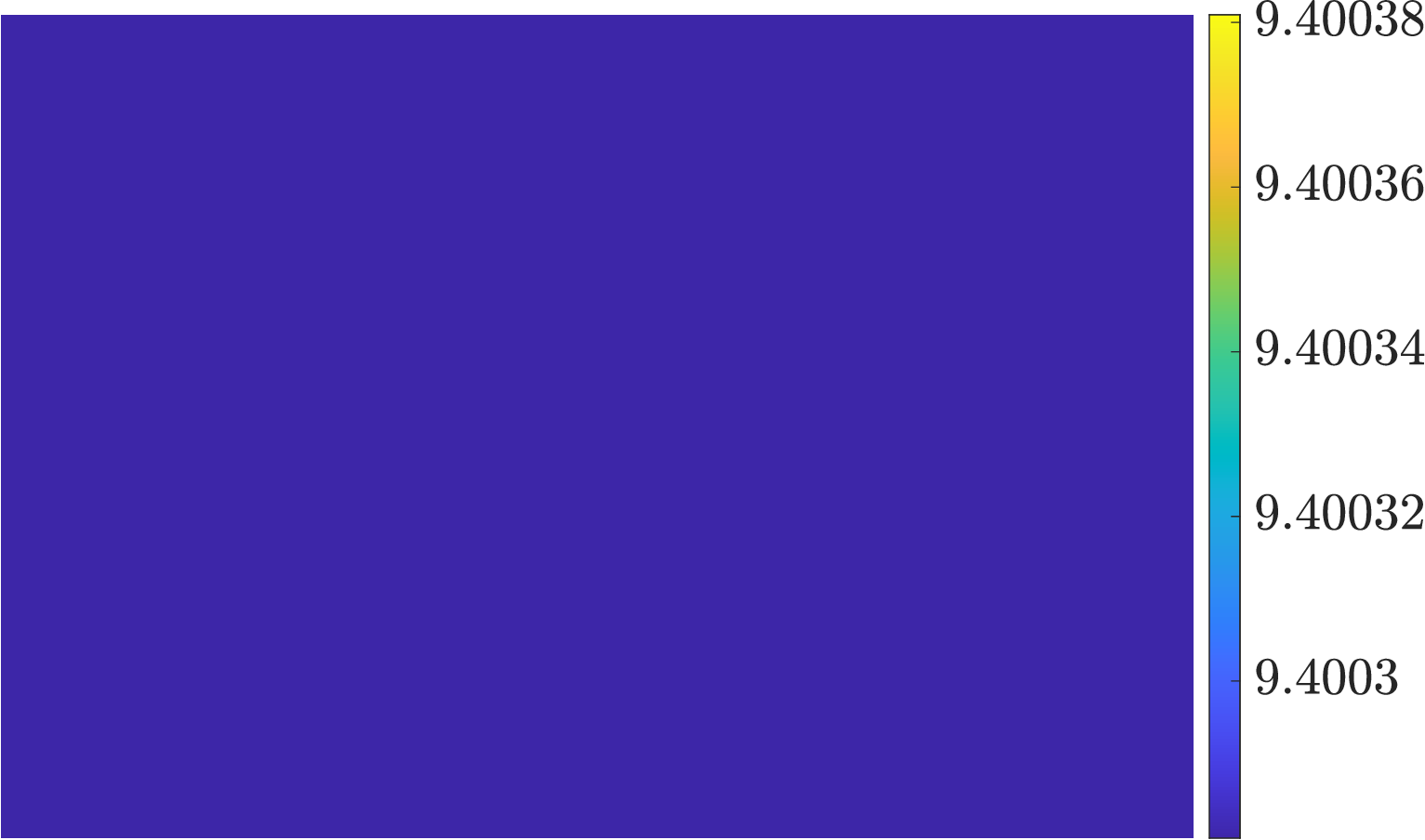}   %[Para su tamaño:anchura con respecto al ancho de la hoja]{Imágenes\Nombre_de_la_imagen}
    \caption{Real normal component ($P_{xx}$).}
 \end{subfigure}
 \hfill
 \begin{subfigure}[t]{0.495\linewidth}
     \centering %Para centrar la imagen
    \includegraphics[scale=0.3]{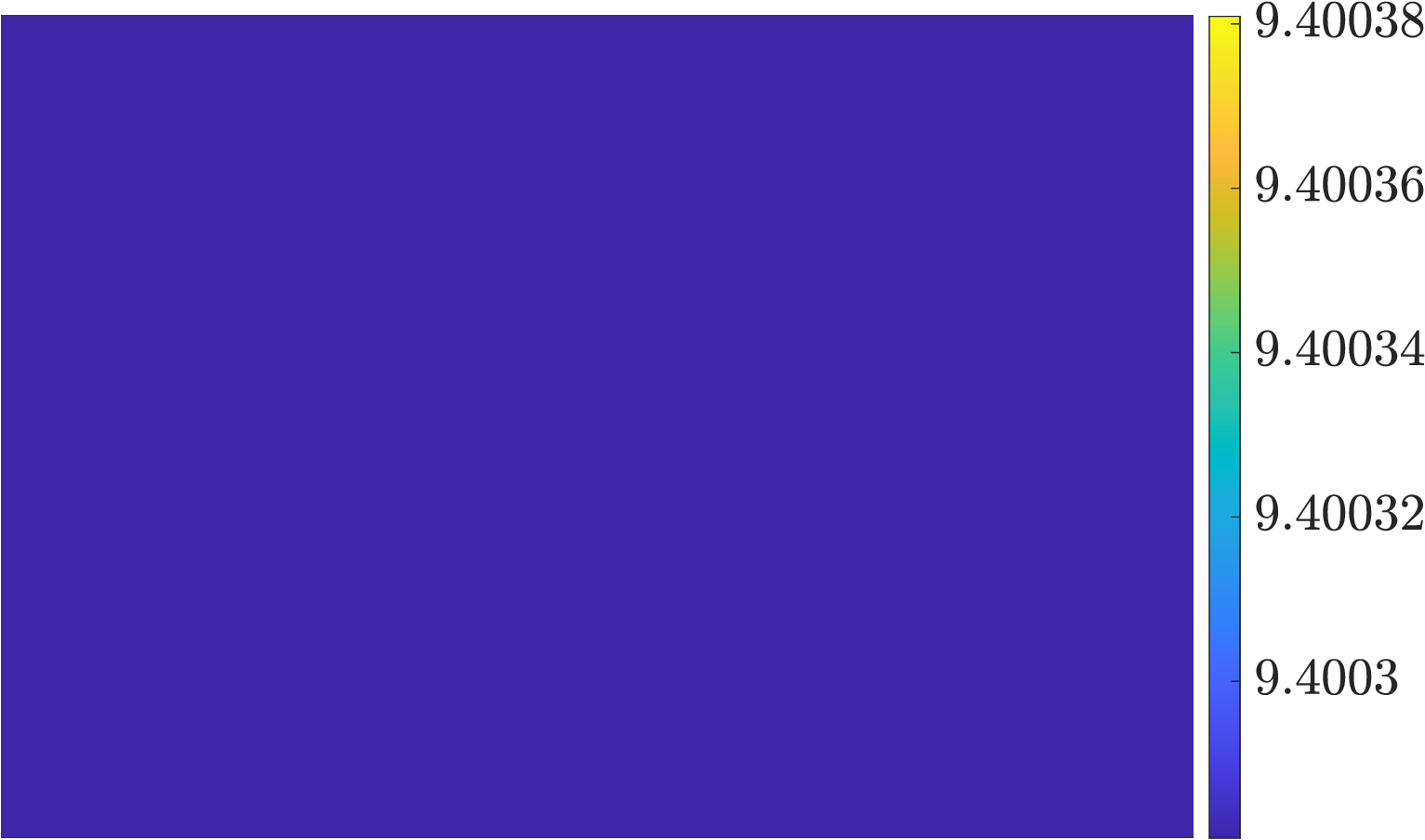}   %[Para su tamaño:anchura con respecto al ancho de la hoja]{Imágenes\Nombre_de_la_imagen}
    \caption{Predicted normal component ($P_{xx}$).}
 \end{subfigure}
 \hfill
 \begin{subfigure}[t]{0.495\linewidth}
    \centering %Para centrar la imagen
    \includegraphics[scale=0.3]{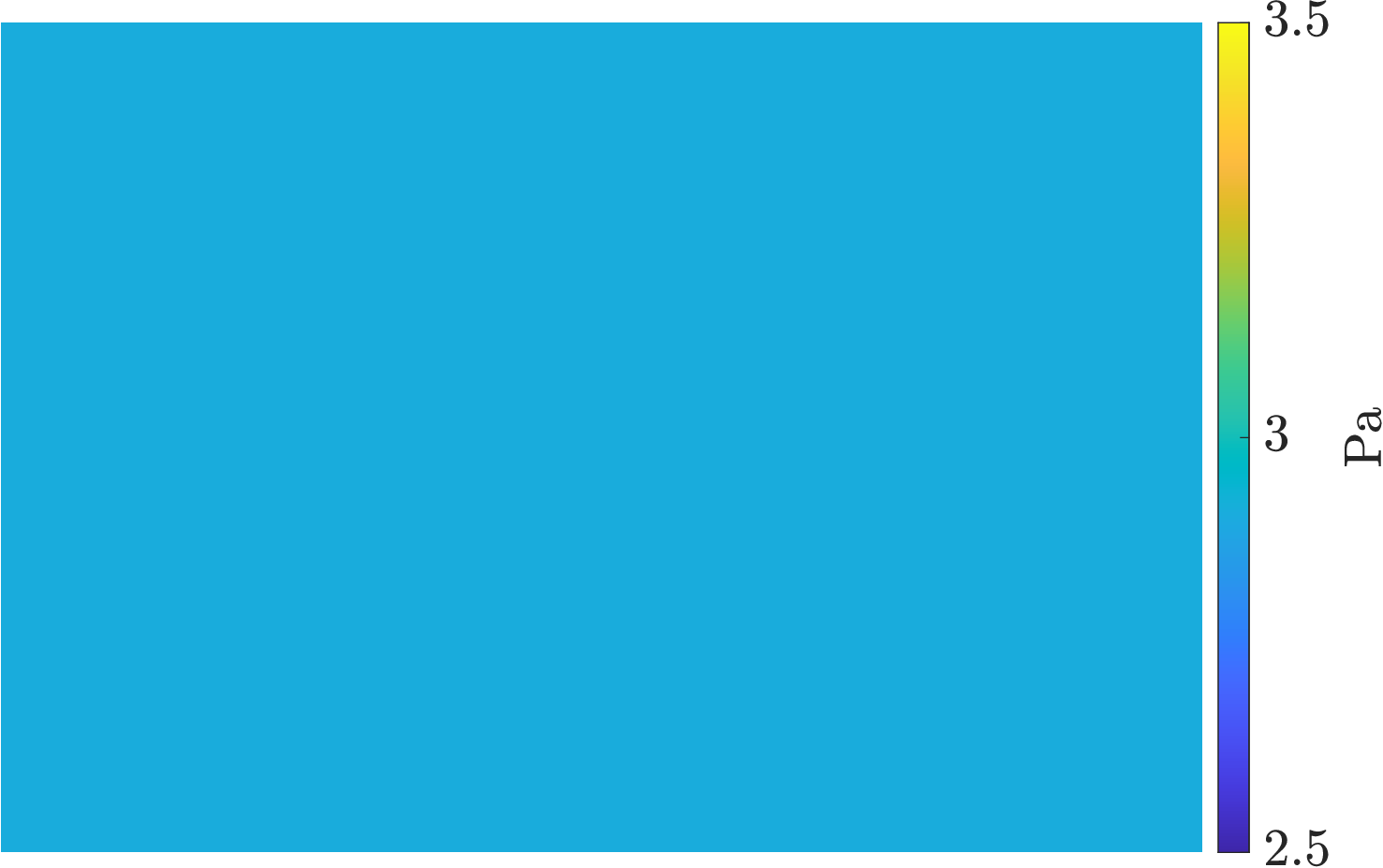}   %[Para su tamaño:anchura con respecto al ancho de la hoja]{Imágenes\Nombre_de_la_imagen}
    \caption{Real normal component ($P_{yy}$).}
 \end{subfigure}
  \begin{subfigure}[t]{0.495\linewidth}
     \centering %Para centrar la imagen
    \includegraphics[scale=0.3]{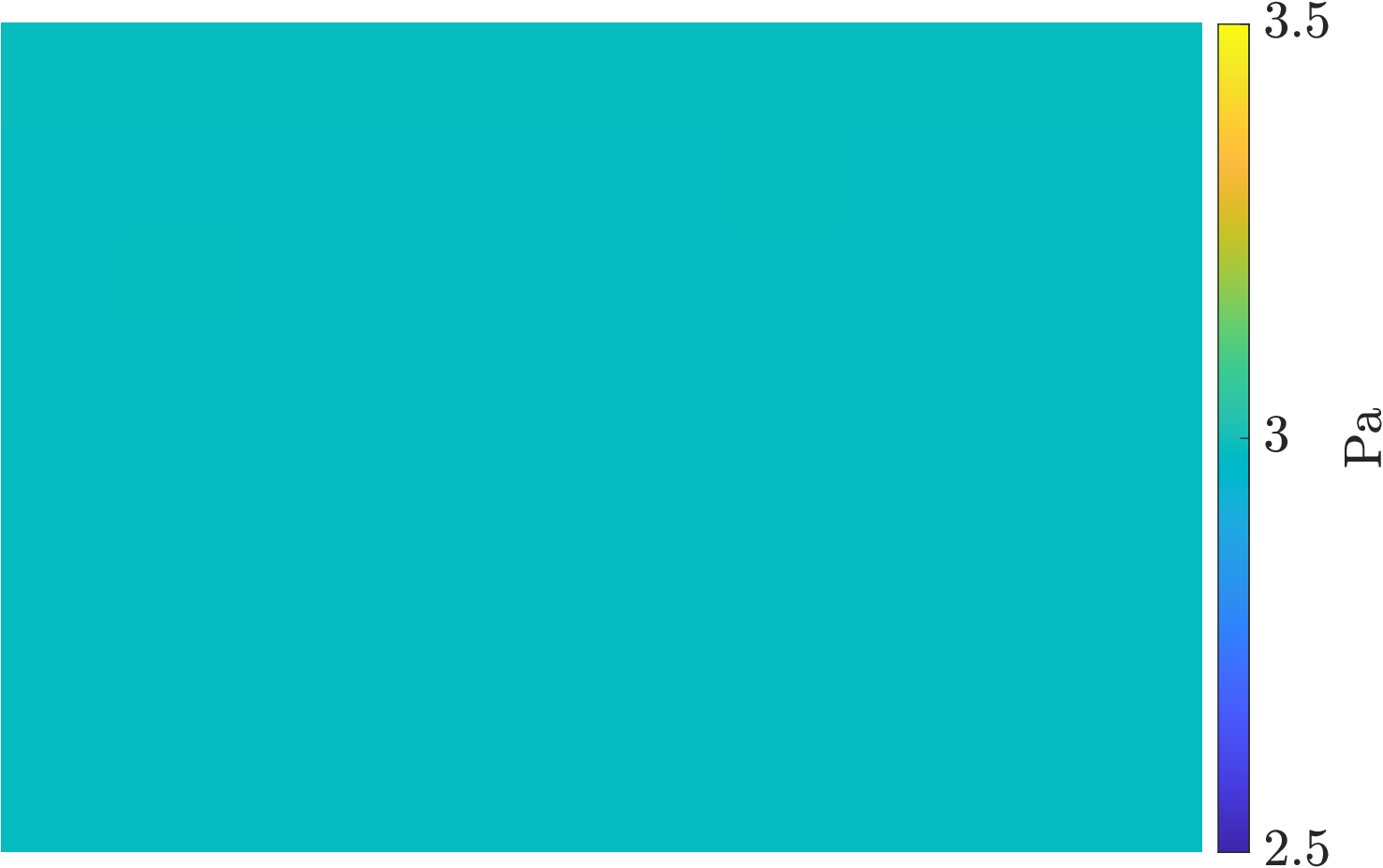}   %[Para su tamaño:anchura con respecto al ancho de la hoja]{Imágenes\Nombre_de_la_imagen}
    \caption{Predicted normal component ($P_{yy}$).}
 \end{subfigure}
 \hfill
 \begin{subfigure}[t]{0.495\linewidth}
     \centering %Para centrar la imagen
    \includegraphics[scale=0.3]{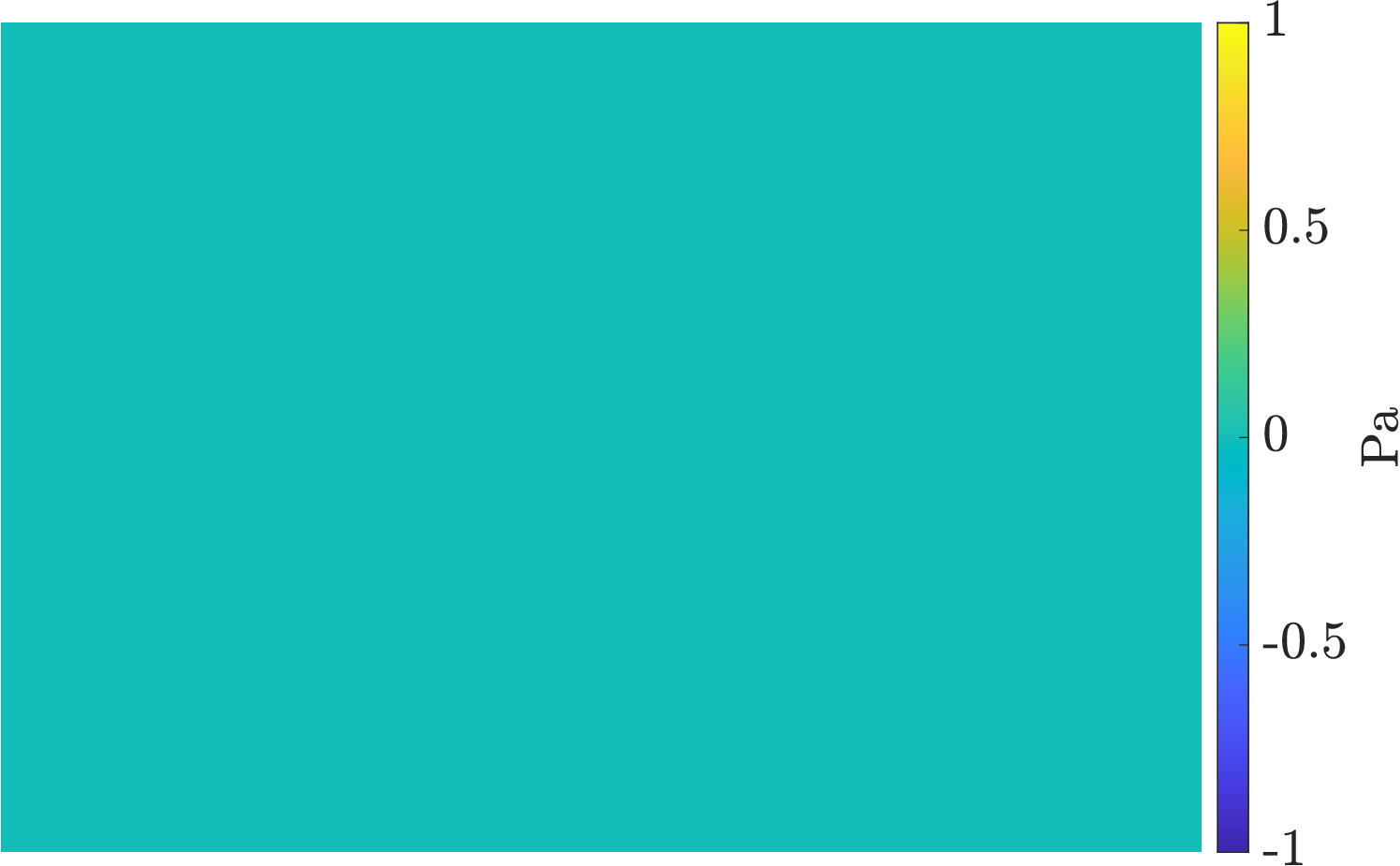}   %[Para su tamaño:anchura con respecto al ancho de la hoja]{Imágenes\Nombre_de_la_imagen}
    \caption{Real shear component ($P_{xy}$).}
 \end{subfigure}
 \hfill
 \begin{subfigure}[t]{0.495\linewidth}
    \centering %Para centrar la imagen
    \includegraphics[scale=0.3]{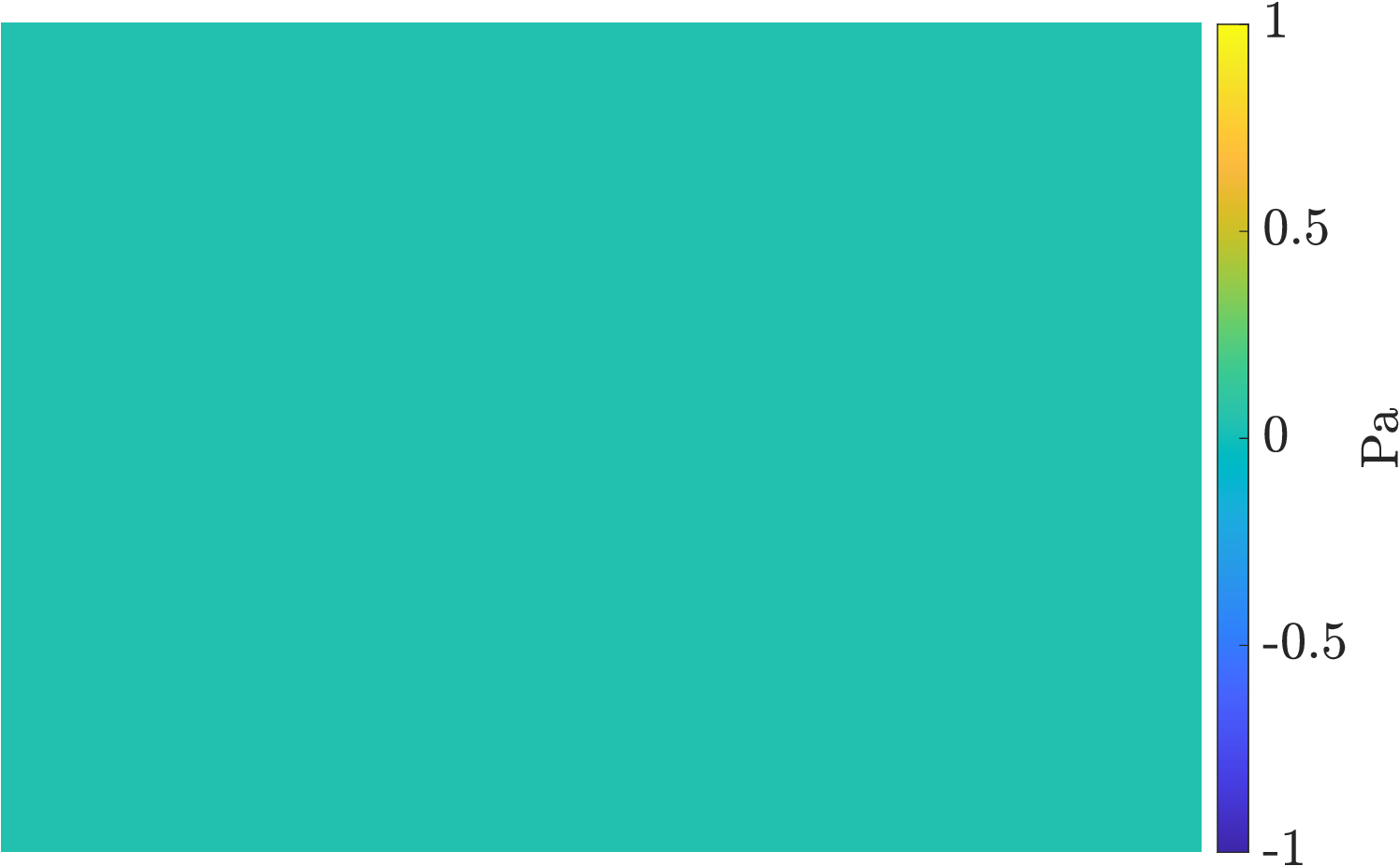}   %[Para su tamaño:anchura con respecto al ancho de la hoja]{Imágenes\Nombre_de_la_imagen}
    \caption{Predicted hear component ($P_{xy}$).}
 \end{subfigure}
\caption{\textbf{PGNNIV prediction versus FEM solution of the components of the stress fields for a single test-set example of the Ogden material.}}
\label{ogdens}
\end{figure}

In Fig. \ref{ogdenu} the displacement field (analytic solution versus PGNNIV prediction) is represented for a test example. In Fig. \ref{ogdene} the components of the strain tensor are illustrated and in Fig. \ref{ogdens} the components of the stress tensor. We observe high similarity between the ground truth and predictive fields despite the coarse discretization.

%\clearpage

\end{appendices}

\end{document}